\documentclass[Afour,times,sageh]{sagej}
\usepackage{moreverb,url}
\usepackage[colorlinks,bookmarksopen,bookmarksnumbered,citecolor=red,urlcolor=red]{hyperref}
\usepackage[noend]{algpseudocode}
\usepackage{algorithm}
\usepackage{amssymb}
\usepackage{bbm}
\usepackage{dsfont}

\usepackage[export]{adjustbox}
\usepackage{graphicx}
\usepackage{import}
\usepackage{xspace}
\usepackage{xurl}
\hypersetup{urlcolor=black}
\definecolor{softgreen}{RGB}{34,139,34} % A visually appealing forest green
\usepackage{array}
\usepackage{flushend}

\usepackage[textsize=small]{todonotes}
\usepackage{listings}
\usepackage{comment}
\usepackage{amsmath} % assumes amsmath package installed
\usepackage{subcaption}
\usepackage{makecell}
\usepackage{xcolor}
\usepackage{url}
\definecolor{gg}{RGB}{0, 155, 85} 
\definecolor{json-key}{rgb}{0.13,0.55,0.13}
\definecolor{json-value}{rgb}{0.25,0.25,0.25}
\definecolor{json-string}{rgb}{0.9,0.3,0.3}
\usepackage{caption}
\usepackage{afterpage}
\usepackage{flafter}      % don't let floats appear before they are defined
\usepackage{placeins}

\usepackage{dblfloatfix} % combines fix2col + stfloats

\begin{document}
%%%%%%%%%%%%%%%%%%%%%%%%%%%%%%%%%%%%%%%%%%%%%%%%%%%%%%%
%%% *** TITLE, AUTHORS, EMAILS, HEADERS *** %%%%%%%%%%%
%%%%%%%%%%%%%%%%%%%%%%%%%%%%%%%%%%%%%%%%%%%%%%%%%%%%%%%
\runninghead{Chatterjee et al.}

\title{Utilizing Inpainting for Keypoint Detection for Vision-Based Control of Robotic Manipulators}

\author{Sreejani Chatterjee\affilnum{1}, Venkatesh Mullur\affilnum{1}, Abhinav Gandhi\affilnum{1}, and Berk Calli\affilnum{1}}

\affiliation{\affilnum{1}Robotics Engineering Program, Worcester Polytechnic Institute, MA, USA}

\corrauth{Sreejani Chatterjee, Worcester Polytechnic Institute, 100 Institute Rd, Worcester, MA 01609, USA.}
\email{schatterjee@wpi.edu}

%%%%%%%%%%%%%%%%%%%%%%%%%%%%%%%%%%%%%%%%%%%%%%%%%%%%%%%
%%% *** ABSTRACTS, KEYWORDS *** %%%%%%%%%%%%%%%%%%%%%%%
%%%%%%%%%%%%%%%%%%%%%%%%%%%%%%%%%%%%%%%%%%%%%%%%%%%%%%%
% NO math, special symbols or citations in the abstract or keywords.

\begin{abstract}
In this paper we present a novel visual servoing framework to control a robotic manipulator in the configuration space by using purely natural visual features. Our goal is to develop methods that can robustly detect and track natural features or keypoints on robotic manipulators that would be used for vision-based control, especially for scenarios where placing external markers on the robot is not feasible or preferred at runtime. For the model training process of our data driven approach, we create a data collection pipeline where we attach ArUco markers along the robot's body, label their centers as keypoints, and then utilize an inpainting method to remove the markers and reconstruct the occluded regions. By doing so, we generate natural (markerless) robot images that are automatically labeled with the marker locations. These images are used to train a keypoint detection algorithm, which is used to control the robot configuration using natural features of the robot. Unlike the prior methods that rely on accurate camera calibration and robot models for labeling training images, our approach eliminates these dependencies through inpainting. 
To achieve robust keypoint detection even in the presence of occlusion, we introduce a second inpainting model, this time to utilize during runtime, that reconstructs occluded regions of the robot in real time, enabling continuous keypoint detection. To further enhance the consistency and robustness of keypoint predictions, we integrate an Unscented Kalman Filter (UKF) that refines the keypoint estimates over time, adding to stable and reliable control performance. We obtained successful control results with this model-free and purely vision-based control strategy, utilizing natural robot features in the runtime, both under full visibility and partial occlusion.
To indicate the broader applicability of the framework, we additionally extend the perception pipeline to a two-module and a three-module soft origami arm and qualitatively evaluate keypoint detection and temporal tracking on this platform.
\end{abstract}

\maketitle

%%%%%%%%%%%%%%%%%%%%%%%%%%%%%%%%%%%%%%%%%%%%%%%%%%%%%%%
%%% *** SECTIONS *** %%%%%%%%%%%%%%%%%%%%%%%%%%%%%%%%%%
%%%%%%%%%%%%%%%%%%%%%%%%%%%%%%%%%%%%%%%%%%%%%%%%%%%%%%%
\section{Introduction} \label{sec:introduction}

\begin{figure*}[!t]
    \centering
    \includegraphics[width=\textwidth]{images/fig1_p1_v4.pdf}

    \includegraphics[width=\textwidth]{images/fig1_p2_v4.pdf}

    \includegraphics[width=\textwidth]{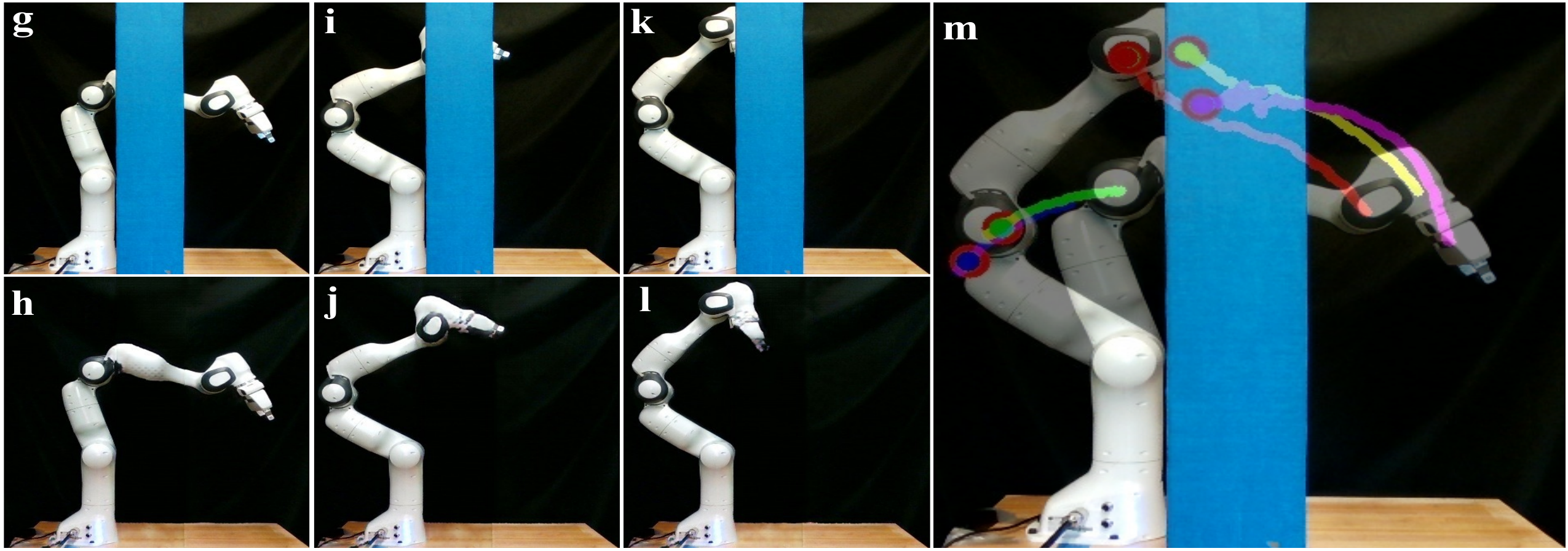}
    \caption{
    Image (a) shows the original images for different robot configurations covering the workspace with ArUco markers attached, and image (b) presents the corresponding binary masks highlighting these markers. Both (a) and (b) are provided as input to a modified LaMa inpainting model, which generates the corresponding reconstructed images shown in (c). The reconstructed images, along with the original ArUco marker centers, are then used to generate a labeled training dataset, as shown in image (d), where the green bounding boxes are drawn with the same orientation of the ArUco markers and the red dots are the marker centers. This dataset is used to train a modified \textit{keypointrcnn\_resnet50\_fpn} model for real-time keypoint prediction. Image (e) illustrates the model's prediction for an arbitrary configuration. (f) depicts the start and goal configurations, along with the resulting control trajectory generated using the predicted keypoints. (g–l) depicts the same control experiment as (f) with occlusion in the environment. (g), (i), (k) show occluded start, intermediate, and goal frames; (h), (j), (l) are their inpainted counterparts. (m) depicts predicted control trajectory under occlusion, corresponding to (f), using inpainted keypoints.
    }
    \label{fig:intro-image}
\end{figure*}

\begin{figure*}[!t]
    \centering
    \includegraphics[width=\textwidth]{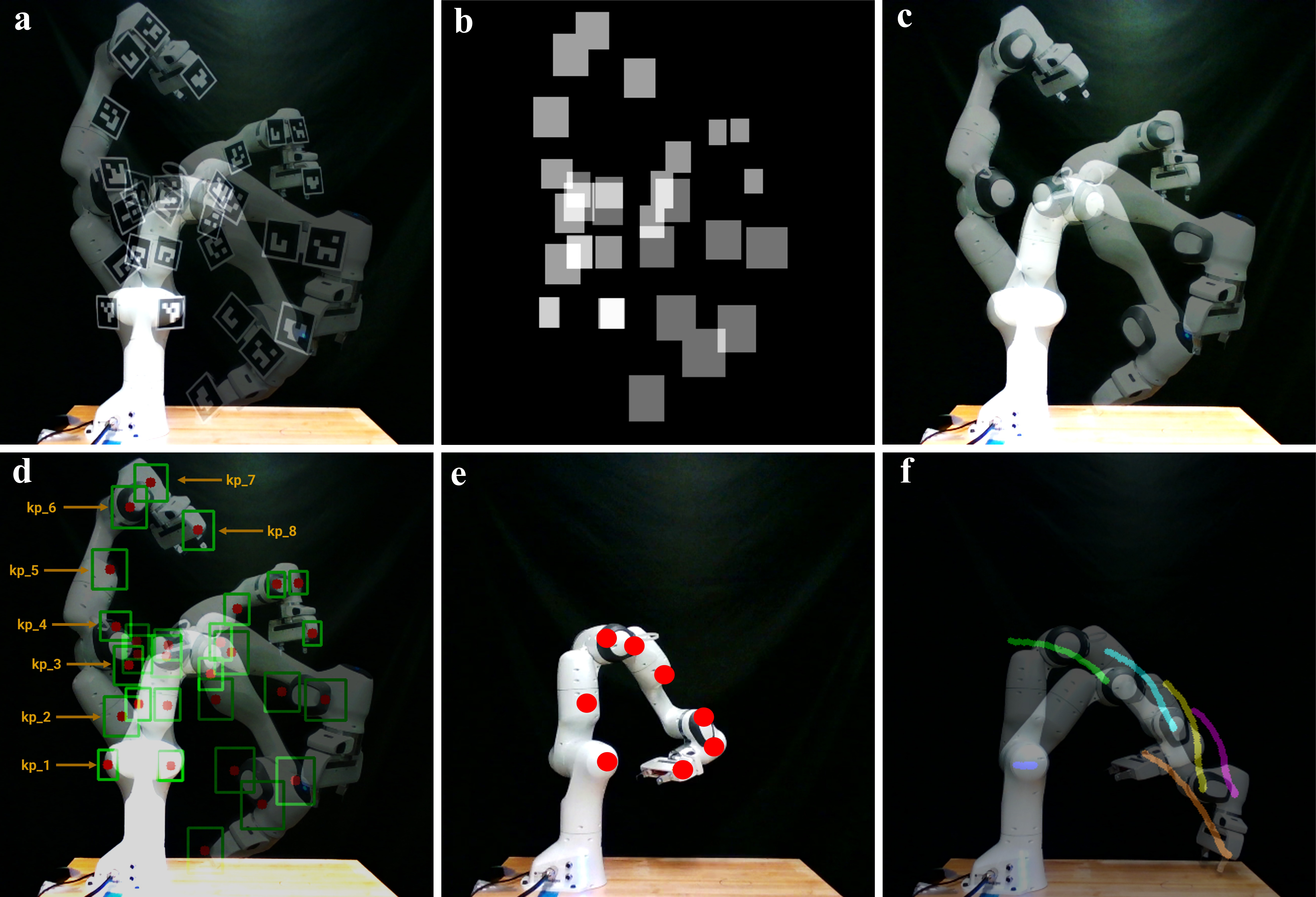}
    \caption{
    The data collection phase for out-of-plane motion. Image (a) shows the original images for different robot configurations covering the workspace in 3D with ArUco markers attached, and image (b) presents the corresponding binary masks highlighting these markers. Both (a) and (b) are provided as input to a modified LaMa inpainting model, which generates the corresponding reconstructed images shown in (c). The reconstructed images, along with the original ArUco marker centers, are then used to generate a labeled training dataset, as shown in image (d), where the green bounding boxes are drawn with the same orientation of the ArUco markers and the red dots are the marker centers. This dataset is used to train a modified \textit{keypointrcnn\_resnet50\_fpn} model for real-time keypoint prediction. Image (e) illustrates the model's prediction for an arbitrary out-of-plane configuration. (f) depicts the start and goal configurations, along with the resulting control trajectory generated using the predicted keypoints. 
    }
    \label{fig:3d_wkflow}
\end{figure*}

Vision-based control methods ~\citep{hutchinson1996tutorial, hashimoto2003review} enable robotic manipulators to operate effectively in unstructured or dynamic environments by utilizing task-relevant visual feedback to close the control loop. These approaches improve robustness to model inaccuracies making them ideal to control systems with complex or variable dynamics \citep{Calli2016, cuevas2018hybrid, ardon2018reaching} such as soft or continuum robots. Model-free visual servoing further minimizes the reliance on predefined robot models \citep{wang2018adaptive, navarro2017fourier} by learning robot-feature motion model during control. This technique is particularly advantageous for robots that are difficult to model accurately, such as soft robots \citep{Lai2020, luo2018orisnake}, underactuated systems \citep{liu2020survey, gandhi2023shape}, 3D-printed platforms \citep{chavdarov2019design, onal2014origami}, and robots built with low-cost hardware \citep{adzeman2020kinematic}. 

Our work focuses on image-based visual servoing (IBVS) in an eye-to-hand configuration \citep{tokuda2021convolutional}, where visual features are extracted from a fixed external camera observing the robot. Mostly in literature, eye-to-hand visual servoing implementations have primarily relied on feature-rich patterns or fiducial markers attached to the robot’s end effector \citep{Lai2020, Calli2016, cuevas2018hybrid, ardon2018reaching} and/or along the body of the robot \citep{gandhi2022skeleton}. If the markers are occluded possibly due to self-occlusion or other objects in the scene \citep{lippiello2005occlusion}, or remain undetected due to motion blur, reflections or lighting conditions \citep{mondejar2018robust}, the control operation may fail. Also, utilizing markers may be inconvenient, impractical, or infeasible in various applications. Our research goal is to advance the limits of image-based visual servoing by removing the reliance on these external markers by using purely natural features along the robot's body in image space, without attaching any external markers. While doing so, we also strive to reduce the dependencies on explicit robot or camera models or proprioceptive sensors on the robot. 

To achieve this we took inspiration from the success of human pose estimation in computer vision for robust keypoint detection in image without using external markers. These algorithms detect the location of specific keypoints corresponding to human body parts, e.g. hands, shoulders, eyes in an image. There are ample datasets available in the literature where these body parts are annotated as keypoints such as MPII Human Pose \citep{Andriluka2014}, COCO test-dev \citep{Lin2014}, Densepose-COCO \citep{guler2018densepose}. These datasets have enabled the training of deep learning frameworks such as OpenPose \citep{Cao2021} and DeepPose \citep{Toshev2014}, which can reliably detect and track human body keypoints in images captured under unconstrained, real-world conditions. Despite their proven success in human pose estimation, similar keypoint datasets for robotic manipulators are sparse in the literature. In \citet{Lee2020} the authors create datasets of three rigid robotic manipulators (Panda \citep{haddadin2024franka}, KUKA \citep{shepherd2014kuka} and Baxter \citep{fitzgerald2013developing}) in simulation. These datasets are then used in a deep learning framework (also known as DREAM) to detect keypoints on real robots. However, in this work, the dataset was created in simulation and the resulting sim-to-real gap prevents the framework's ability to reliably and continuously detect keypoints, which is crucial for real-time control in robotic applications. This was not an issue with DREAM since their goal was one-shot camera calibration and not control. To address the lack of real world data, in our earlier work \citep{chatterjee2023keypoints} we introduced an automated data collection pipeline for annotating keypoints along the body of robotic manipulators. This method labels keypoints on a robot body by combining the camera's intrinsic and extrinsic parameters with the arm's forward kinematics to project 3D joint frames into 2D image space using the pinhole camera model \citep{Zhang2000}. The resulting 2D projections are used as annotated keypoints on the robot's body. The pipeline was designed for physical robotic manipulators, with no reliance on simulation, and is readily extendable to other rigid manipulator platforms. While the approach enabled the identification of robust keypoints for markerless task-space control without online robot or camera models, the data-collection stage still required precise camera calibration and an accurate robot model. As a result, the pipeline is not directly extendable to robots where obtaining kinematic models is often challenging, such as soft or underactuated robots, or robots with inexpensive hardware. 

In this paper we address this shortcoming by removing the reliance of explicit robot or camera model during the data collection phase. To collect data, we placed ArUco markers on the robot's body to mark the locations of the keypoints, as illustrated in \autoref{fig:intro-image}(a) for planar motion and \autoref{fig:3d_wkflow}(a) for out-of-plane motion. We then artificially remove these markers using a modified mask-based GAN inpainting model (LaMa \citep{Suvorov_2022_WACV}), depicted in \autoref{fig:intro-image}(c). The center of the markers are our annotated keypoints and the reconstructed image generated after removing the markers are our training images, as shown in \autoref{fig:intro-image}(d). This data is then used to train a modified Keypoint R-CNN \citep{Patil2021} model to generate keypoints in run-time to be used as natural visual features to achieve markerless visual servoing \citep{chatterjee2023keypoints, chatterjee2024utilizing} as demonstrated in \autoref{fig:intro-image}(e),(f). Up until this phase, the pipeline operates under the assumption that the entire manipulator remains visible throughout the task. However, this assumption may not hold in unstructured or uncertain environments where continuous visibility is often impractical. 

We address this challenge of partial occlusion in the robot's image-space representation in purely vision-based control systems by introducing an attention U-Net based WGAN image inpainting model to reconstruct parts of the robot occluded by obstructions in the image. In \autoref{fig:intro-image}(g),(i),(k) are the different configurations of moving robot with occlusion in the environment and \autoref{fig:intro-image}(h),(j),(l) are the corresponding reconstruction where the occlusions are removed from the images.  Our novel mask-free approach for inpainting leverages Wasserstein Generative Adversarial Networks with Gradient Penalty \citep{gulrajani2017improved}. Specifically, we employ an Attention U-Net \citep{oktay2018attention} as the generator and a spectral-normalized convolutional critic that outputs a single realism score over the whole image, rather than a patch-based discriminator \citep{isola2017image}, so that the critic captures the global structural consistency of the reconstructed robot. The gradient penalty ensures stable and smooth gradient flow during training. Our method directly transforms occluded regions of the robot into their occlusion-free counterparts, without requiring explicit masks or prior modeling of the occlusions. The reconstructed images are then passed through the trained Keypoint R-CNN model for real-time keypoint detection. As demonstrated in \autoref{fig:intro-image}(m), the detected keypoints are then used as natural visual features in an adaptive visual servoing scheme, where the Jacobian is estimated online using a window-based least squares optimization technique \citep{navarro2017fourier}, significantly reducing the reliance on explicit robot or camera model to compute the interaction matrix. To enhance robustness, we further incorporate an Unscented Kalman Filter (UKF) \citep{wan2000unscented} into the pipeline. This is particularly useful in cases where occasional noise or motion blur in the reconstructed images leads to inaccurate or missing keypoint detections. Given the adaptive and nonlinear nature of our system, we choose the UKF, which is better suited than linear filters for handling missing keypoints and ensuring continuity in the control loop.

To verify the effectiveness of the inpainting model, we use objects of different sizes, shapes, textures, and colors to occlude the robot in the image. We observe that larger occlusions degrade reconstruction quality, leading to an increased rate of missing keypoints and causing the UKF to rely more heavily on its prediction step rather than the measurement update, ensuring continuity in the control loop. 

To evaluate the robustness of our pipeline, we conduct adaptive visual servoing experiments under three conditions: full visibility, partial occlusion with smaller coverage, and partial occlusion with larger coverage. Under full visibility, we also evaluate the pipeline’s out-of-plane (3D) motion-control capability. In this setting the system produces smooth trajectories and robust convergence. With smaller occlusions, we observe only minimal degradation in tracking accuracy and control performance. In contrast, larger occlusions lead to noisier trajectories and slower convergence. Nevertheless, the system still reliably reaches the goal, demonstrating strong resilience to partial visual obstructions and uncertainties in the environment. To the best of our knowledge, this is the first eye-to-hand IBVS pipeline to achieve 3D motion control using a single camera and natural 2D keypoints, with a control loop that uses no depth sensing, no camera calibration, and no kinematic robot model at runtime. 

We validate the full control pipeline on a rigid Franka Emika Panda arm, which gives us a controlled testbed with reliable ground truth for both the perception and the control evaluation. To indicate that the perception pipeline carries over to platforms where accurate kinematic models are difficult to obtain, we additionally apply it to a soft origami arm and report qualitative keypoint detection and tracking in \autoref{ssec:origami_extension}.

Our framework introduces the following key novelties:
\vspace{-0.4em}
\begin{itemize}
    \item Eliminates reliance on external markers without requiring manually labeled keypoint datasets.
    \item Enables model-free, vision-only control by reducing dependence on calibrated robot or camera models or proprioceptive sensing, making it ideal for robots with limited or unreliable onboard sensing.
    \item Resolves configuration ambiguity in redundant manipulators by leveraging full-body keypoints for more consistent control.
    \item Enables out-of-plane motion control using a single camera and 2D image keypoints without depth estimation, stereo, or explicit robot/camera models.
    \item Enables vision-based control without the need for computationally expensive mask generation for occlusion detection during training. This also allows the method to generalize to arbitrary shapes, sizes, and textures of occlusion, since predefined masks are not used to determine occlusion boundaries.

\end{itemize}

We clarify that the keypoint detector is trained on images of the specific robot and therefore encodes the robot's visual appearance. Our claim of being model-free refers to the data-collection and runtime control stages, neither of which requires a kinematic model, camera calibration, depth, or encoder feedback. The learned detector replaces these with a mapping obtained from data rather than from an explicit model.

This work has the following assumptions and limitations:
\vspace{-0.4em}
\begin{itemize}
    \item The training data for both the inpainting and keypoint-detection models were collected in our experimental setup against a black backdrop, and we did not retrain on varied backgrounds. We do, however, evaluate how the clean-trained detector transfers to cluttered scenes without retraining in \autoref{ssec:bkgrnd_robustness}. Broader robustness would be expected from a larger dataset spanning more conditions, but our main focus here is to present the novel strategies for vision-only control with natural keypoints.
    \item Occlusion covering a larger area and occlusion with a similar shade as the robot often leads to reconstruction errors and missed keypoints.
    \item The experimental setup used static occlusion; dynamic occlusion scenarios were not evaluated.
    \item The occlusion dataset and the control evaluation under occlusions were limited to planar (2D) configurations.
    \item The extension to the soft origami arm validates keypoint detection and temporal tracking only. Closed-loop control on the soft arm is left for future work.
     \item Occlusion handling stays tied to the background it was trained on. The detector transfers to a cluttered scene on the keypoints that remain visible, but neither inpainting nor a detector retrained on occlusion recovers the occluded keypoints against an unfamiliar background, as shown in \autoref{ssec:occ_clutter}. Occlusion in a changed background remains an open problem.
\end{itemize}

To the best of our knowledge, this work presents the first vision-based control approach that uses natural features along the robot's body and requires no kinematic robot model, no camera calibration, and no sensing other than the camera, in particular no encoders, at control time. The robot-specific information that the system relies on is acquired by training the keypoint detector on robot images, and not supplied through an explicit geometric model.

An overview of our entire pipeline, including marker removal for dataset generation, keypoint training, prediction, and visual servoing with and without occlusion, is shown in \autoref{fig:intro-image}.

% This article is an extension of a conference paper that appeared in the proceedings of the 2024 IEEE International Conference on Robotics and Automation (ICRA) \citep{chatterjee2024utilizing}. In the original paper, we controlled $2$ planar joints using $3$ natural features without any collision avoidance. In this current article, we developed and integrated a novel occlusion handling component strategy. We have also designed a temporal filtering (UKF) to improve the system's robustness to detection uncertainties. We added experiments for assessing the performance of the system under visual occlusions. We also extended the setup to $3$ planar joints with $5$ keypoints, increasing the number of joints and showing demonstrating the framework's capability of handling actuation redundancy.

\section{Novelty Statement and Differences from Prior Publication} \label{sec:novelty}
    % Novelty Statement and Differences from Prior Publication
This article is an extension of our conference paper that appeared in the proceedings of the 2024 IEEE International Conference on Robotics and Automation (ICRA) \citep{chatterjee2024utilizing} [DOI: 10.1109/ICRA57147.2024.10610006]. In the original paper, we controlled $2$ planar joints using $3$ natural keypoints without considering any occlusion in the workspace during control. In this current article, we scaled the setup to $3$ planar joints controlled by $5$ keypoints and $3$ planar and $1$ spatial joints, controlled by $6$ keypoints. We expanded the data collection and control experiments according to the newly scaled setup. To the best of our knowledge the current system demonstrated control in out-of-plane motion using IBVS in eye-to-hand setup without using depth, stereo setup or robot/camera model. We also developed and integrated a novel occlusion handling component that preserves keypoint detection under partial occlusion during control. We have also added a temporal filtering (UKF) to improve the system's robustness to detection uncertainties. We added experiments for assessing the performance of the system under visual occlusions. We evaluated whether the inpainting stage is necessary by comparing it against temporal filtering alone and against a keypoint detector retrained on occluded images. We measured the per-frame latency of every stage of the perception pipeline to establish that the system operates within the budget of the control loop. We assessed the robustness of the pipeline to a cluttered laboratory background that is absent from the training data, examined whether the occlusion handling transfers to that background, and closed the control loop in the same scene for out-of-plane motion. We also evaluated the online interaction-matrix estimate directly through its conditioning and through the agreement between the feature motion it predicts and the feature motion that is observed, since the analytical Jacobian that would serve as a reference is unavailable in our model-free setting. Finally, we extended the perception pipeline to a two-module and a three-module soft origami arm to indicate that the approach carries to platforms where accurate kinematic models are difficult to obtain.
\section{Related Work} \label{sec:related}
\subsection{Visual Servoing:}
By definition, visual servoing is the seamless integration of robotic vision with control \citep{sun2018review}, where the robot adjusts its motion based on what it sees \citep{chaumette2016visual}. The concept of controlling a robot using a camera was first proposed by \citet{hill1979real} to give the robot the flexibility to see the task space and react to changes in the environment in real time. This makes visual servoing systems much more reliable than traditional control methods \citep{siciliano2009robotics} in uncertain conditions, especially when controlling robots without precise models or when encoder feedback is unreliable or unavailable. 

Broadly, visual servoing methods in the literature are classified into two main categories: position-based visual servoing (PBVS) and image-based visual servoing (IBVS) \citep{hutchinson1996tutorial}. In PBVS \citep{dong2015position}, image features are used to estimate the pose of the target relative to the camera by leveraging the geometric model of the object and the known camera parameters. In IBVS \citep{sanderson1983image}, the image features are directly used to compute the control error, making the control operate in the image space rather than in Cartesian space. Our work focuses on IBVS methods, as they have much less reliance on accurate models of the robot, camera, or environment. 

IBVS can be broadly categorized into two camera configurations: eye-in-hand \citep{mohebbi2016comparative} and eye-to-hand \citep{gonccalves2008uncalibrated}. In an eye-in-hand configuration, the camera is mounted on the robot and moves along with it. In contrast, an eye-to-hand setup places the camera in a fixed position, observing the robot from the outside. Our pipeline is similar to eye-to-hand image-based visual servoing, but utilizes features on the robot body, not only the end effector: In conventional image-based visual servoing setups, especially in an eye-to-hand configuration, the image features are typically extracted from the robot's end-effector. However for redundant manipulators \citep{de2008visual, nazari2022visual}, this can lead to configuration ambiguity, where multiple joint configurations produce the same end-effector position. As demonstrated in \citet{huang2023novel}, relying solely on end-effector visual features can result in convergence instabilities due to excess degrees of freedom, potentially leading the robot to suboptimal or unstable poses that violate task constraints. To overcome this limitation, \citet{gandhi2022skeleton} proposed representing the robot’s shape as a B-spline curve and using the control points of the curve as visual features. These control points, distributed along the robot’s body, enable control in configuration space rather than relying solely on the end-effector. Building on this idea, our approach annotates and predicts keypoints spread across the entire robot body, providing a more informative representation of visual features. This design allows the system to uniquely identify full-body goal configurations that result in the same end-effector position in the image, ensuring more stable control trajectories.

\subsection{Markerless Image-Based Visual Servoing:}
    In the literature on traditional IBVS, control features are frequently derived from fiducial markers such as ArUco \citep{garrido2014automatic}, AprilTags \citep{wang2016apriltag}, or QR codes \citep{bach2023application}, which are attached to the robot or its environment. Apart from practical limitations of using markers, their effectiveness can degrade under self-occlusion \citep{lippiello2005occlusion}, challenging lighting conditions, or detection failures caused by motion blur, reflections \citep{mondejar2018robust}, or incorrect identification by marker libraries. Such issues introduce risk to the stability and reliability of the visual servoing system, especially in dynamic or unstructured environments. 
    To mitigate this \citet{vicente2017towards} presents a markerless IBVS pipeline for humanoid robots using stereo vision to perform grasping tasks. Their approach estimates the 6D pose of the robot’s hand by matching rendered hand images with observed stereo pairs using HOG features and a particle filter, followed by decoupled image-based control laws for translation and rotation. The authors of \citet{fantacci2018markerless} also propose a markerless visual servoing pipeline for humanoid robots using stereo vision. Their method involves explicit 3D object modeling via superquadrics, 6D end-effector pose estimation with a particle filter, and a decoupled image-based visual servoing. In contrast to both the above approaches our system bypasses the need for stereo vision, explicit 3D pose estimation or object geometry by relying solely on 2D keypoints extracted from monocular RGB images, enabling a more generalizable markerless visual servoing framework.
    Even though authors of \citet{xie2024markless} control an uncalibrated and underactuated robot using a markerless IBVS scheme, their system uses an eye-in-hand setup and depends on a few hand-crafted global features, such as centroids, higher-order moments, and brightness fields, that are closely tied to the specific geometry of their robotic assembly. In contrast, our approach uses 2D keypoints extracted from RGB images, making it more generalizable across different robots and applications.

% --- ADDED (R2.3a): point-cloud / ICP registration as a different design point ---
    Another family of markerless methods registers an observed point cloud to a known model of the robot or target through iterative closest point matching \citep{besl1992icp}, and tracks the configuration by aligning the model to incoming depth measurements \citep{schmidt2014dart}. These approaches require both a depth sensor and an accurate geometric model of the body being tracked, which are the two dependencies our method avoids. They are also tied to objects for which a rigid or articulated model already exists, and do not extend naturally to under-modeled platforms. In contrast, our approach operates on 2D keypoints from a single RGB image, without depth or a geometric model, and therefore applies to platforms for which such a model is unavailable.
% --- end added ---

\subsection{Image-based Visual Servoing in out-of-plane motion:}
In literature, 3D IBVS with an eye-to-hand setup is quite sparse. \citet{chen2005adaptive} use the classic \(2\tfrac{1}{2}\text{-D}\) formulation, relying on homography-based feedback with calibrated camera intrinsics to track end-effector motion. In \citet{kulpate2005eye}, the authors propose an inventive fixed-camera scheme that places a flat mirror in the field of view to create a virtual stereo geometry, enabling 3D positioning without a second camera or prior calibration. However, the mirror alters the scene and adds hardware constraints. In \citet{xu2016uncalibrated} the authors derive a depth-independent interaction matrix and identify the camera projection model online, avoiding offline calibration but still imposing a parametric projection model during control. Their evaluation is limited to simulation with no real-world control experiments conducted. In contrast, our system uses only a single fixed camera and natural 2D keypoints, estimates the Jacobian online, and demonstrates 3D motion control without depth sensing, homographies, stereo, or any robot/camera model, with extensive real-world experiments that include within and out-of-plane motions.

\subsection{Visual Servoing under Occlusions:} 
    While IBVS enables a robot to complete tasks by directly controlling image space error, it inherently relies on continuous visibility of identifiable control features. However, a problem arises when these features are occluded by other objects in the scene. 

    To mitigate the above limitation, authors of \citet{kragic2003robust} designed a robust visual servoing pipeline that combine multiple image cues such as color, motion and template correlation to track objects under visually challenging conditions such as partial occlusion. Their method leverages a geometric 3D wireframe model of the target object. In case of temporary loss of object features due to occlusion, the system uses the known object model and robot pose estimation to predict where the missing features should be in the image, enabling continuous tracking. However, this approach relies on accurate camera calibration and robot's kinematic to align 3D pose with 2D image features. In \citet{iwatani2008visual} authors introduced a visual servoing framework that handles occlusion by reconstructing missing image features using the image Jacobian and previously extracted features of a partially visible target object. Their approach also relies on accurate camera and robot model to compute the image Jacobian. Their method relies on a linear approximation of the mapping from joint space to image space, requiring accurate knowledge of both the robot's kinematic model and the camera extrinsics to compute the Jacobian and perform feature reconstruction. While the above two methods can only mitigate partial occlusion of the target object, authors of \citet{fleurmond2016handling} proposed a framework to handle complete visual feature loss during IBVS, by reconstructing the 3D structure of the target object using multi-view geometry. Their method builds a 3D structure of the target object when it is fully visible using multiple cameras pointing from different angles. In the event of complete or partial visual feature loss, the previously estimated 3D structure is used to reproject the object's features in the image space. This approach, however, requires accurate geometric robot model and precise camera calibration to compute the necessary transformations between the camera and end-effector frames. Among the most recent works, authors of \citet{zhang2023occlusion} proposed an occlusion-aware IBVS framework, that combines Model Predictive Control (MPC) Probabilistic Control Barrier Certificates (PrCBC) to proactively avoid occlusion. Their method predicts whether visual features will become occluded by projecting known 3D features and obstacles into image space and estimating occlusion risk probability, based on depth uncertainty. Robot motion is then constrained to maintain visibility with high probability. While effective, this approach relies on accurate robot models, camera calibration, and prior knowledge of obstacle geometry. Similarly in \citet{nicolis2018occlusion}, the authors propose an occlusion-avoidance strategy for dual-arm teleoperation. In their setup, one arm is teleoperated by a user to perform a task, while the second arm, equipped with an eye-in-hand camera, autonomously moves to maintain an unobstructed view of the tool in the other arm. Their method detects potential occlusions by monitoring the image-space distance between the tool and nearby objects and formulates this as a visibility constraint in a real-time optimization problem. The camera arm is then adjusted only when occlusion is imminent, ensuring continuous visual feedback. While this method effectively maintains visibility, it relies on accurate robot and camera calibration and known 3D geometry of the environment. 
 
    All the above approaches rely on explicit robot model and/or accurate camera calibration. Moreover, these methods are primarily designed for IBVS in eye-in-hand setups, where the camera moves with the manipulator and the occlusion typically affects a static target object. In comparison, occlusion handling in eye-to-hand configurations, where a fixed camera observes a moving robot's body, is virtually underexplored, especially when the robot itself becomes the occluded subject, as is the case in our research setup. Unlike above methods, our system avoids reliance on known robot models, or precise camera calibration. Instead, we handle occlusion by using a deep learning-based image inpainting model to reconstruct the occluded robot parts directly in image space, enabling robust feature detection and control.

% --- ADDED (R2.2): distinguish from learning-based / generative control policies ---
\subsection{Learning-Based Visual Control:}
A separate class of vision-based manipulation methods learns control policies directly from demonstrations using generative models, such as diffusion policies \citep{chi2023diffusion} and flow-matching policies \citep{lipman2023flow, chisari2024flowmatch}, often evaluated on planar manipulation tasks such as block pushing. Diffusion policies in particular operate from a single fixed camera using natural visual features, without explicit depth or robot models. However, these methods differ from our setting in two respects. They learn a task-specific policy from a dataset of demonstrations, whereas our controller uses no demonstrations and closes the loop online through adaptive Jacobian estimation toward a goal configuration specified directly in image space. The keypoints we detect form a general representation of the robot body that is reused across tasks, rather than a policy trained for a single task. In contrast to these imitation-based generative policies, our framework performs closed-loop, demonstration-free adaptive control.
% --- end added ---

\vspace{1mm}
\subsection{Image Inpainting Techniques:} 
    With the advent of deep learning, image inpainting \citep{Romero_2022_CVPR}, the technique of restoring damaged or absent areas of an image, has advanced significantly. Traditional methods used patch-based \citep{guo2017patch} or diffusion-based techniques \citep{bertalmio2000image}, filled in missing areas with low-level features, assuming that there exist a number of similar patches in the image. Despite being computationally efficient, these approaches frequently yielded hazy or unrealistic results, especially for intricate textures, non-repetitive, and rotation variant regions. \par
    
    Using texture generation techniques, early image inpainting works by \citet{criminisi2004region} and \citet{efros1999texture} tried to address the issue by replicating background patches into holes either propagating from hole boundaries or spreading from low-resolution to high- resolution. Although these methods are effective for background inpainting applications that include consistent textures or linear structures, they struggle with complex scenarios involving intricate and diverse structures, such as real-world scenes. Furthermore, these methods rely on low-level features and lack the high-level semantic understanding required for coherent reconstruction, making them inadequate for generating novel but contextually appropriate content. \par

    Before the advent of adversarial methods, early deep learning approaches relied on autoencoders, that used neural networks to reconstruct missing image regions. Autoencoders \citep{tu2019facial, givkashi2022image} compress the input into latent distribution parameters using downsampling, and the decoder attempts to reconstruct the image. The decoder is trained to minimize reconstruction loss (e.g., mean squared error). \citet{givkashi2022image} investigated a guided selection process to improve the quality of autoencoder-based inpainting under specific conditions. However, the effectiveness of the model depends heavily on the shape and size of the occluded region, and it is uncertain whether the network truly captures the semantic context (i.e., what kind of object or structure should be filled in) needed for coherent reconstruction. \citet{tu2019facial} examined the use of variational autoencoders for face image inpainting leveraging deep features of the AR face dataset. Although this model showed promise in reconstructing appearance details, its reliance on reference averaging and pixel-wise loss often led to blurry outputs and lacked the high-level semantic understanding needed to accurately reconstruct structured and articulated objects like robotic manipulators, where the spatial arrangement and continuity between links must be preserved for functional integrity.\par

    Convolutional neural networks, generative adversarial networks, variational autoencoders \citep{quan2024deep, XIANG2023109046} have revolutionized image inpainting by learning semantic priors and meaningful hidden representations through end-to-end training. In order to handle the multi-modal nature of the inpainting problem (i.e., for any given occluded region, there may exist multiple plausible ways to fill it in, based on the neighboring context), \citet{pathak2016context} created Context Encoders, which were the first to employ convolutional neural networks for picture inpainting with an encoder-decoder architecture trained using both reconstruction and adversarial losses. Building on this foundation, \citet{yu2018generative} developed a generative inpainting framework that incorporates a contextual attention, allowing the network to borrow features from distant spatial locations and enhance performance on irregular occlusions. By using Fast Fourier Convolutions to capture image-wide context and preserve structural consistency over vast missing sections, \citet{Suvorov_2022_WACV} advanced the field with Large Mask Inpainting (LaMa). \par

    A significant challenge in the conventional inpainting techniques is the reliance on explicit masks to identify the occluded regions for a smooth reconstruction of an image. In real-world robotic applications, accurate mask generation is often impractical due to the uncertainty of occlusions, especially in dynamic environments. A self-supervised method for occlusion-aware learning was presented by \citet{wang2018occlusion}, which addresses some of the above issues but still necessitates an explicit occlusion map during training. Many existing occlusion-handling networks rely on modeling the occlusions \citep{wang2018occlusion} or explicit occlusion detection, both of which impose extra sensory and processing demands that could impede real-time applications. \par

% Related Work
    \subsection{Generative Adversarial Networks for Image Reconstruction}
    Through adversarial training, in which a generator competes with a discriminator network, Generative Adversarial Networks (GANs) have revolutionized image synthesis and reconstruction. This efficacy was shown by \citet{demir2018patch} using a patch-based GAN for image inpainting. Traditional GANs, however, suffer from training instability. Wasserstein GANs \citep{arjovsky2017wasserstein} provided a more stable process through the use of the Wasserstein distance, and the gradient penalty term of \citet{gulrajani2017improved} (WGAN-GP) further enhanced stability. \par

    Context Encoders \citep{pathak2016context} were the first to use encoder-decoder topologies with adversarial losses for producing semantically meaningful output in large missing regions, whereas \citet{wei2022image} effectively integrated U-Net architectures with adversarial discrimination for inpainting. GAN-based inpainting has been enhanced by a number of significant developments like Global and local discriminators proposed by \citet{iizuka2017globally} in order to preserve both local consistency and overall coherence. Recent advances include specialized convolution operations that improve outcomes for irregular masks, such as Gated Convolutions \citep{yu2019free}, which generalizes binary masking to a learnable mechanism, and Partial Convolutions by \citet{liu2018image}, which employ masked and re-normalized convolutions.

    Recent inpainting advances include diffusion models like RePaint \citep{lugmayr2022repaint} that yield excellent results but necessitate computationally costly iterative sampling. Swin Transformers \citep{zhou2023superior} that use self-attention with shift windows for long-range dependencies but result in artifacts at patch boundaries that impact keypoint detection. Global structure is difficult to capture by conventional CNN-based techniques, particularly for huge missing regions. This is addressed by MAT (Mask-Aware Transformer) \citep{li2022mat}, a revolutionary inpainting-oriented transformer block that ensures more coherent reconstructions without imposing undue processing expense by dynamically updating masks to refine attention across valid regions. Similar to this, the T-former reduces the quadratic complexity of conventional transformers by proposing an effective linear attention mechanism, which makes them more appropriate for high-resolution image inpainting \citep{deng2022t}. Transformer-based self-supervised GANs have also been investigated as a way to improve reconstruction quality while preserving computational effectiveness \citep{zhou2023superior}.
    
     Mask-free and dynamically generated mask techniques have been investigated in recent inpainting developments, which lessen the need for explicit occlusion detection. Using GAN-inversion, \citet{mou2023rgi} automatically detects and restores missing areas without the need for predetermined masks. Relaxed-RGI fine-tunes the GAN manifold to improve the accuracy under severe corruptions. Similarly, to detect inconsistencies, VCNet \citep{wang2020vcnet} uses a mask prediction network to do blind inpainting. This is followed by robust filling, which is appropriate for uncertain occlusions. For interactive mask creation, Inpaint Anything \citep{yu2023inpaint} combines segmentation (e.g., SAM) with models such as LaMa \citep{Suvorov_2022_WACV}, whereas Empty Cities \citep{bescos2019empty} eliminates dynamic items using semantic segmentation, producing masks for GAN-based inpainting.

    Even though Large Mask Inpainting (LaMa) \citep{Suvorov_2022_WACV}, which leverages Fast Fourier Convolutions to capture image-wide context, was highly effective in our data collection pipeline, specifically for removing ArUco markers from static images where masks could be generated easily and accurately, it remains unsuitable for real-time occlusion handling in robotic control tasks. In dynamic settings, where occlusions are unpredictable and vary over time, LaMa’s dependence on explicitly defined masks poses a significant limitation. These observations motivate our mask-free inpainting strategy, which combines WGAN-GP and Attention U-Net architectures, specifically designed to operate in real-time robotic visual servoing scenarios. Our method addresses the gap left by conventional inpainting models, which are often tailored for general image restoration tasks and not optimized for the fast, online operation and structural integrity required in real-time robotics applications. \par

\section{Methodology} \label{sec:overiew}

\begin{figure*}[!t]
    \centering
    \includegraphics[width=0.75\textwidth]{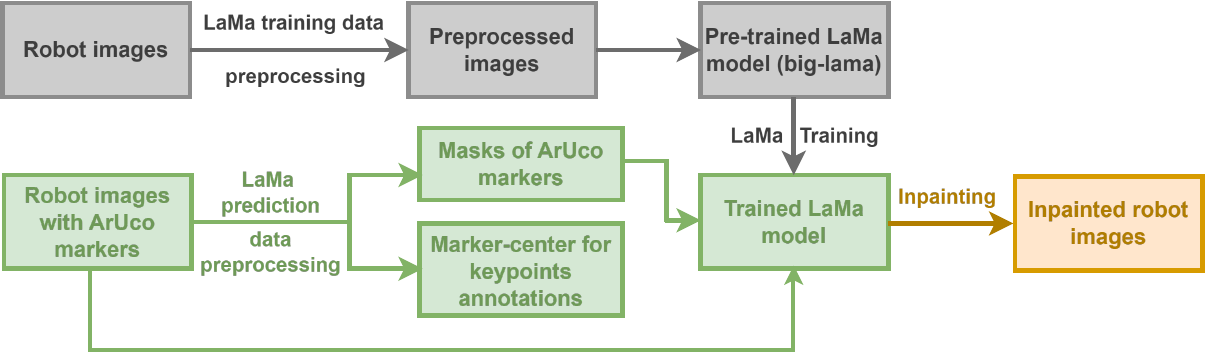}
    \caption{
    Overall data-collection pipeline. 
    }
    \label{fig:data_flow}
\end{figure*}
\begin{figure*}[!t]
    \centering
    \includegraphics[width=0.75\textwidth]{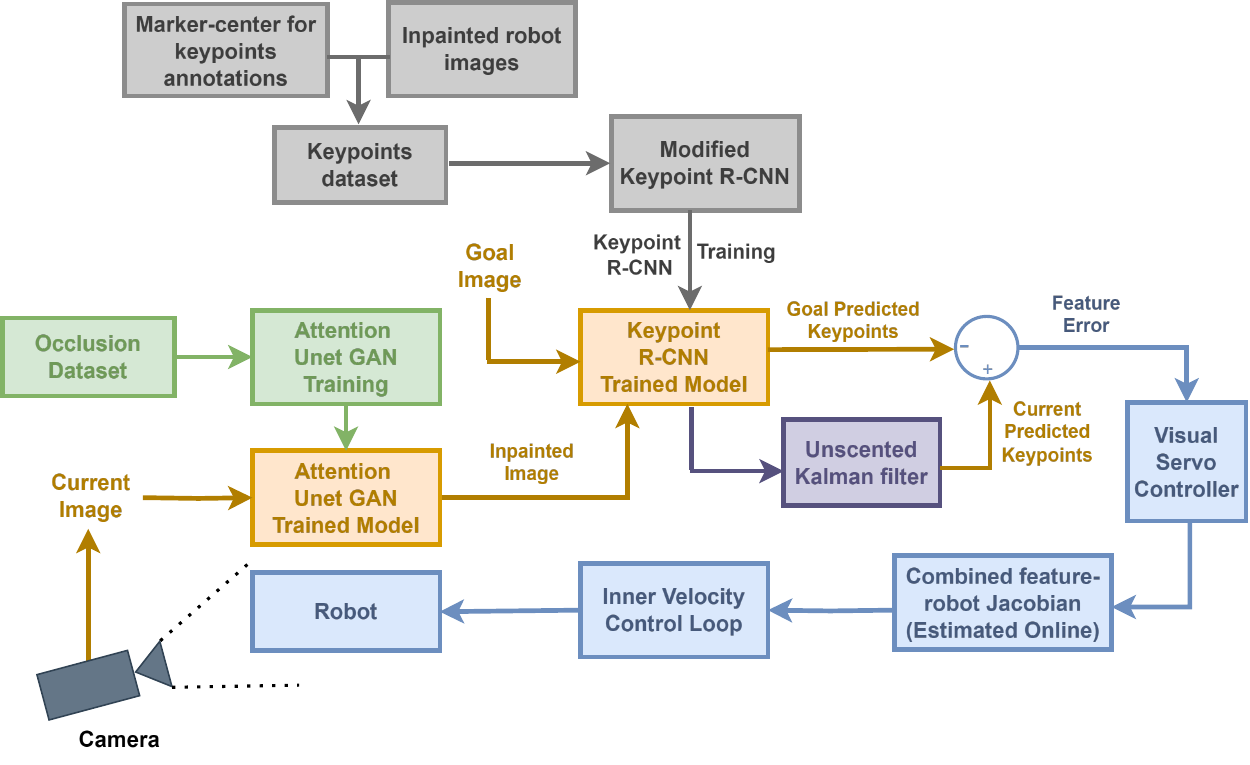}
    \caption{
    Overall pipeline for adaptive visual servoing using predicted keypoints from inpainted images.
    }
    \label{fig:control_flow}
\end{figure*}

The following subsections detail the methodology of our overall pipeline. We begin with an automated data collection process to identify and annotate keypoints along the robot's body. We attach ArUco markers on the robot's body and record their centers as keypoints while capturing images of the robot across various configurations within its visible workspace as described in \autoref{ssec:kp_dataset}. To obtain clean, marker-free training data, we employ a GAN-based inpainting model that removes the ArUco markers and reconstructs the occluded robot regions, detailed in \autoref{ssec:data_coll_img_recon}. The resulting dataset \autoref{ssec:data_gen}, comprised of the annotated keypoint locations and corresponding inpainted images is then used to train a modified keypoint detection model to infer keypoints in the robot's body in run-time, described in \autoref{ssec:kp_pred}. To handle occlusion in the environment during runtime, we first apply a mask-free, attention-based U-Net GAN inpainting model, defined in \autoref{ssec:gan-desc}, to reconstruct occluded regions of the robot in the image. Keypoint detection, similar to \autoref{ssec:kp_pred}, is then performed on the inpainted image. The resulting detections are subsequently passed through an Unscented Kalman Filter (UKF) defined in \autoref{ssec:ukf}, which refines the keypoint trajectories over time. In cases where the keypoint detector fails due to missing or inaccurate reconstructions, the UKF falls back on its internal motion model to predict the keypoint location, enabling robust tracking even under partial or faulty observations. We design an adaptive visual servoing scheme in \autoref{ssec:adaptive_vs} that uses the detected keypoints as control features during runtime.

\begin{figure*}[ht]
  \centering
  % -------- Column Headers --------
  \begin{subfigure}{0.16\textwidth}
    \centering \textbf{Input}
  \end{subfigure}
  \begin{subfigure}{0.16\textwidth}
    \centering \textbf{Label}
  \end{subfigure}
  \begin{subfigure}{0.16\textwidth}
    \centering \textbf{Proposed (AttU-Net)}
  \end{subfigure}
  \begin{subfigure}{0.16\textwidth}
    \centering \textbf{ResU-Net}
  \end{subfigure}
  \begin{subfigure}{0.16\textwidth}
    \centering \textbf{U2NET}
  \end{subfigure}
  \begin{subfigure}{0.16\textwidth}
    \centering \textbf{CoModGAN}
  \end{subfigure}

  \vspace{6pt}

  % -------- Row 1 --------
  \begin{subfigure}{0.16\textwidth}
    \includegraphics[width=\linewidth]{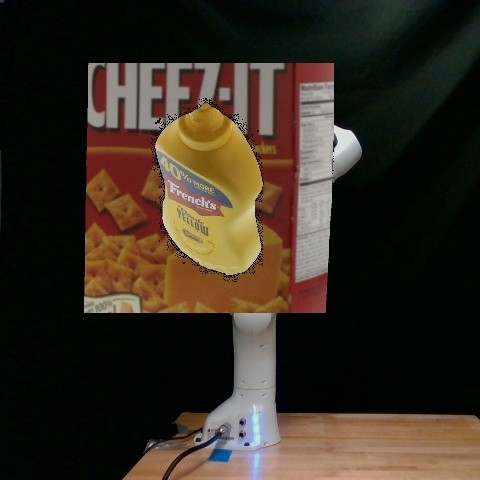}
  \end{subfigure}
  \begin{subfigure}{0.16\textwidth}
    \includegraphics[width=\linewidth]{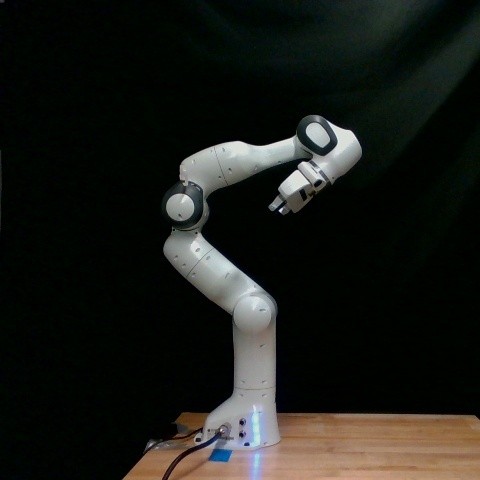}
  \end{subfigure}
  \begin{subfigure}{0.16\textwidth}
    \includegraphics[width=\linewidth]{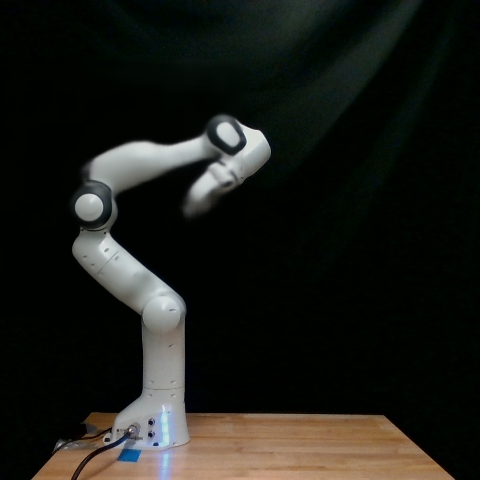}
  \end{subfigure}
  \begin{subfigure}{0.16\textwidth}
    \includegraphics[width=\linewidth]{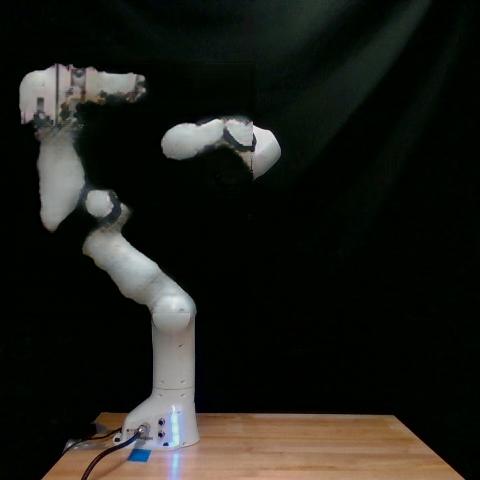}
  \end{subfigure}
  \begin{subfigure}{0.16\textwidth}
    \includegraphics[width=\linewidth]{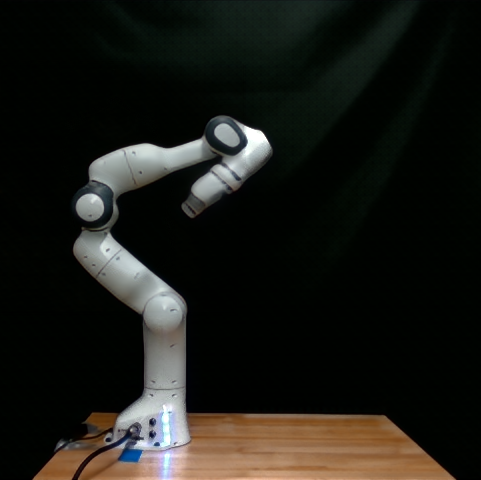}
  \end{subfigure}
  \begin{subfigure}{0.16\textwidth}
    \includegraphics[width=\linewidth]{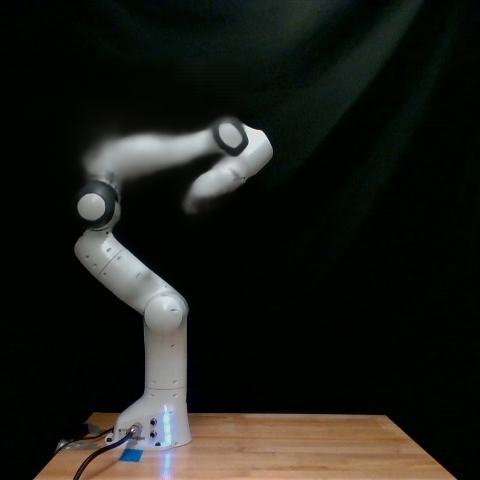}
  \end{subfigure}

  \vspace{8pt}

  % -------- Row 2 --------
  \begin{subfigure}{0.16\textwidth}
    \includegraphics[width=\linewidth]{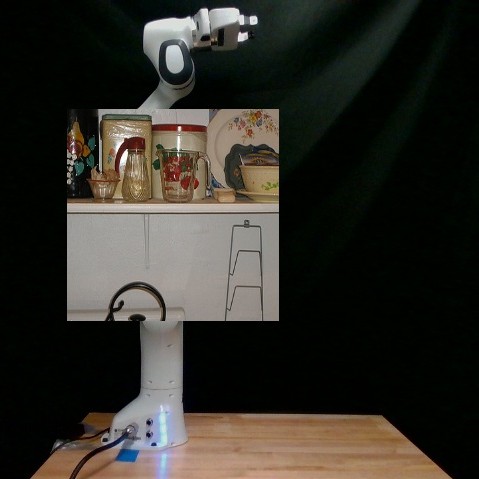}
  \end{subfigure}
  \begin{subfigure}{0.16\textwidth}
    \includegraphics[width=\linewidth]{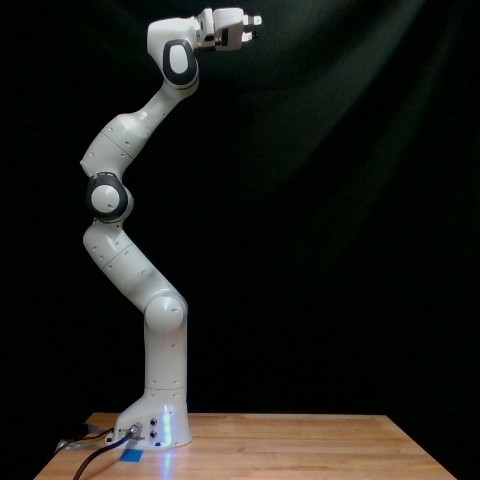}
  \end{subfigure}
  \begin{subfigure}{0.16\textwidth}
    \includegraphics[width=\linewidth]{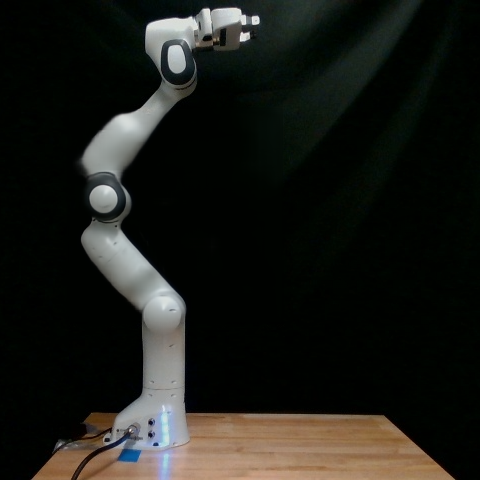}
  \end{subfigure}
  \begin{subfigure}{0.16\textwidth}
    \includegraphics[width=\linewidth]{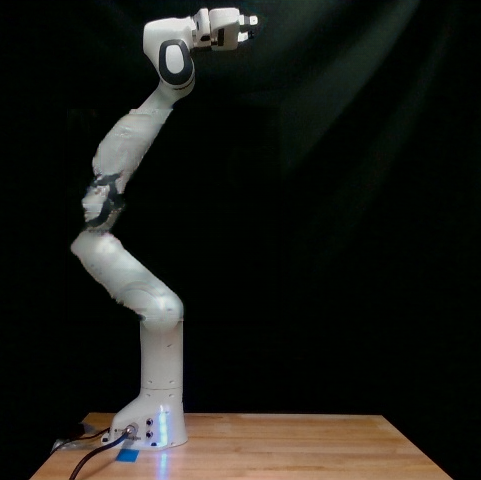}
  \end{subfigure}
  \begin{subfigure}{0.16\textwidth}
    \includegraphics[width=\linewidth]{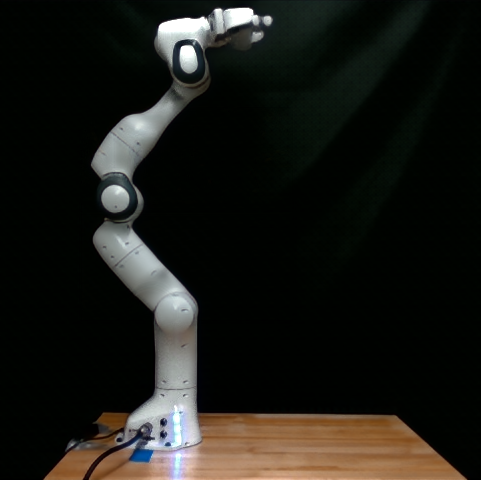}
  \end{subfigure}
  \begin{subfigure}{0.16\textwidth}
    \includegraphics[width=\linewidth]{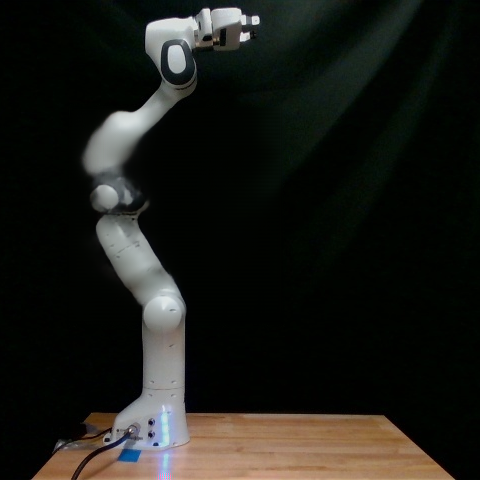}
  \end{subfigure}

  \vspace{8pt}

  % -------- Row 3 --------
  \begin{subfigure}{0.16\textwidth}
    \includegraphics[width=\linewidth]{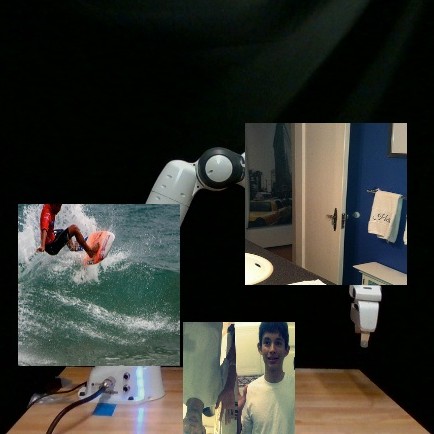}
  \end{subfigure}
  \begin{subfigure}{0.16\textwidth}
    \includegraphics[width=\linewidth]{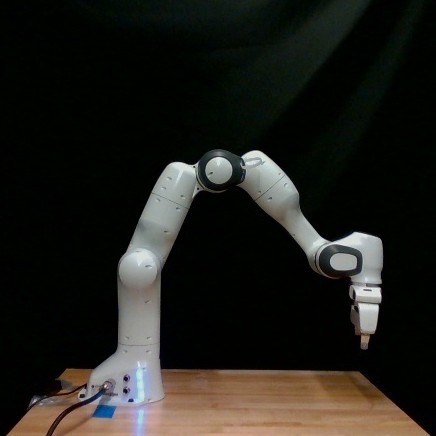}
  \end{subfigure}
  \begin{subfigure}{0.16\textwidth}
    \includegraphics[width=\linewidth]{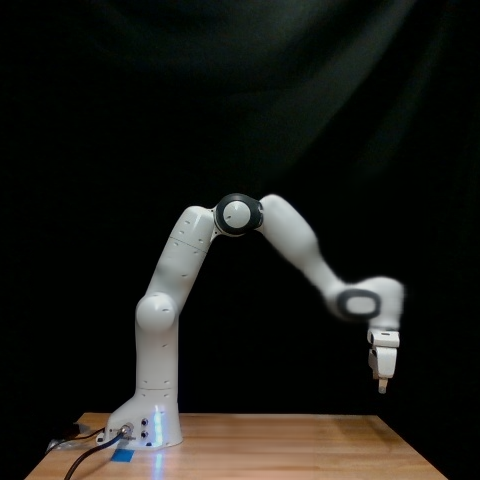}
  \end{subfigure}
  \begin{subfigure}{0.16\textwidth}
    \includegraphics[width=\linewidth]{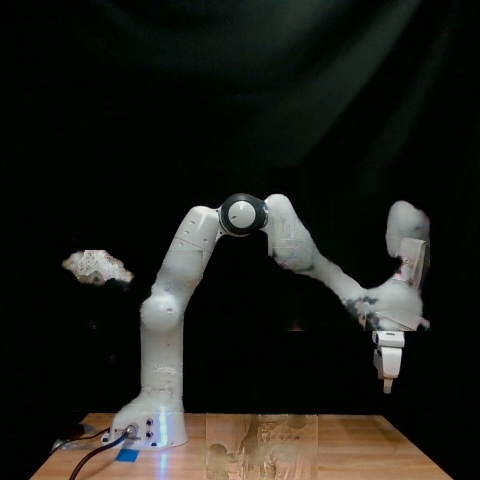}
  \end{subfigure}
  \begin{subfigure}{0.16\textwidth}
    \includegraphics[width=\linewidth]{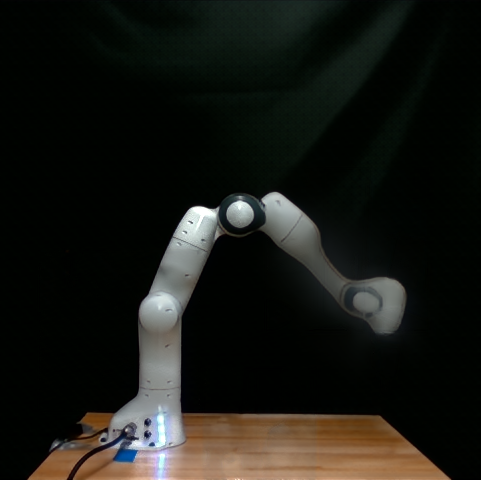}
  \end{subfigure}
  \begin{subfigure}{0.16\textwidth}
    \includegraphics[width=\linewidth]{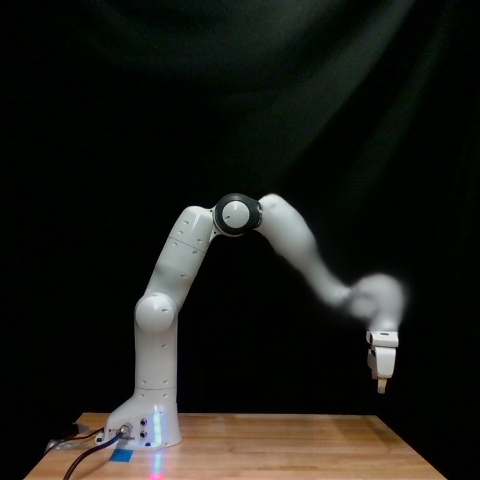}
  \end{subfigure}

  \vspace{8pt}

  % -------- Row 4 --------
  \begin{subfigure}{0.16\textwidth}
    \includegraphics[width=\linewidth]{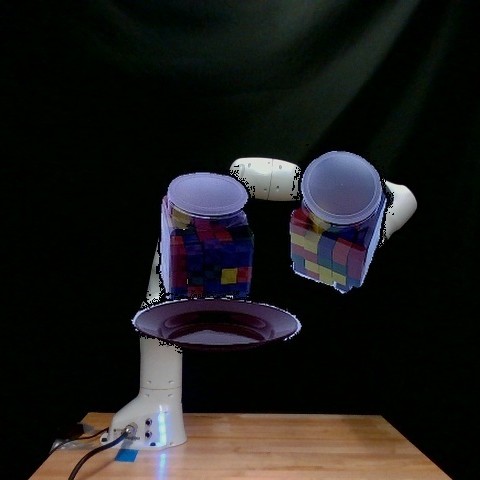}
  \end{subfigure}
  \begin{subfigure}{0.16\textwidth}
    \includegraphics[width=\linewidth]{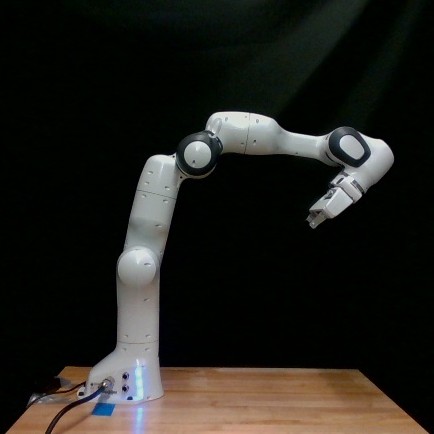}
  \end{subfigure}
  \begin{subfigure}{0.16\textwidth}
    \includegraphics[width=\linewidth]{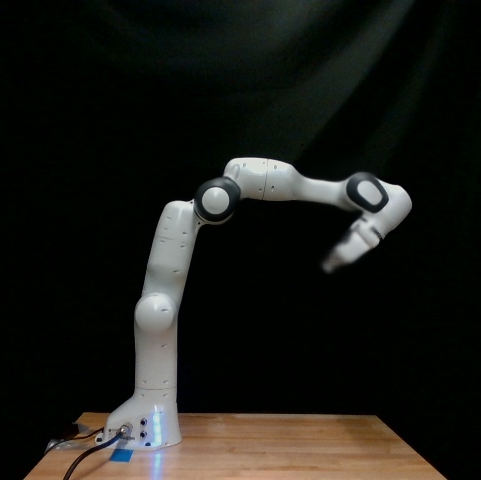}
  \end{subfigure}
  \begin{subfigure}{0.16\textwidth}
    \includegraphics[width=\linewidth]{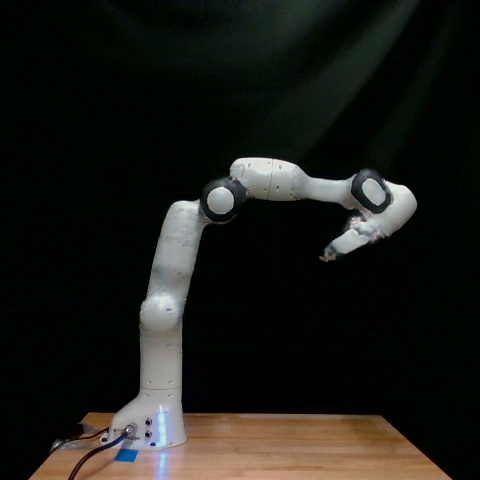}
  \end{subfigure}
  \begin{subfigure}{0.16\textwidth}
    \includegraphics[width=\linewidth]{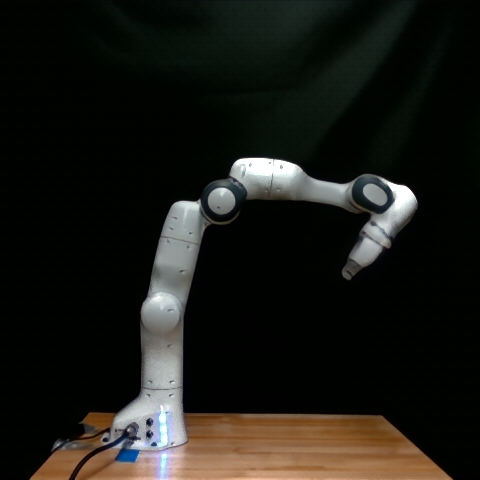}
  \end{subfigure}
  \begin{subfigure}{0.16\textwidth}
    \includegraphics[width=\linewidth]{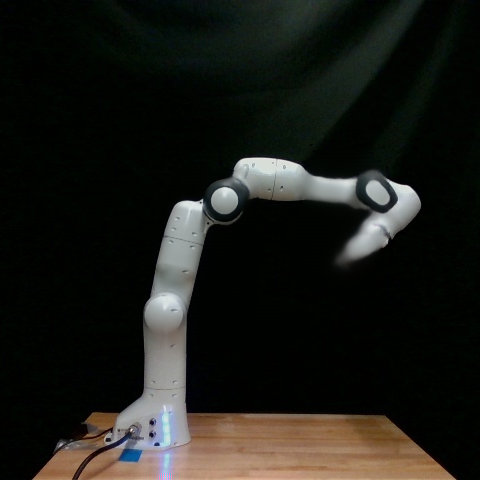}
  \end{subfigure}

  \vspace{8pt}

  % -------- Row 5 --------
  \begin{subfigure}{0.16\textwidth}
    \includegraphics[width=\linewidth]{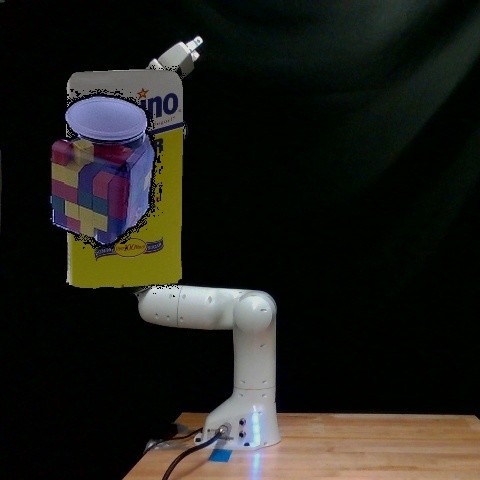}
  \end{subfigure}
  \begin{subfigure}{0.16\textwidth}
    \includegraphics[width=\linewidth]{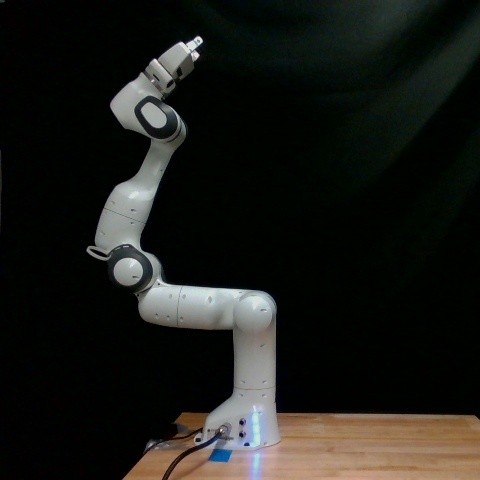}
  \end{subfigure}
  \begin{subfigure}{0.16\textwidth}
    \includegraphics[width=\linewidth]{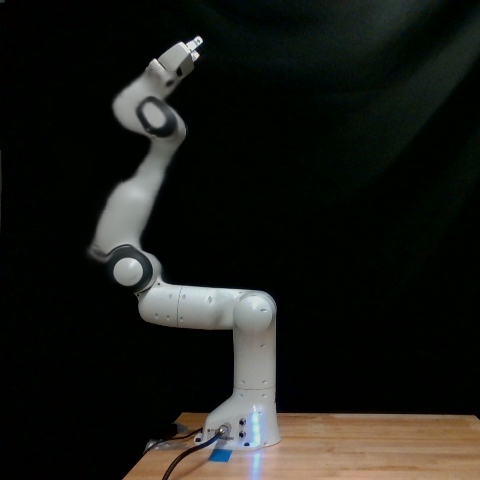}
  \end{subfigure}
  \begin{subfigure}{0.16\textwidth}
    \includegraphics[width=\linewidth]{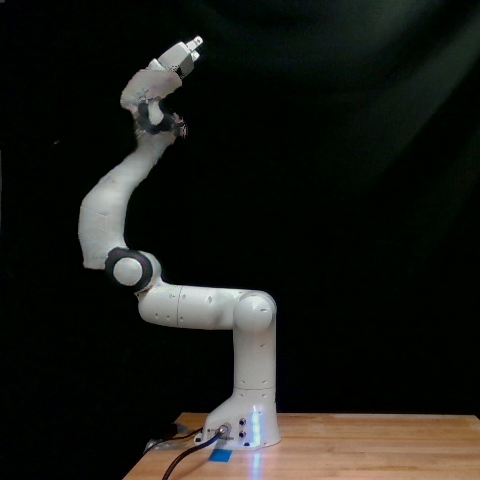}
  \end{subfigure}
  \begin{subfigure}{0.16\textwidth}
    \includegraphics[width=\linewidth]{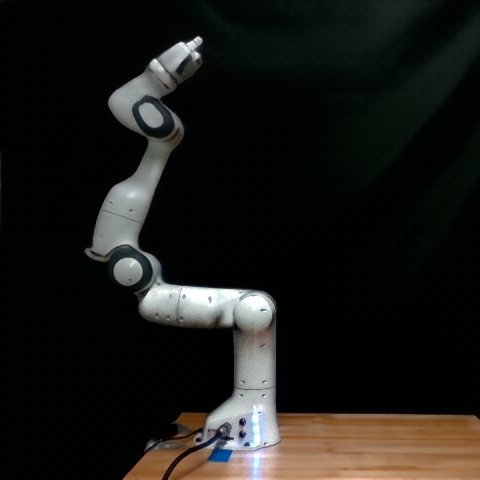}
  \end{subfigure}
  \begin{subfigure}{0.16\textwidth}
    \includegraphics[width=\linewidth]{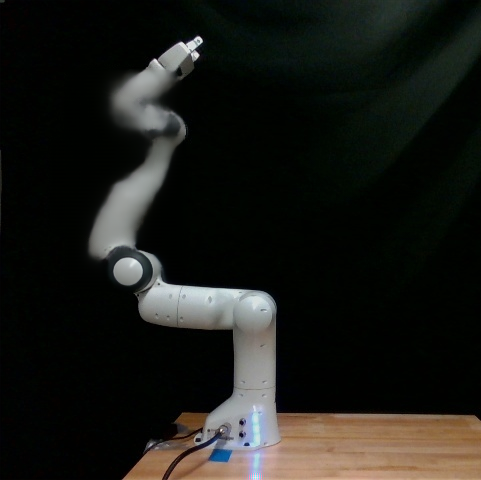}
  \end{subfigure}

  \vspace{8pt}

  % -------- Row 6 --------
  \begin{subfigure}{0.16\textwidth}
    \includegraphics[width=\linewidth]{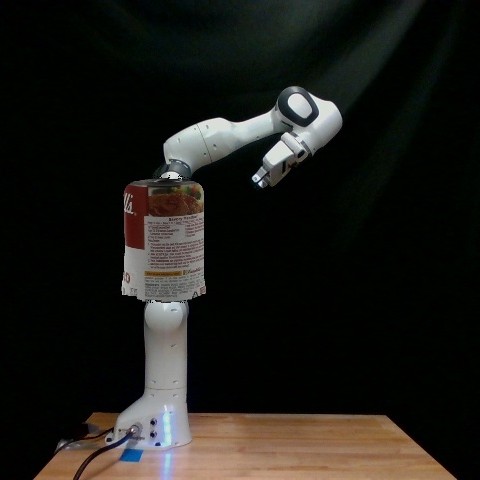}
  \end{subfigure}
  \begin{subfigure}{0.16\textwidth}
    \includegraphics[width=\linewidth]{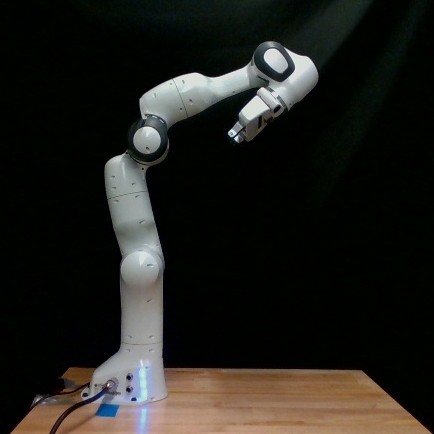}
  \end{subfigure}
  \begin{subfigure}{0.16\textwidth}
    \includegraphics[width=\linewidth]{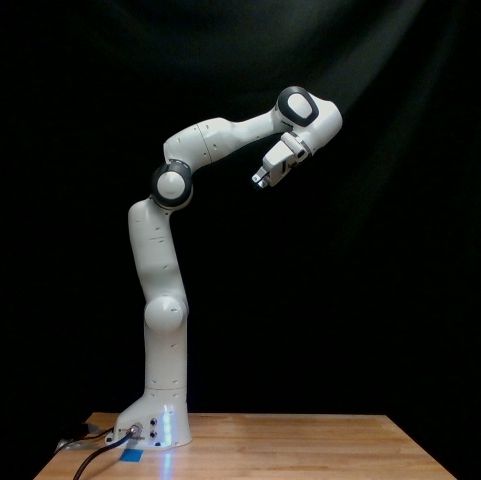}
  \end{subfigure}
  \begin{subfigure}{0.16\textwidth}
    \includegraphics[width=\linewidth]{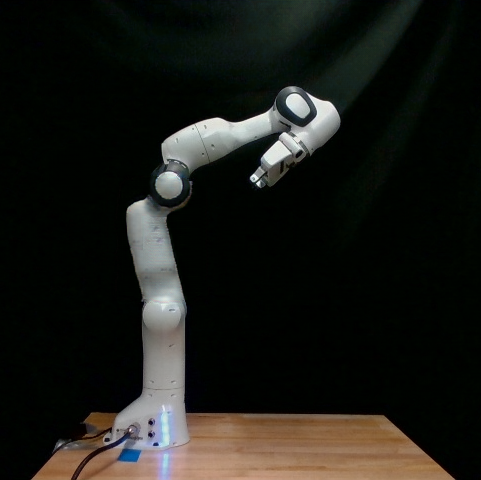}
  \end{subfigure}
  \begin{subfigure}{0.16\textwidth}
    \includegraphics[width=\linewidth]{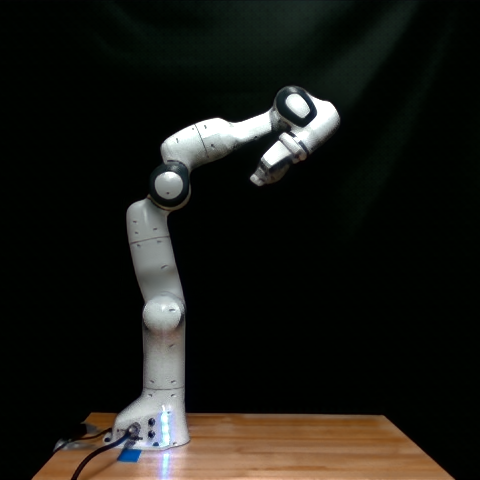}
  \end{subfigure}
  \begin{subfigure}{0.16\textwidth}
    \includegraphics[width=\linewidth]{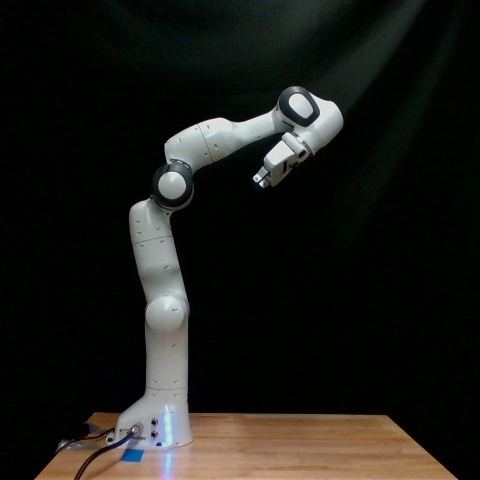}
  \end{subfigure}
  \caption{Comparison of inpainting results across multiple models. Each row shows an input image, its ground truth label, and outputs from AttU-Net, ResU-Net, U2NET, and CoModGAN. Column headers are shown only once for clarity.}
  \label{fig:inpainting_grid}
\end{figure*}
  
\begin{figure*}[!t]
    \centering
    \includegraphics[width=1.0\textwidth, center, trim=0 0 0 0, clip]{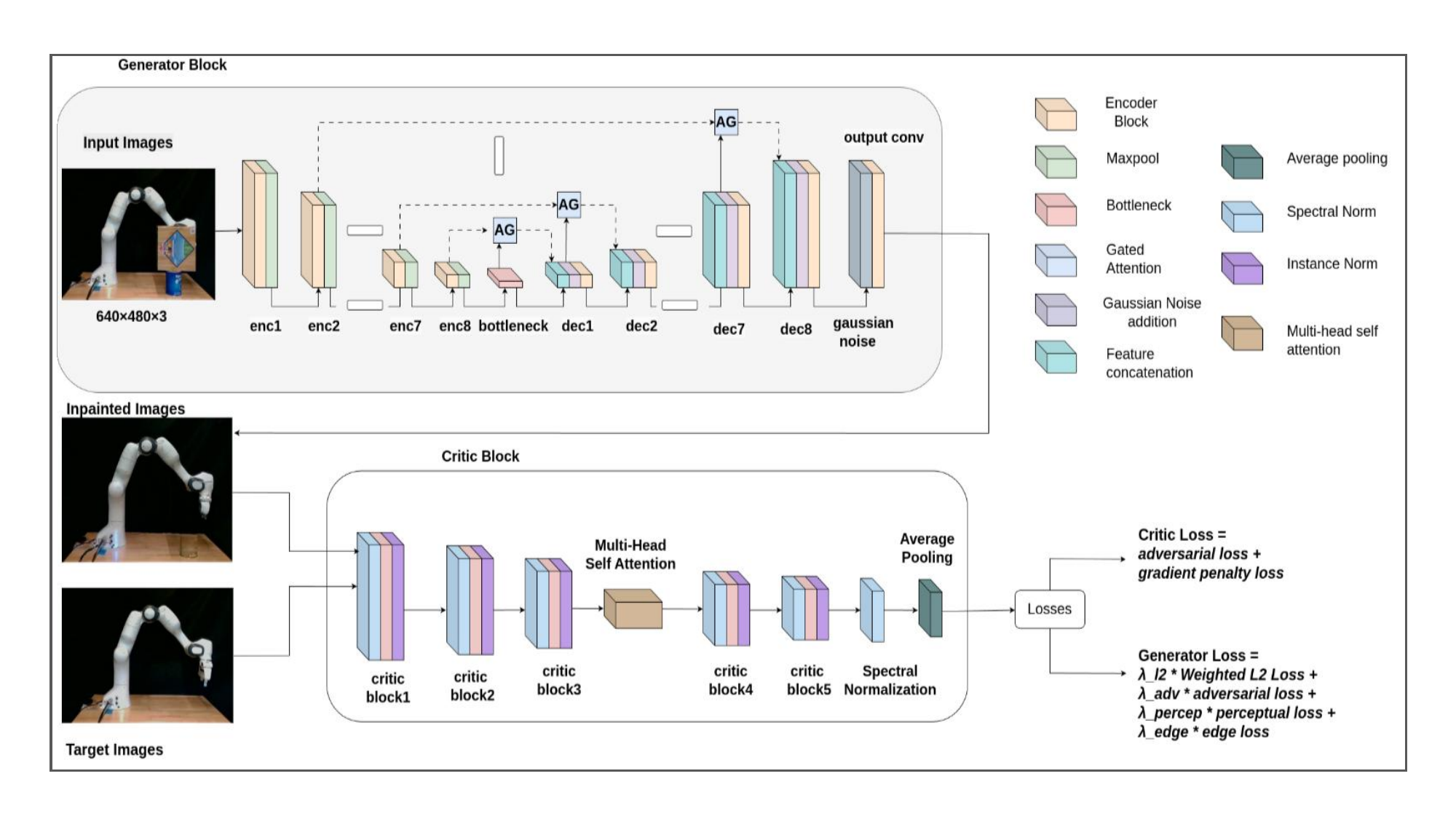}
    \caption{
    Overview of the Attention U-Net based GAN model.
    }
    \label{fig:net-arch}
\end{figure*}

\begin{figure}[!htb]
    \centering
    \includegraphics[width=1.05\linewidth, center, trim=0 0 0 0, clip]{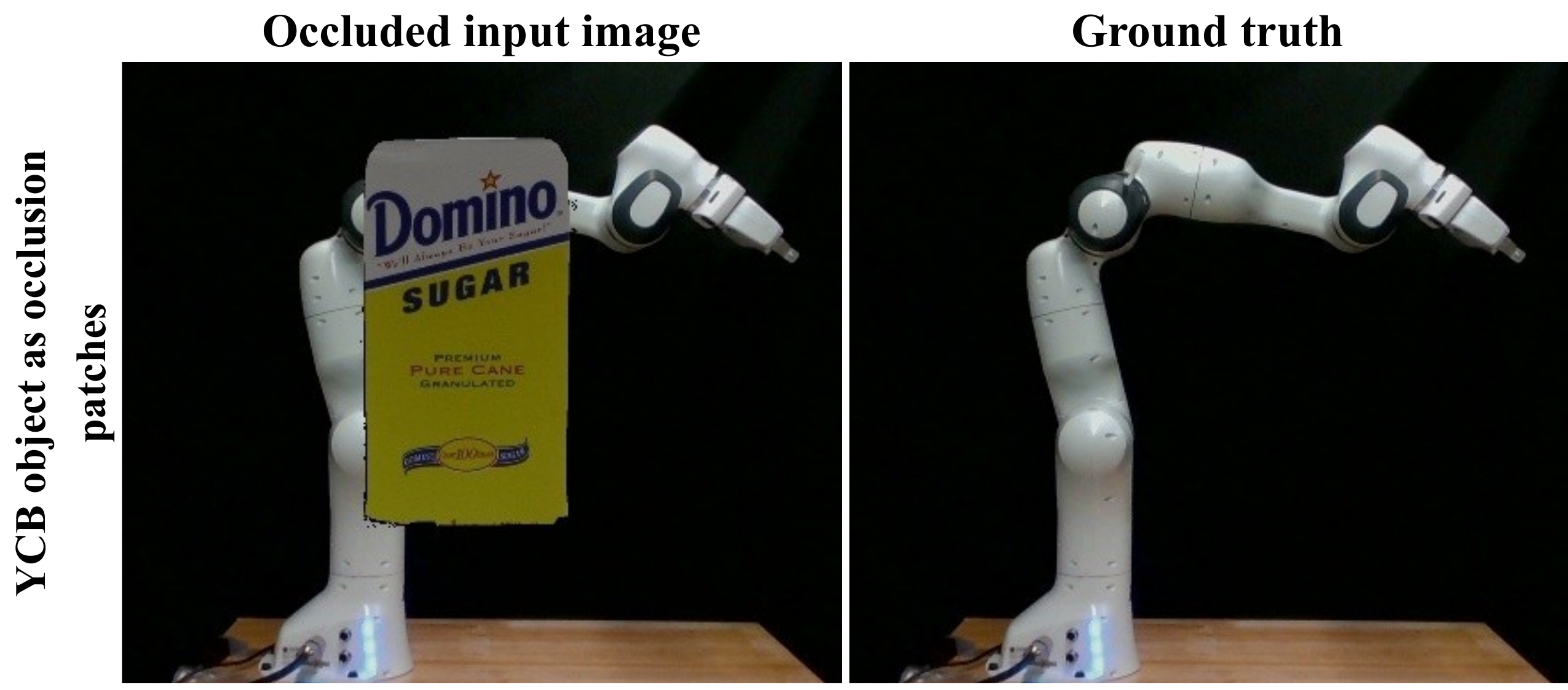}
    
    \includegraphics[width=1.05\linewidth, center, trim=0 0 0 0, clip]{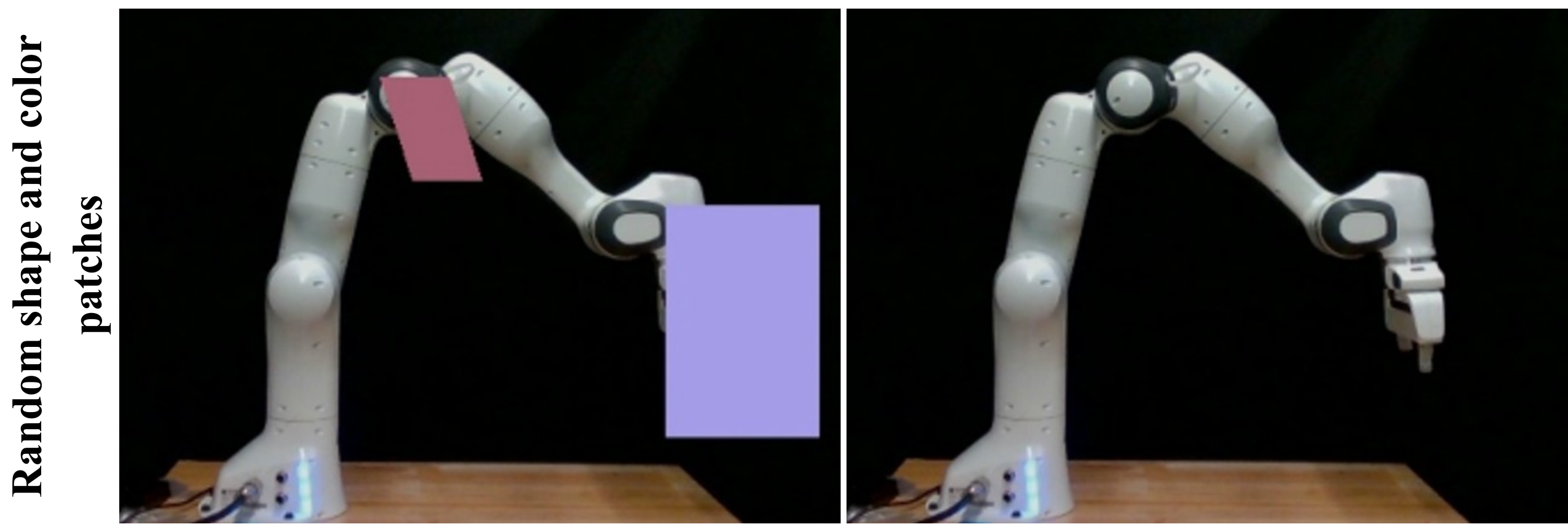}
    
     \includegraphics[width=1.05\linewidth, center, trim=0 0 0 0, clip]{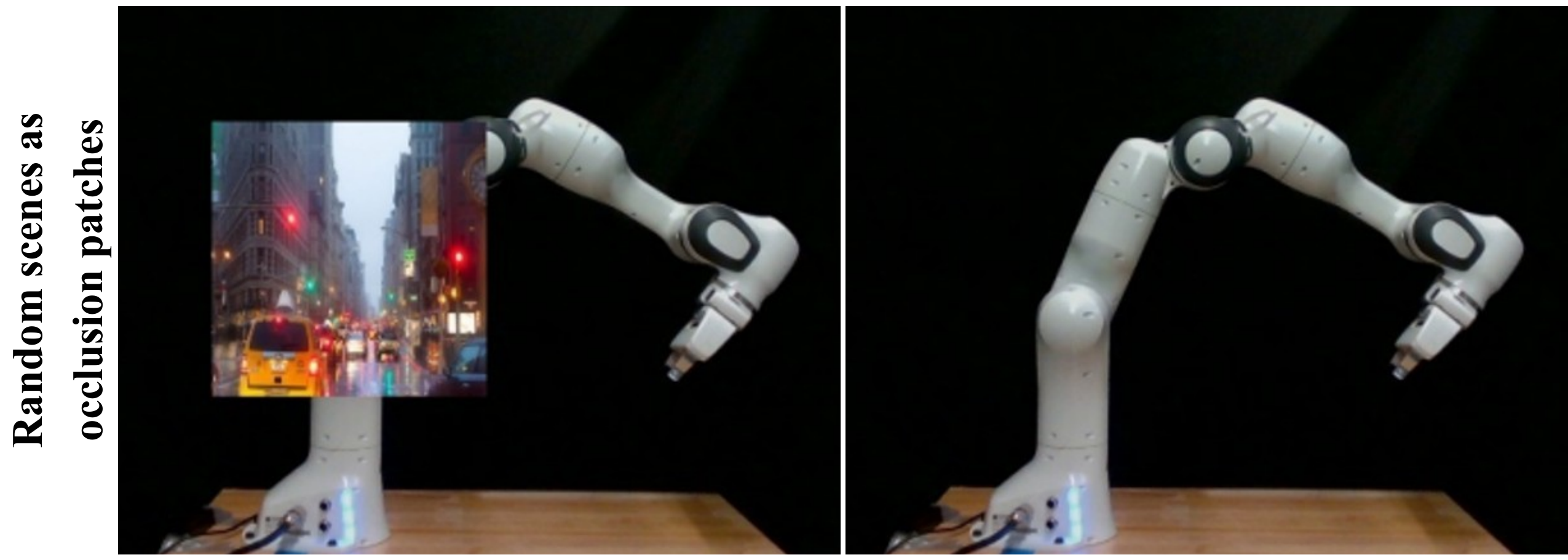}     
    \caption{
   Samples from the dataset used to train the Attention U-Net GAN model. The first row shows occlusions created by overlaying patches of YCB objects onto the robot images. The second row depicts occlusions simulated using randomly shaped and colored patches. The third row presents occlusions generated from real-world scene fragments, enhancing the model’s robustness to diverse visual clutter.
    }
    \label{fig:sample}
    \vspace{-0.5em}
\end{figure}

\subsection{Automated Data collection pipeline}  
\label{ssec:kp_dataset}

Our data collection process is described in detail in the following subsections, with an overview of the entire pipeline illustrated in the flowchart shown in \autoref{fig:data_flow}.

\subsubsection{Training Phase of Inpainting Algorithm:} 
\label{ssec:data_collection}
We collected training images in two phases. \textbf{Phase I (planar)}: we moved the robot within its visible workspace by applying small velocities to the planar joints, computed by \autoref{comp_vel} following \citep{chatterjee2025image}, and captured images at a fixed frequency. \textbf{Phase II (out-of-plane)}: we applied the same velocity law to all planar joints and additionally to the first spatial joint to induce out-of-plane motion, again recording images at the same capture rate. 

\begin{equation}
\small
\label{comp_vel}
v_j = \min\left(\frac{M_j}{\ensuremath{res} \cdot dt}, v_{\text{max}}\right),
\end{equation}

where, \( M_j \) is the motion range or difference of limits of joint \( j \), \ensuremath{res}, is the number of discrete  steps used to traverse \( M_j \), \( dt \) is the duration allocated to complete each step, and \( v_{\text{max}} \) is the maximum allowable velocity for joint \( j \).
By dividing the motion range of each joint into uniform increments based on \ensuremath{res}, the robot is guided to systematically traverse its configuration space, ensuring comprehensive coverage of its reachable workspace. We capture a set of images while moving the robot using the above process. The training images are archived on Zenodo \textit{(DOI: 10.5281/zenodo.17309054)}

We use this dataset to train a state-of-the-art mask-based GAN inpainting model LaMa \citep{Suvorov_2022_WACV}. Our primary motivation to choose LaMa is that it surpasses the majority baseline models including other state-of-the-art solutions such as Deepfill\_v2 \citep{yu2019free}, EdgeConnect \citep{nazeri2019edgeconnect}, HiFill \citep{yi2020contextual}, MADF \citep{zhu2021image} and CoModGan \citep{zhao2021large}, according to evaluations using the Learned Perceptual Image Patch similarity (LPIPS) and Frechet Inception Distance (FID) metrics. These metrics were applied to assess performance across various test mask generation strategies, such as narrow, wide and segmentation masks. It is not only robust to different image resolutions but also easy to customize, due to clear, well-documented configuration files, making it flexible for inpainting a variety of data. 

LaMa \citep{Suvorov_2022_WACV} provides a number of pre-trained models in their GitHub repository. However, since they were not trained with robot images, we fine tune the best pre-trained model named as `big-lama' in the repository by further training it with our collected robot images. We preprocess the collected
images following the instructions in \citet{Suvorov_2022_WACV}. An important aspect of preprocessing the data is to crop it to a suitable size. For our purpose, we cropped the images to $480 \times 480$. During training run time, by default, LaMa crops the images further to a size of $256 \times 256$, which can be customized by changing specific parameters inside the LaMa repository. We changed it to $480 \times 480$ for our datasets. For further details on data preprocessing, refer to our data collection page at https://github.com/Jani-C/KPDataGenerator. We train the customized model for $80$ epochs of batch size $30$. The part of the workflow in \autoref{fig:data_flow} in grey represents this phase.

\subsubsection{Image Reconstruction using LaMa:} 
\label{ssec:data_coll_img_recon}

We first place ArUco markers along the robot's body as shown in \autoref{fig:intro-image}(a) and \autoref{fig:3d_wkflow}(a). Data are collected in two phases, mirroring \autoref{ssec:data_collection}. \textbf{Phase I (planar):} we apply small joint velocities computed with \autoref{comp_vel} to all planar joints while continuously capturing images across the reachable workspace. \textbf{Phase II (out-of-plane):} we use the same velocity law on all planar joints and additionally actuate the first spatial joint to induce out-of-plane motion, again recording images at the same capture rate. We would like to emphasize that, in our proposed method, the user has the flexibility to place the markers anywhere according to their individual purposes. As shown in \autoref{fig:intro-image}(a), we use $5$ markers for planar-motion data collection, and as shown in \autoref{fig:3d_wkflow}(a), we use $8$ markers distributed along the entire length of the robot for 3D-motion data collection. One thing worth noticing is the size of the markers for both cases. For 3D data collection we employ larger ArUco markers, since increased camera–robot distance and oblique viewpoints reduce the markers’ apparent size and contrast, making reliable detection more difficult. Larger markers provide robust detection across depth variation. 

Once the markers are placed in the desired locations on the manipulator and the images are collected, we automatically detect the marker locations in each image and create a binary mask around the markers as depicted in \autoref{fig:intro-image}(b) and \autoref{fig:3d_wkflow}(b). The RGB images and their corresponding masks are then inpainted by the newly trained LaMa model (archived on Zenodo as \textit{DOI: 10.5281/zenodo.17334695}), with refinement enabled to improve the inpainting quality as described in \citet{kulshreshtha2022feature}. \autoref{fig:intro-image}(c) and \autoref{fig:3d_wkflow}(c) show the LaMa prediction, in which the ArUco markers are removed. The centers of the ArUco markers in the original images provide our keypoint labels for both planar and out-of-plane views. The green portion of \autoref{fig:data_flow} represents this phase. The images of the arm with ArUco markers on it are archived on Zenodo \textit{(DOI: 10.5281/zenodo.17333705)}. The combined dataset of images with markers and corresponding binary masks are also archived on Zenodo \textit{(DOI: 10.5281/zenodo.17308981)}.

\subsubsection{Dataset Generation:}
\label{ssec:data_gen}
We save the center coordinates of the ArUco markers on all images collected in \autoref{ssec:data_coll_img_recon} in JSON files as keypoint annotations, together with the corresponding reconstructed images. The `keypoints' field in the JSON file is of the format [$x$, $y$, $v$], where $v$ is the visibility of the keypoint and $x$ and $y$ are the pixel coordinates of the keypoint in image space. Along with the keypoints, we also compute a bounding box with each keypoint as the center and save those in the same JSON file. \textbf{For planar-motion images}, each of these `bboxes' (bounding boxes), is a fixed square of side $bb\_size$ centered at \((x,y)\), and its orientation follows the marker’s detected rotation (via the ArUco corner detection). \textbf{For 3D-motion images}, instead of a fixed size we use the rectangle induced by the detected ArUco marker corners (i.e., the tight axis-aligned box enclosing the four corners), and store those coordinates in the same field. So the `bboxes' field has the format [$x-(bb\_size$$/$$2$), $y-(bb\_size$$/$$2$), $x+(bb\_size$$/$$2$), $y+(bb\_size$$/$$2$)] for planar images (fixed-size square) with $bb\_size$ as the side length and for 3D-motion images, `bboxes' store [$x\_{\min}$, $y\_{\min}$, $x\_{\max}$, $y\_{\max}$] computed directly from the detected marker corners. 

This constitutes the training dataset for our customized keypoint detector network described in \autoref{ssec:kp_pred}. We have $5$ keypoints (for planar motion) and $8$ keypoints (for out-of-plane motion) for our Panda robot along with their corresponding bounding boxes. \autoref{fig:intro-image}(d) and \autoref{fig:3d_wkflow}(d) illustrate images of sample robot configurations labeled with keypoints, bounding boxes, and respective keypoint names, respectively for planar and out-of-plane motions. The orange part of the workflow in \autoref{fig:data_flow} represents this phase of the methodology.

The reconstructed images with annotated json files are archived \textit{(DOI: 10.5281/zenodo.17309869)}.

\subsection{Network Architecture of the Keypoint Detector Model:}
\label{ssec:kp_train}
We use PyTorch vision library's \citep{ayyadevara2020modern} \textit{keypointrcnn\_resnet50\_fpn} to train our dataset generated in \autoref{ssec:data_gen}. The \textit{keypointrcnn\_resnet50\_fpn}, which is pretrained to detect $17$ keypoints in the human body, is an extension of the Mask R-CNN \citep{he2017mask} deep learning model and is state of the art for keypoint detection for human body pose estimation. We customized this model \citep{P2021} for the datasets that we created with $5$ keypoints (planar) and $8$ keypoints (3D) for the Panda robot. In our custom PyTorch Dataset, keypoints/bounding boxes are labeled sequentially: for planar, $1$ to $5$ (i.e., \textit{kp\_1} to \textit{kp\_5}) as shown in \autoref{fig:intro-image}(d); for 3D, $1$ to $8$ (i.e., \textit{kp\_1} to \textit{kp\_8}) as shown in \autoref{fig:3d_wkflow}(d).

The \textit{keypointrcnn\_resnet50\_fpn} uses the backbone of ResNet50 with pre-trained weights of IMAGENET1K\_V2 from the COCO Dataset \citep{Lin2014}. We refined our model on our datasets using an optimizer and hyperparameter configuration selected via Optuna hyperparameter optimizer \citep{akiba2019optuna}. The tuning process identified Stochastic Gradient Descent (SGD) as the optimal optimizer, with a learning rate of $0.0001$, momentum of $0.9$, and a weight decay of $0.0005$ to mitigate overfitting. Training was conducted using a batch size of $1$ over $50$ epochs for the planar model and $2$ over $50$ epochs for the 3D model. The evaluation and prediction of keypoints follow the same procedure as in our previous work \citep{chatterjee2023keypoints}. The grey part of the workflow illustrated in \autoref{fig:control_flow} represents keypoint detector training phase of the pipeline.

\subsection{Mask-free Inpainting Model for Handling Occlusion}
\label{ssec:gan-desc}
This section details our mask-free GAN inpainting approach to reconstruct occlusion-free robot images. This part of the pipeline is implemented and tested only for planar motion, hence, only planar datasets are used.
The three primary parts of our approach are:
\begin{itemize}
    \item an Attention U-Net-based generator architecture,
    \item a discriminator architecture that uses spectral normalization, and
    \item a training strategy that uses Wasserstein loss with gradient penalty.
\end{itemize}

The green part of the workflow in \autoref{fig:control_flow} depicts this phase of the control pipeline.

We first present our methodological rationalization, followed by detailed descriptions of each component.

\subsubsection{Methodological Rationale:}
We choose our proposed GAN architecture for mask-free inpainting as it offers several key advantages. 
\begin{itemize}
    \item GANs do not require explicit modeling of complex occlusions as they learn them through adversarial training. 
    \item The generator's encoder-decoder architecture with attention-based skip connections helps preserve fine details of the robot's body such as the edges of the links or the joint regions, which are important for accurate keypoint detection. This is reinforced through adversarial training, improving overall control accuracy.
    \item Our GAN-based method can detect and reconstruct occluded regions directly, without needing extra post-processing or explicit masks like traditional inpainting methods.
\end{itemize}

We also ran preliminary studies with other methods like Transformers, Swin Transformers, LaMa, and diffusion models, but they demonstrated major limitations for our application as listed below:

\begin{itemize}
    \item Transformer-based models introduced block-like artifacts at patch boundaries, which reduced the accuracy of keypoint detection.
    \item  Swin Transformers were too computationally heavy for real-time use and also produced artifacts in mask-free settings.
    \item LaMa relies on explicit binary masks, which goes against our goal of avoiding mask-based supervision.
    \item Diffusion models are unsuitable for real-time applications due to their slow sampling process and the difficulty of modeling occlusions across many small steps.
\end{itemize}

\autoref{fig:inpainting_grid} illustrates a qualitative analysis of the accuracy of image reconstruction across tested methods. As demonstrated for robotic visual servoing, our GAN model with an Attention U-Net generator and WGAN-GP training offers the best balance of image quality, speed, and mask-free operation.

\autoref{fig:net-arch} illustrates the overview of the network architecture of our proposed Attention U-Net based GAN model.

\subsubsection{Data Preparation and Preprocessing:}
Our dataset for GAN consists of $80,000$ image pairs collected using the real Franka Emika Panda robot. Each pair includes an image of the robot with synthetically added occlusions and its corresponding unoccluded version. The occlusions are generated using a combination of YCB objects, randomly placed geometric shapes, and real-life scenes as occlusion patches to mimic real-world visual obstructions. The clean, unoccluded images serve as ground truth targets for training the network, allowing it to learn how to reconstruct the robot's appearance under various occlusion conditions. \autoref{fig:sample} illustrates examples from the dataset, showing different types of occlusion patches used to generate the training data. The processed dataset as training input is archived on Zenodo \textit{(DOI: 10.5281/zenodo.17310309)}.

\subsubsection{Generator Architecture - Attention U-Net:}
The generator uses an Attention U-Net architecture with gated skip connections to turn occluded regions into clear, unoccluded ones. Since we do not use occlusion masks, this design helps the network learn to tell apart parts of the robot that should be kept from occlusions that should be removed. This is especially important for accurately reconstructing detailed robot structures by focusing on important areas and ignoring irrelevant ones.

Unlike traditional U-Net implementations, our generator employs a deeper encoder-decoder structure with eight stages of feature extraction and reconstruction. This depth makes it possible to perform a thorough multiscale analysis to counteract the maskless application. 

The encoder consists of eight convolutional blocks that progressively reduce the image size while increasing the feature channels. Each block includes two $3\times3$ convolution layers, followed by batch normalization, ReLU activation, and $2\times2$ max pooling. The decoder mirrors the encoder with eight blocks that restore the image size while reducing feature channels. A key feature of our design is the gated attention mechanism in each skip connection, which selectively filters the information passed from the encoder to the decoder.

Each gated attention block in the skip connections uses information from the decoder to compute an attention score, which controls how much of the encoder’s features should be used for reconstruction \citep{SCHLEMPER2019197}. For each spatial location $(i,j)$ on the feature map, the gate determines the importance of the features using the following equation

\begin{equation}
\small
    (\alpha(i,j) = \sigma(W^T \cdot [F_{\text{enc}}(i,j); F_{\text{dec}}(i,j)] + b))
\end{equation}

where \(\sigma\) is the sigmoid function, 
\(W\) and \(b\) are learnable parameters, and
\([F_{\text{enc}}; F_{\text{dec}}]\) concatenates encoder and decoder features at position
$(i,j)$ 

The attended feature map that enhances reconstruction fidelity is computed as follows:
\begin{equation}
\small
    F_{\text{att}}(i,j) = \alpha(i,j) \cdot F_{\text{enc}}(i,j),
\end{equation}

A final $1\times1$ convolution followed by a hyperbolic tangent activation generates the RGB output image, normalized to the range \([-1, 1]\).

By suppressing occlusion patterns and preserving the robot's structural details, this method enables selective feature propagation. The attention coefficients serve as soft, learned masks that detect relevant features without needing occlusion masks. 

\subsubsection{Discriminator Architecture - Spectral-Normalized Convolutional Network:}
To provide localized feedback to the generator, the discriminator network uses a convolutional neural network with spectral normalization architecture to assess how realistic the generated images are. It outputs a single scalar score, computed using five downsampling blocks, a padded convolution, an instance normalization, and a final linear layer. We use a scalar output instead of a spatial score map because it simplifies training and helps maintain stability. Spectral normalization ensures Lipschitz continuity, which is important for stabilizing WGAN-GP by keeping gradient norms within a safe range. Unlike PatchGAN, which focuses on small local regions, our discriminator captures global structure. While PatchGAN is good at judging local details, it may miss larger inconsistencies. Our approach provides feedback across the whole image, helping preserve global realism and structural consistency. All convolutional layers are spectrally normalized to control gradient flow. This works alongside the WGAN-GP gradient penalty to stabilize training, prevent mode collapse, and ensure accurate evaluation of important robot features like joints and links. 

\subsubsection{Training Strategy - WGAN-GP:}
We train our model using the Wasserstein GAN with Gradient Penalty (WGAN-GP) framework, which offers more stable and higher-quality results than standard GANs. The loss function is defined as:
\begin{equation}
\small
    L = \mathbb{E}[D(\tilde{x})] - \mathbb{E}[D(x)] + \lambda \cdot \mathbb{E}\left[\left(\|\nabla_{\hat{x}} D(\hat{x})\|_2 - 1\right)^2\right]
\end{equation}

where
\begin{itemize}
    \item \(\tilde{x}\) is the generator’s output (inpainted image),
    \item \(x\) is the real (occlusion-free) image,
    \item \(\hat{x}\) is an interpolated sample between \(\tilde{x}\) and \(x\) (real and generated samples),
    \item \(\lambda = 10\) is the gradient penalty coefficient, and
    \item \(D\) denotes the discriminator.
\end{itemize} 

The generator minimizes: 

{\small
\[
L_G = -\mathbb{E}[D(G(x_{\text{occluded}}))]
\]
}

% $$(L_G = -\mathbb{E}[D(G(x_{\text{occluded}}))])$$ 

The discriminator minimizes:\\
{\small
\[
L_D = \mathbb{E}[D(G(x_{\text{occluded}}))] - \mathbb{E}[D(x)] + \lambda \cdot \mathbb{E}\left[\left(\|\nabla_{\hat{x}} D(\hat{x})\|_2 - 1\right)^2\right]
\]
}

Wasserstein distance gives smoother and more useful gradients than binary cross-entropy, helping to avoid vanishing gradients. The gradient penalty enforces Lipschitz continuity, making training more stable. We use RMSprop with a learning rate of $0.0001$ and a batch size of $4$, updating the generator once for every five discriminator iterations to keep their convergence rates balanced. 

The final trained model is archived on Zenodo (\textit{DOI: 10.5281/zenodo.17334749}).

\subsection{Keypoint Prediction in Real-time using Trained Keypoint R-CNN:}
\label{ssec:kp_pred}
In this phase we use the trained Keypoint R-CNN model as described in \autoref{ssec:kp_train}. We pass image frames of different robot configuration in runtime. Keypoints along the robot's body are then predicted for different configurations achieved by the robot as illustrated by the sample configuration in \autoref{fig:intro-image}(e). These predicted keypoints are then used as visual control features for our adaptive visual servoing scheme described in \autoref{ssec:adaptive_vs}. If there is occlusion present in the environment then the images are passed through the Attention U-Net-GAN inpainting model trained in \autoref{ssec:gan-desc}. This inpainted image is then passed to the trained Keypoint R-CNN model to predict keypoints under occlusion. The orange part of the pipeline in \autoref{fig:control_flow} covers this area of the methodology.

\subsection{Estimating Missing and Inaccurate Keypoints Using Unscented Kalman Filter}
\label{ssec:ukf}
We use the inpainted images generated using the method described in \autoref{ssec:gan-desc} to detect keypoints with the method described in \autoref{ssec:kp_pred}. 
While most occlusions are resolved through our inpainting-based reconstruction pipeline, certain challenging cases, such as occlusion spanning a wider area, or ambiguous reconstructions, can still lead to missing or inaccurate keypoint detections. To handle these residual complexities and ensure continuity of keypoint tracking, we integrate an Unscented Kalman Filter (UKF) for each visual keypoint.
The UKF maintains a 6-dimensional state vector \([x, y, v_x, v_y, a_x, a_y]\), representing the 2D position, velocity, and acceleration of each keypoint in image space. An important point to be noted, the UKF is an integral part of the pipeline and operates at every timestep, smoothing and predicting keypoint trajectories even when all keypoints are successfully detected.

We use the Van der Merwe scaled sigma points formulation \citep{merwe2004sigma} to compute sigma points and propagate the nonlinear motion dynamics through the filter. The state transition function models constant acceleration motion, updating positions and velocities over a fixed timestep \(\Delta t = 0.1\) seconds using:

\begin{equation}
\begin{aligned}
x' &= x + v_x \cdot \Delta t + 0.5 \cdot a_x \cdot \Delta t^2 \\
y' &= y + v_y \cdot \Delta t + 0.5 \cdot a_y \cdot \Delta t^2 \\
v_x' &= v_x + a_x \cdot \Delta t \\
v_y' &= v_y + a_y \cdot \Delta t \\
a_x' &= a_x \\
a_y' &= a_y
\end{aligned}
\end{equation}

We set \(\Delta t = 0.1\) seconds to match the $10$ Hz control loop, so the UKF runs in sync with the robot’s update rate.
The observation model directly measures the position and velocity of each keypoint, i.e., \([x, y, v_x, v_y]\), which are computed from detected keypoints and their temporal differences across image frames.

The filter uses three tunable covariance matrices: the state uncertainty \(P\), process noise \(Q\), and measurement noise \(R\). These matrices allow the filter to adjust its trust in the predicted state versus incoming measurements.

In our implementation, we use the following values:

\begin{itemize}
    \item \textbf{State covariance P}:
    This diagonal matrix defines the initial uncertainty in each state variable: position \((x, y)\), velocity \((v_x, v_y)\), and acceleration \((a_x, a_y)\). For our experimental setup we specified the diagonal entries as \textbf{\boldmath{\small $[50, 50, 30, 30, 10, 10]$}}. Higher values for position reflect greater initial uncertainty in where keypoints might appear, while relatively lower values for velocity and acceleration reflect a smoother motion assumption.

    \item \textbf{Process noise covariance Q}:
    This matrix represents uncertainty in the motion model. In our implementation, we assume uncorrelated noise across state variables and define $Q$ as a diagonal matrix with entries \textbf{\boldmath{\small $[0.2, 0.2, 1.0, 1.0, 0.2, 0.2]$}}. The higher values for velocity terms indicate allowance for sudden velocity changes, which helps the filter respond to unmodeled accelerations. Lower values for position and acceleration help maintain smoother updates and prevent jitter in predictions.

    \item \textbf{Measurement noise covariance R}:
    This matrix encodes the expected noise in the keypoint detector’s output. In our implementation, we assume the measurements are uncorrelated and represent 
    $R$ as a diagonal matrix with entries \textbf{\boldmath{\small $[0.1, 0.1, 30.0, 30.0]$}}. We assume low noise in position measurements \((x, y)\) due to the precision of the \textit{keypointrcnn\_resnet50\_fpn} on inpainted images. In contrast, velocity measurements are estimated by comparing positions across consecutive frames, which can be noisy. To account for this, we assign much higher uncertainty to \((v_x, v_y)\), reducing their influence on the filter’s update.
\end{itemize}   

These values are chosen based on experiments to make the filter both responsive and stable. This helps the filter keep keypoints accurately tracked while avoiding sudden jumps caused by noise or detection errors.

The purple part in the workflow \autoref{fig:control_flow} depicts the phase where UKF is implemented in the pipeline.

\hypertarget{ukf_thresholding} {%
To handle a missing keypoint in the current frame, we substitute its position with the predicted value from the UKF. To estimate its motion, we use the velocity of the nearest visible keypoint, under the assumption that neighboring points often move similarly, especially on a rigid robotic structure. In cases where a keypoint is detected but shows an unusually large displacement compared to its location in the previous frame, we perform a reliability check. The system compares the UKF's predicted location with the new detection. If the prediction is close enough to the detection (within a pre-defined threshold), we accept the detection as valid. However, if the difference is too large, the detection is assumed to be an outlier. In such cases, we generate a substitute measurement using the UKF-predicted position and assign velocity from a nearby keypoint, effectively smoothing out the discontinuity.
}

% --- ADDED (R1.5): names and quantifies the reliability check as the outlier gate ---
In our implementation, this reliability check uses a fixed displacement threshold of $20$ px between consecutive frames, which we refer to as the outlier gate. When a detection falls more than $20$ px from the previously accepted position of the same keypoint, the gate rejects it and substitutes the UKF-predicted position before the measurement update, while detections within the threshold pass to the filter unchanged. Because the gate runs before the filter, the UKF does not update on a spurious large jump. This is the component that absorbs the residual detection errors introduced by background clutter in \autoref{ssec:bkgrnd_robustness}.
% --- end added ---

This approach ensures that all keypoints, even those that are intermittently occluded or mislocalized, receive smooth and spatially consistent estimates. This is critical for achieving robust visual servoing in cluttered or uncertain environments.

In the following subsections, we present our rationale for choosing the Unscented Kalman Filter (UKF) \citep{wan2000unscented} over the Kalman Filter (KF) \citep{bishop2001introduction} and Extended Kalman Filter (EKF) \citep{fujii2013extended}, and explain why we adopt an inpainting-based approach for occlusion handling instead of relying solely on filtering to estimate missing keypoints after \textit{keypointrcnn\_resnet50\_fpn} prediction.

\subsubsection{Filter Evaluation Strategy and Metrics:}
\label{ssec:filter_videos}
We designed a controlled experimental evaluation strategy to quantitatively assess the effectiveness of different filters under occlusion.
We evaluate keypoint detection accuracy using five occlusion-rich videos, each captured by moving the robot through a predefined set of configurations in its workspace. For each experiment, we record two separate videos: one with visual occlusions introduced using real objects, and another without any occlusions. Both are captured using a fixed external camera in an eye-to-hand configuration. The video without occlusion serves as a ground truth reference i.e., keypoints predicted by the keypoint detection model described in \autoref{ssec:kp_pred} in these unoccluded frames are treated as the true keypoint locations.

For the occluded videos, we again ran the same keypoint detection model, and wherever keypoints were missing due to occlusions, we estimated them using three different filters: Kalman Filter (KF), Extended Kalman Filter (EKF), and Unscented Kalman Filter (UKF).

To compare the performance of these filters, we evaluated the predicted keypoints against the ground truth using the following three metrics:

\begin{itemize}
    \item \textbf{Mean Error (px)}: The average Euclidean distance in pixels between the predicted and ground truth keypoints, computed across all frames and all visible keypoints.
    \item \textbf{Root Mean Squared Error (RMSE):} The square root of the average squared error over all keypoints, providing a robust measure of overall deviation.
    \item \textbf{Failure Rate (\%)}: The percentage of keypoints for which the prediction error exceeded a defined threshold ($10$ pixels). A keypoint that the detector does not report is also counted as a failure, so the failure rate reflects both mislocalization and non-detection, while the mean error and RMSE are computed only over the keypoints each pipeline reports.
\end{itemize}

The comparative results for KF, EKF, and UKF, averaged across all five experiments, are summarized in Table~\ref{tab:filter_eval_results}. Each value represents the mean performance for the five evaluation runs.

The following table reports the effect of temporal filtering on its own, applied directly to the occluded detections with no image reconstruction. It serves as a baseline rather than the proposed pipeline. We evaluate the full pipeline, including inpainting, in \autoref{ssec:inpaint_ablation}.

\begin{table}[t]
\caption{Filtering-only baseline without inpainting. Keypoint estimation error under occlusion for the three temporal filters, averaged across the five sequences. These rows apply temporal filtering alone to the occluded detections and serve as a baseline rather than the proposed pipeline, which we evaluate in \autoref{tab:inpaint_ablation}.}
\label{tab:filter_eval_results}
\centering
\resizebox{0.5\textwidth}{!}{%
\begin{tabular}{|>{\centering\arraybackslash}p{3.0cm}|>{\centering\arraybackslash}p{2.2cm}||>{\centering\arraybackslash}p{2.0cm}|>{\centering\arraybackslash}p{2.0cm}|}
\hline
\textbf{Filters} & \textbf{Mean Error (px)} & \textbf{RMSE (px)} & \textbf{Failure Rate \%} \\ 
\hline
\hline
Kalman            & $18.41$ & $39.34$ & $24.94$ \\
\hline
Extended Kalman   & $18.41$ & $39.34$ & $24.94$ \\
\hline
Unscented Kalman  & $18.32$ & $39.37$ & $24.66$ \\
\hline
\end{tabular}
}
\end{table}

Although the numerical performance of the Unscented Kalman Filter (UKF) is only marginally better than that of the Kalman Filter (KF) and Extended Kalman Filter (EKF) in our baseline evaluation, we choose to use UKF for its ability to naturally handle nonlinear motion dynamics without requiring Jacobian computations. The Kalman Filter (KF) assumes a linear motion model, which is often insufficient in our setup where keypoint motion can exhibit nonlinear dynamics due to the adaptive control loop and the online Jacobian estimation process. The Extended Kalman Filter (EKF) improves upon this by linearizing the system around the current estimate using Jacobians. However, EKF still relies on accurate and consistent linear approximations, which are difficult to obtain in our vision-based setting. Specifically, because we do not model the robot kinematics or camera explicitly, we cannot derive reliable Jacobians for the keypoint motion model—making EKF prone to errors when keypoints undergo abrupt shifts or when visibility conditions change rapidly.
In practice, we observe that EKF and KF produce nearly identical results in our system. This is because the motion and measurement models we use are relatively simple, and the measurement function (direct keypoint observation) is linear. Moreover, without access to reliable Jacobians, EKF effectively behaves like KF, offering no additional advantage.
In contrast, the Unscented Kalman Filter (UKF) handles nonlinearities more robustly by using sigma points to propagate the full distribution of the state through the nonlinear motion model. It does not require explicit Jacobian computation, which aligns well with our model-free, image-based approach. By providing smoother and more consistent keypoint trajectories even under occlusion, UKF helps preserve the integrity of the Jacobian estimation and, consequently, the stability of the overall visual servoing control loop. Thus, UKF is not only more robust to missing or noisy detections, but also better suited to the adaptive Jacobian estimation mechanism critical to our visual servoing system.

% --- ADDED (R2.3c): why not constrained estimators such as MHE ---
Constrained estimators, such as moving horizon estimation or joint-limit-aware filters, can enforce physical feasibility on the estimated state. However, they require the kinematic model and joint geometry that our framework avoids, since their constraints are expressed in terms of link lengths, joint limits, and forward kinematics, which we do not assume to be available. The UKF operates only on image-space keypoint trajectories and does not need such a model, which is why we use it here. For deployments where a kinematic model is available, a constrained estimator that encodes joint limits and link geometry is a reasonable extension, which we leave for future work.
% --- end added ---

\subsubsection{Filtering With vs. Without Inpainting:}
As shown in \autoref{tab:filter_eval_results}, relying solely on filtering, even with the UKF, results in a high failure rate when handling occluded keypoints. To address this, we repeated the same set of experiments described earlier, but with our inpainting model applied prior to keypoint detection using \textit{keypointrcnn\_resnet50\_fpn}. As summarized in \autoref{tab:inpainting_eval_results}, the combination of inpainting and UKF significantly reduces prediction errors and failure rates, demonstrating the shortcomings of only temporal filtering and effectiveness of inpainting in restoring key visual features for accurate keypoint estimation.

% \begin{table}[t]
% \caption{Inpainting with Filter evaluation results averaged across five experiments}
% \label{tab:inpainting_eval_results}
% \centering
% \resizebox{0.5\textwidth}{!}{%
% \begin{tabular}{|>{\centering\arraybackslash}p{3.0cm}||>{\centering\arraybackslash}p{2.2cm}|>{\centering\arraybackslash}p{2.2cm}|>{\centering\arraybackslash}p{2.2cm}|>{\centering\arraybackslash}p{2.4cm}|}
% \hline
% \textbf{Filters} & \textbf{Mean Error (px)} & \textbf{RMSE (px)} & \textbf{Failure Rate \%} & \textbf{Inference Time (s)} \\ 
% \hline
% \hline
% Only UKF            & $18.32$ & $39.37$ & $24.66$ & $19.91$  \\
% \hline
% Inpainting with UKF   & $3.38$ & $7.29$ & $6.06$ & $29.22$ \\
% \hline
% \end{tabular}
% }
% \end{table}

\begin{table}[t]
\caption{Inpainting with Filter evaluation results averaged across five experiments}
\label{tab:inpainting_eval_results}
\centering
\resizebox{0.5\textwidth}{!}{%
\begin{tabular}{|>{\centering\arraybackslash}p{3.0cm}||>{\centering\arraybackslash}p{2.4cm}|>{\centering\arraybackslash}p{2.4cm}|>{\centering\arraybackslash}p{2.4cm}|}
\hline
\textbf{Filters} & \textbf{Mean Error (px)} & \textbf{RMSE (px)} & \textbf{Failure Rate \%} \\
\hline
\hline
Only UKF            & $18.32$ & $39.37$ & $24.66$ \\
\hline
Inpainting with UKF   & $3.38$ & $7.29$ & $6.06$ \\
\hline
\end{tabular}
}
\end{table}

\subsection{Adaptive Visual Servoing for Configuration Space Control}
\label{ssec:adaptive_vs}
The keypoints inferred from our trained model, as described in \autoref{ssec:kp_pred}, are used as image features in an adaptive visual servoing scheme. We define a feature error vector $\mathbf e$ between the current keypoint locations $p$ and the desired keypoint locations $p^*$ as shown in \autoref{eq:error}.
\begin{equation}\label{eq:error}
    \mathbf{e} = p - p^{*}
\end{equation}
We estimate a combined hand-eye-Jacobian $\hat{J}_{c}$, that relays the change in feature locations in image space ($\dot{p}$) to the robot's joint velocities ($\dot{q}$).
\begin{equation}\label{eq:capital-q}
    \dot{Q} = {[\dot{q}[k-n+1],\ \dot{q}[k-n+2],\ \hdots ,\ \dot{q}[k]]}^T
\end{equation}
\begin{equation}\label{eq:capital-c}
    \dot{P} = {[\dot{p}[k-n+1],\ \dot{p}[k-n+2],\ \hdots ,\ \dot{p}[k]]}^T
\end{equation}
Specifically, we collect pairs of visuo-motor data for the last $n$ samples of applied robot joint velocities $\dot{Q}$, as shown in \autoref{eq:capital-q} and change in image feature (keypoint) locations $\dot{P}$, as shown in \autoref{eq:capital-c}. The collected window of visuo-motor data pairs is used in a single-shot least squares optimization to obtain adaptive updates, for each row of $\hat{J}_{c}$, as described in \autoref{eq:update-rule}. The adaptive updates are added to the current estimation of $\hat{J}_{c}$ as shown in \autoref{eq:jacobian-update}.
\begin{equation}\label{eq:update-rule}
    \Delta{\hat{J}_{c_i}[k]} = \gamma \dot{Q}^T (\dot{Q} J_{c_i}[k] - \dot{P})
\end{equation}
\begin{equation}\label{eq:jacobian-update}
    \hat{J}_{c_i}[k+1] = \Delta{\hat{J}_{c_i}[k]} + \hat{J}_{c_i}[k]
\end{equation}
Finally, the image feature error is utilized with a proportional control law and the current estimate of $\hat{J}_{c}$, as shown in \autoref{eq:control_law}, to generate reference joint velocities that drive the robot to its desired configuration in the image.
\begin{equation} \label{eq:control_law}
    \mathbf{\dot q_r} = -\lambda J_{c}^{+} \mathbf{e}
\end{equation}
In our implementation, we initialize the adaptive scheme online by providing the system with a small excitation velocity trajectory to estimate an initial combined Jacobian. After this initial estimation, the algorithm continuously updates the Jacobian in runtime during control. For adaptive schemes, it is important to select the window size carefully. Large window sizes prevent sudden spikes in robot joint velocities but increase the convergence time of the estimated combined Jacobian whereas smaller window sizes may cause sudden changes and spikes in the joint velocities generated from the estimated Jacobian. The blue part of the workflow in \autoref{fig:control_flow} depicts this part of the proposed pipeline.
The overall control pipeline using predicted keypoints from images is illustrated in \autoref{fig:control_flow}
\section{Experiments and Results} \label{sec:vision}
    % Experiments and Results

In this section, we present both qualitative and quantitative evaluation of our proposed pipeline under uncertainty and challenging visual conditions. The experiments are conducted using a Franka Emika Panda arm \citep{zhang2020modular} and an Intel RealSense D435i \citep{servi2024comparative} camera in an eye-to-hand configuration. We perform five  types of experiments: (i) keypoint detection accuracy under full visibility, (ii) keypoint detection accuracy under partial visibility, (iii) keypoint detection accuracy against a cluttered background, (iv) per-frame latency of the perception pipeline, and (v) control performance, including a direct evaluation of the online interaction-matrix estimate.

To evaluate keypoint detection accuracy under full visibility, we capture images of the Panda arm in $20$ distinct configurations, spanning a large portion of its planar workspace. For each configuration, we record three versions: one without markers and two with ArUco markers—one set containing $4$ markers along the robot’s body and another containing $5$. In addition, to study out-of-plane behavior, we collect a separate set of $20$ images with $8$ markers distributed along the arm while actuating the first spatial joint with the $3$ planar joints to induce out-of-plane motion. The marker-based images serve as ground truth for keypoint locations, while the markerless images are used to evaluate the predictive accuracy of our \textit{keypointrcnn\_resnet50\_fpn} model on natural, unmarked robot features. The rationale behind evaluating multiple keypoint sets is to assess control performance as we increase the controlled degrees of freedom: the $4$-keypoint setup was chosen to control two planar joints using $3$ of the keypoints as visual features; the $5$-keypoint setup controls three planar joints using all $5$ keypoints; and the $8$-keypoint setup controls four joints (three planar $+$ one spatial) using $6$ of the $8$ keypoints as visual features.

The trained keypoint detection models used in our accuracy and control evaluations are archived on Zenodo: 
\begin{itemize}
    \item $2$-joint/$3$-keypoint (\textit{DOI: 10.5281/zenodo.17331025}), 
    \item $3$-joint/$5$-keypoint (\textit{DOI: 10.5281/zenodo.17331054}), and 
    \item out-of-plane ($4$-joint/$6$-keypoint) (\textit{DOI: 10.5281/zenodo.17331102}).
\end{itemize}

We evaluate keypoint detection accuracy under partial visibility using the two sets of videos described in \autoref{ssec:filter_videos}. Each set contains five videos corresponding to the same robot motion trajectories, with one set captured under full visibility and the other with partial occlusions introduced into the scene. The unoccluded videos serve as reference data, where keypoints predicted by the Keypoint R-CNN (\textit{keypointrcnn\_resnet50\_fpn}) model (as described in \autoref{ssec:kp_pred}) are treated as ground truth. Partial visibility experiments are carried out with only the $5$-keypoints setup. For each frame in the occluded videos, we compare two keypoint detection pipelines: (1) Keypoint R-CNN with UKF correction, where keypoints are directly predicted from the occluded image and any missing keypoints are estimated using the Unscented Kalman Filter (UKF), and (2) Keypoint R-CNN with inpainting and UKF, where occluded images are first passed through our proposed inpainting network, followed by keypoint prediction and UKF-based correction. This dual setup allows us to assess the relative contributions of inpainting and temporal filtering to keypoint accuracy. 

Finally, we evaluate how these predicted keypoints perform as visual features in our adaptive visual servoing scheme.

\subsection{Keypoint Detection Accuracy - Full Visibility}
\begin{figure*}[!t]
      \centering
      \vspace{5pt}
      \includegraphics[width=\textwidth]{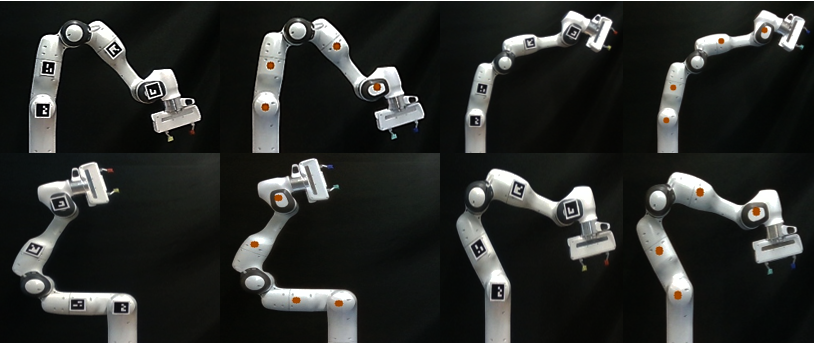}
      \caption{The images with ArUco marker are the ground truth and the red dot on the corresponding image is the keypoint predicted by Keypoint R-CNN model trained with data generated using our present method.}
      \label{fig:acc_comp_04}
\end{figure*}

\begin{figure}[!t]
      \centering
      \includegraphics[width=0.45\textwidth]{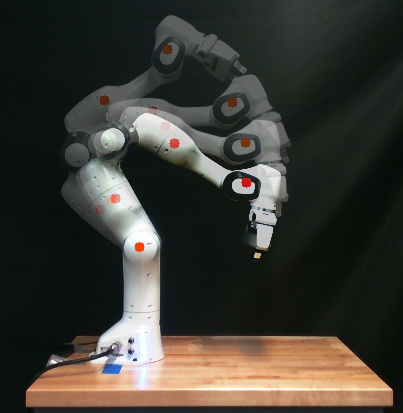}
      \caption{Smooth keypoint detection during continuous robot motion for $4$ keypoints configuration.}
      \label{fig:motion_detect_04}
\end{figure}

We evaluate keypoint detection accuracy under full visibility, using error thresholds of $5$ and $10$ pixels. For the $4$-keypoint setup, we achieve a $100\%$ detection rate within a $10$ pixel margin and $98.75\%$ within a $5$ pixel margin. The average localization error across all detections is $2.19$ pixels. Some examples of the detection are shown in \autoref{fig:acc_comp_04}, and \autoref{fig:motion_detect_04} illustrates consistent keypoint detection during robot motion.

\begin{figure*}[!t]
      \centering
      \vspace{5pt}
      \includegraphics[width=\textwidth]{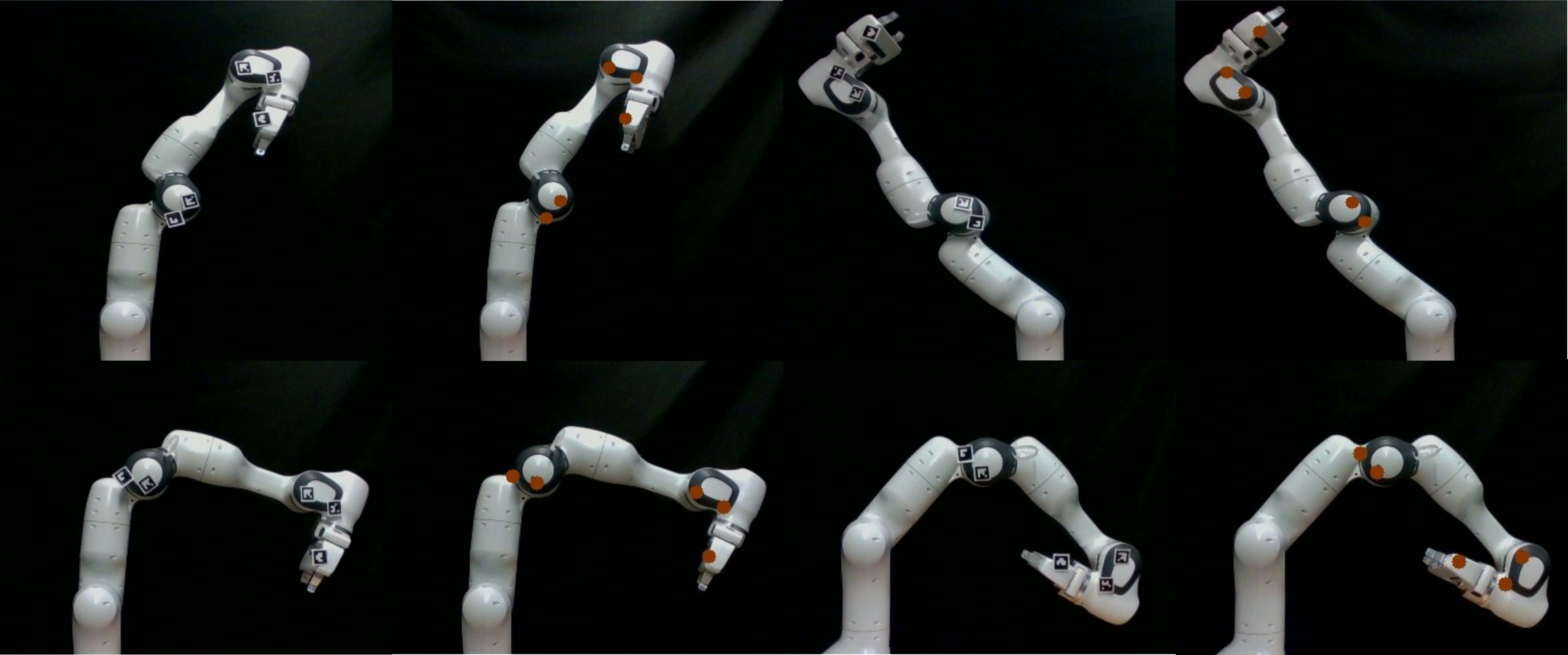}
      \caption{The images with ArUco marker are the ground truth and the red dot on the corresponding image is the keypoint predicted by Keypoint R-CNN model trained with data generated using our present method.}
      \label{fig:acc_comp_05}
\end{figure*}

\begin{figure}[!t]
      \centering
      \includegraphics[width=0.45\textwidth]{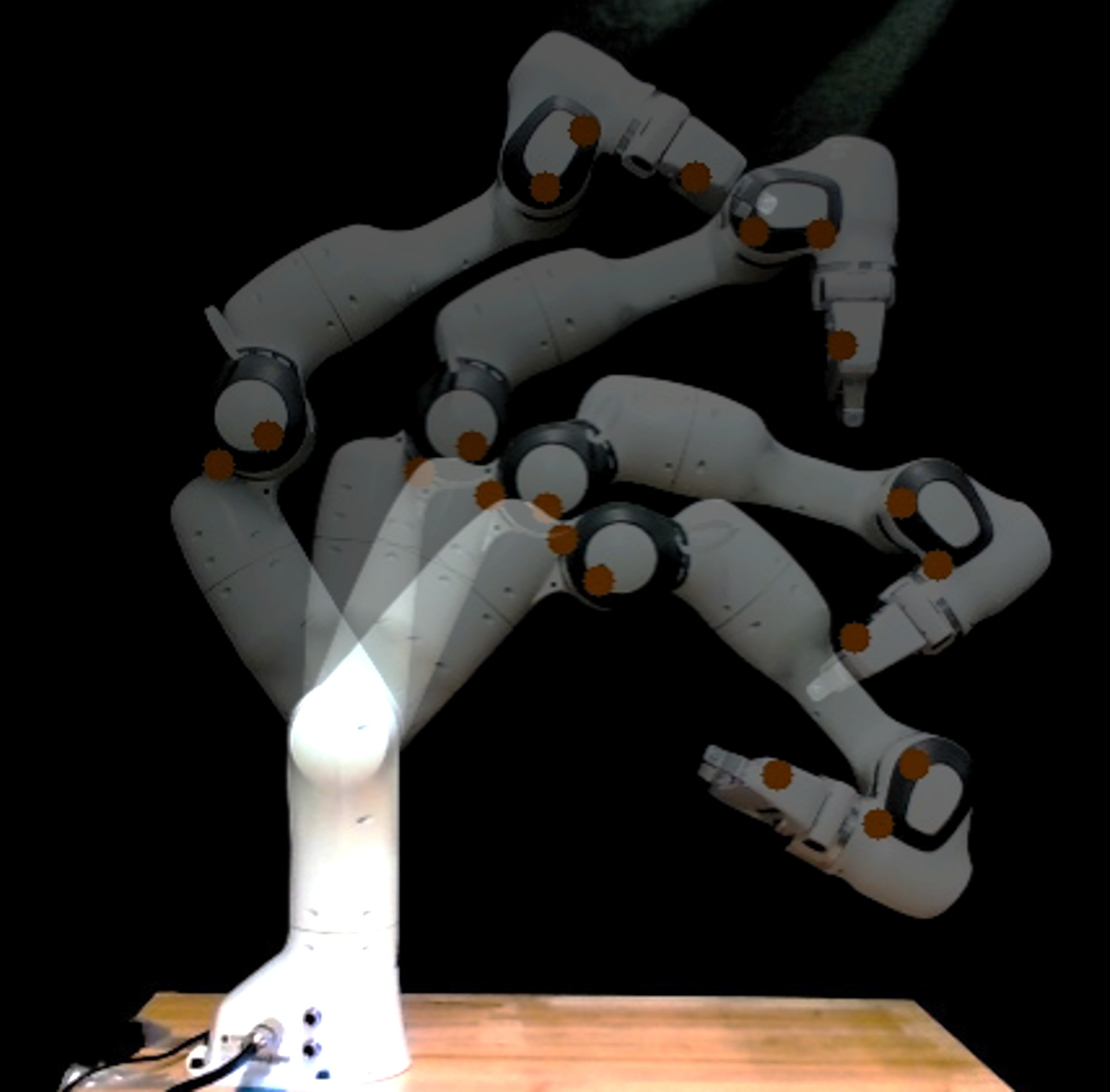}
      \caption{Smooth keypoint detection during continuous robot motion for $5$ keypoints configuration.}
      \label{fig:motion_detect_05}
\end{figure}

For the $5$-keypoint setup, we observe near identical performance: a $100\%$ detection rate within $10$ pixels and $97.78\%$ within $5$ pixels margin of error and an average localization error of $1.04$ pixels. This consistency across both configurations highlights the robustness of the model in accurately localizing natural keypoints on the robot body under ideal visual conditions. Similar to the $4$-keypoint experiments we have conducted a qualitative visualization of the detection accuracy as illustrated in \autoref{fig:acc_comp_05}. \autoref{fig:motion_detect_05} demonstrates consistent keypoint detection during robot motion.

\begin{figure*}[!t]
      \centering
      \vspace{5pt}
      \includegraphics[width=\textwidth]{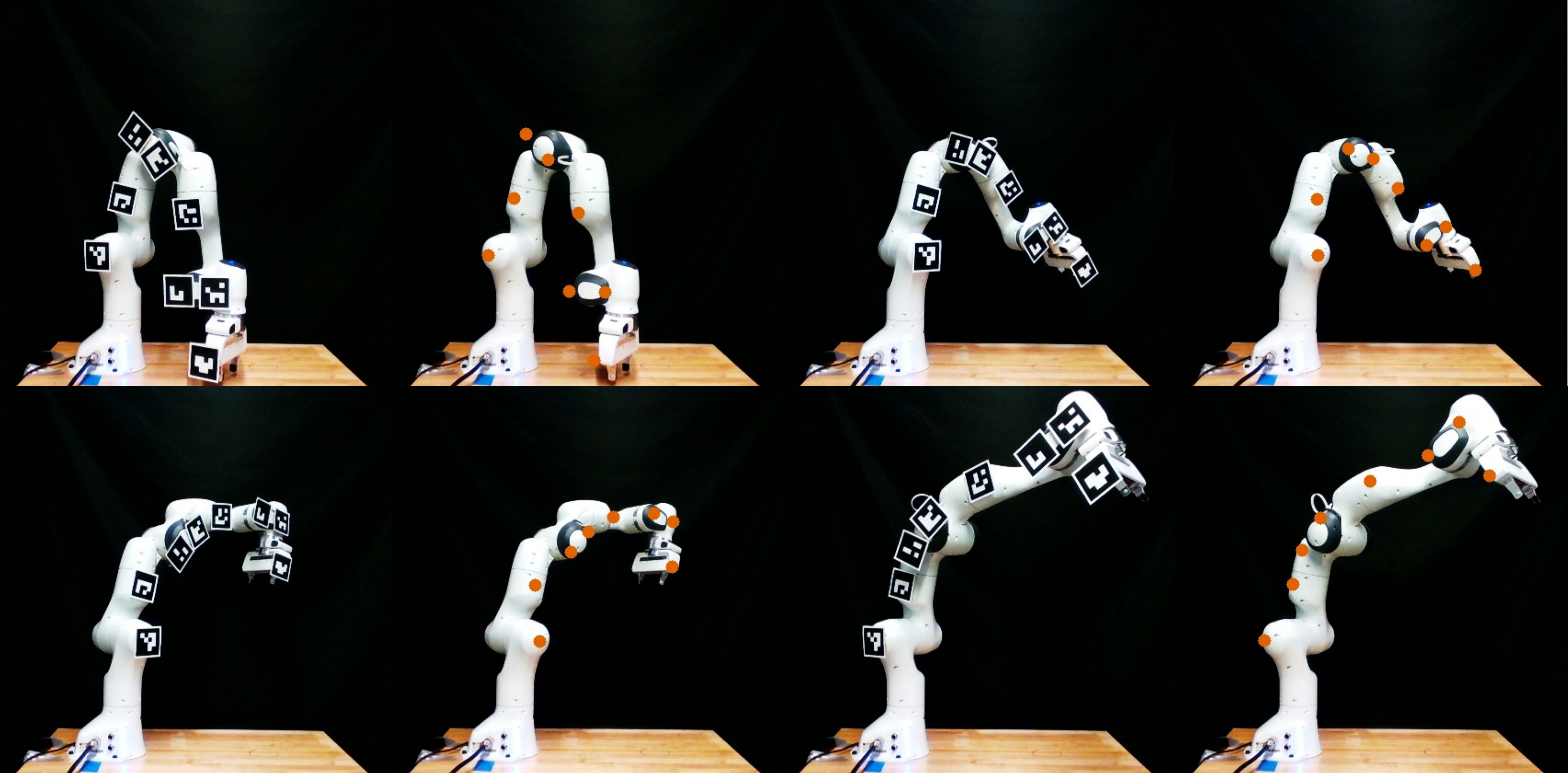}
      \caption{The images with ArUco marker are the ground truth and the red dot on the corresponding image is the keypoint predicted by Keypoint R-CNN model trained with data generated using our present method.}
      \label{fig:acc_comp_3d}
\end{figure*}

\begin{figure}[!t]
      \centering
      \includegraphics[width=0.45\textwidth]{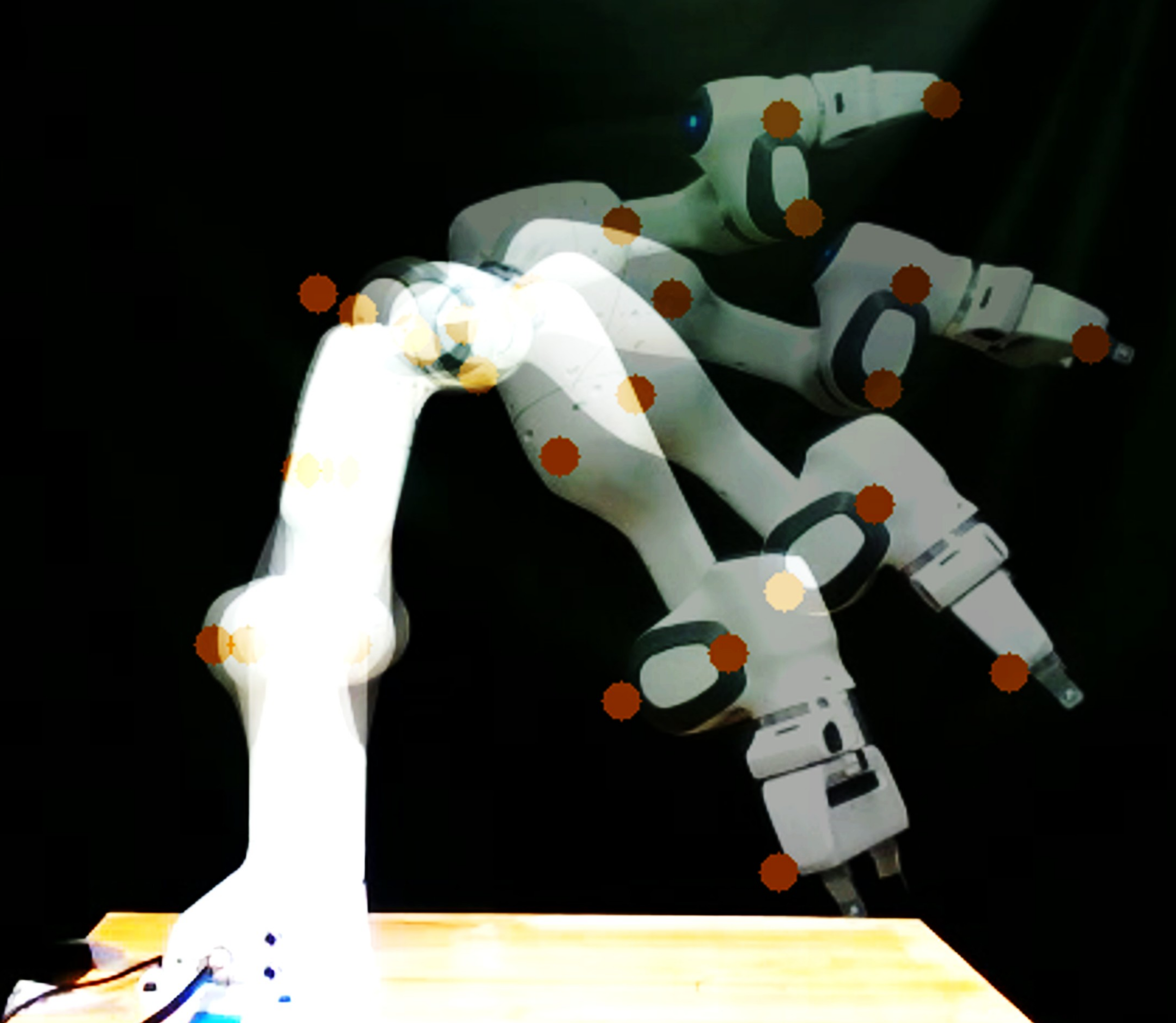}
      \caption{Smooth keypoint detection during continuous robot motion in 3D with $8$ keypoints configuration.}
      \label{fig:motion_detect_3d}
\end{figure}

For the $8$-keypoint (3D) setup, we obtain a $100\%$ detection rate within $10$ pixels, $97.69\%$ within $5$ pixels, and an average localization error of $0.933$ pixels, demonstrating strong localization even with added out-of-plane variability. Qualitative visualizations for the $8$-keypoint (3D) setup are also provided: \autoref{fig:acc_comp_3d} illustrates detection accuracy overlays across out-of-plane viewpoints, and \autoref{fig:motion_detect_3d} shows consistent keypoint detection during out-of-plane robot motion.

\begin{figure*}[!t]
  \centering

  \includegraphics[width=0.775\textwidth]{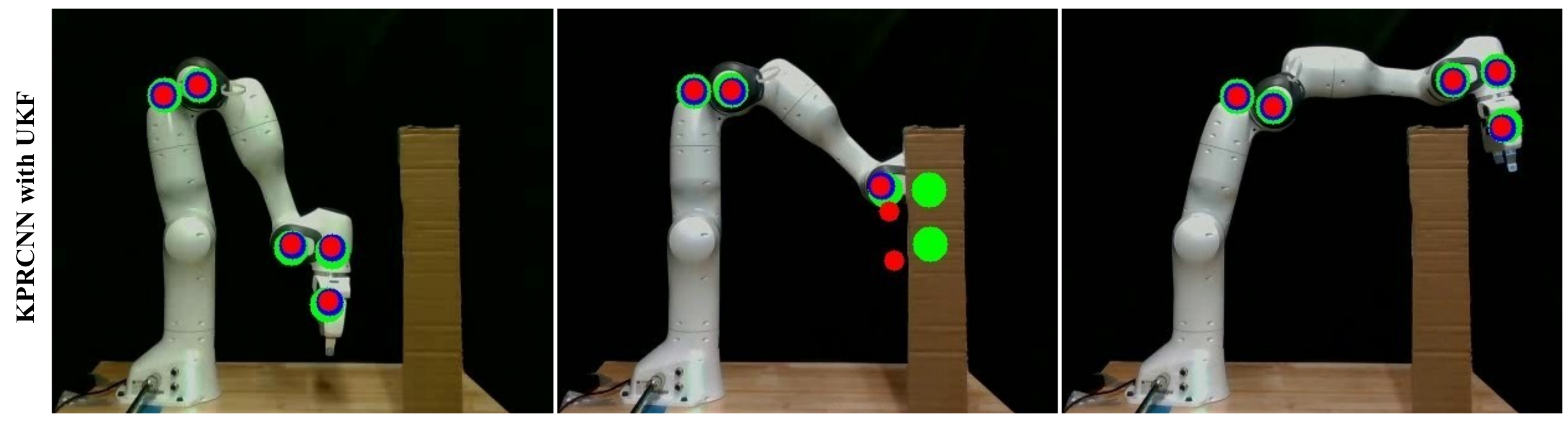}

  \includegraphics[width=0.755\textwidth]{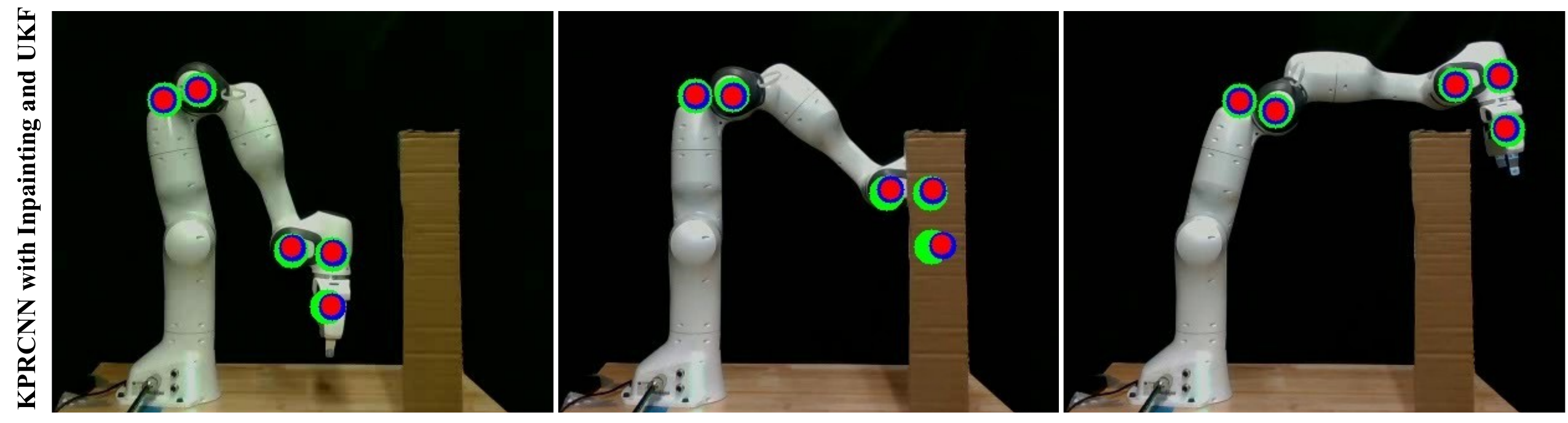}

  \includegraphics[width=0.755\textwidth]{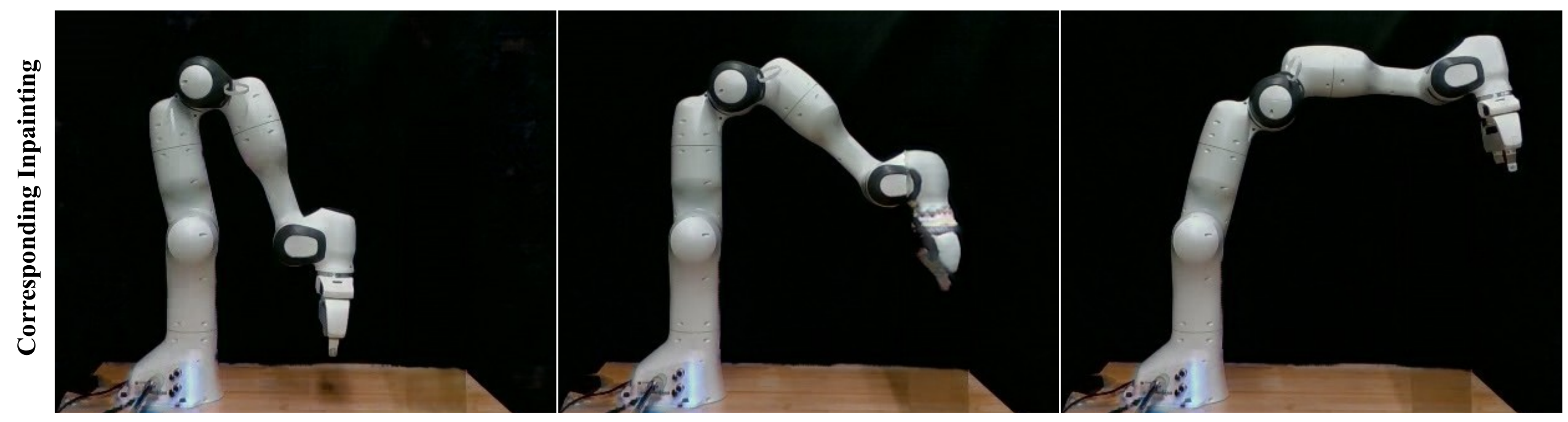}

  \includegraphics[width=0.775\textwidth]{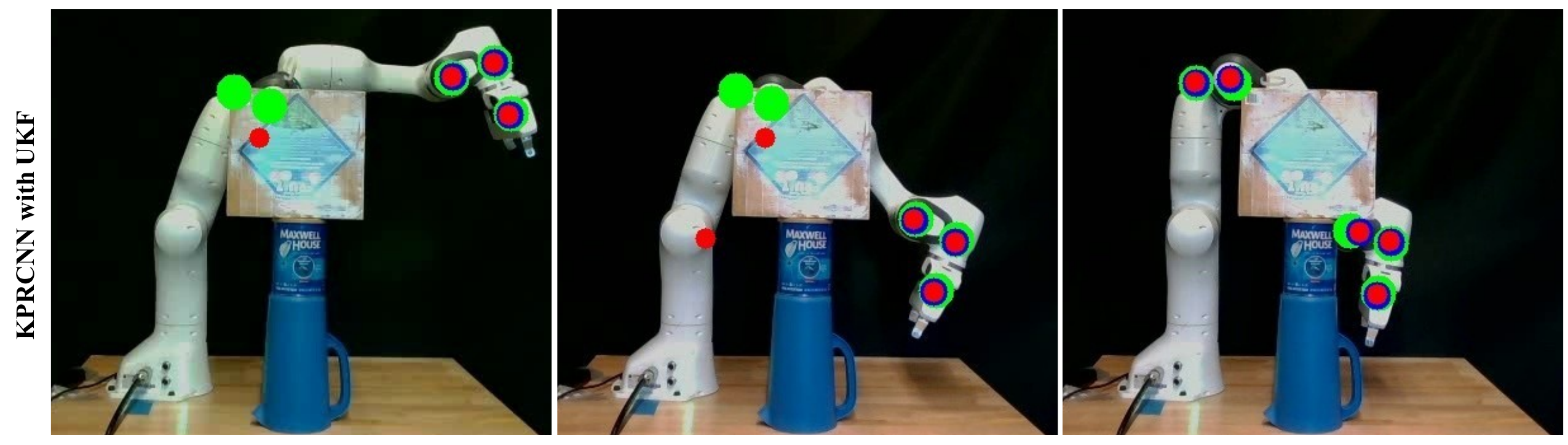}

  \includegraphics[width=0.775\textwidth]{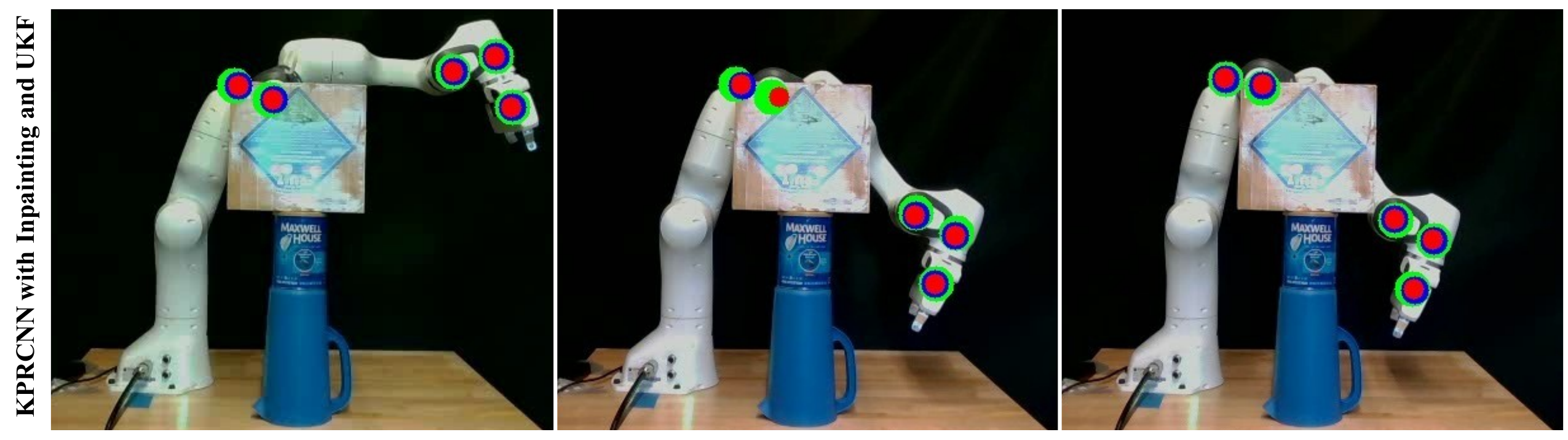}

  \includegraphics[width=0.775\textwidth]{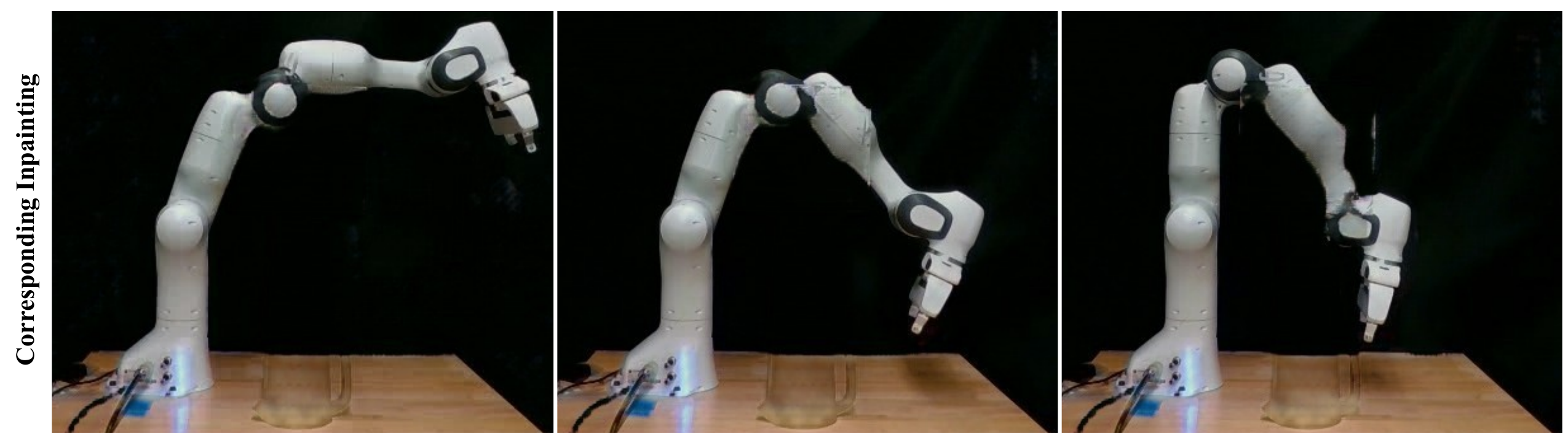}
  
  \vspace{-5pt}

  \caption{Example video frames from two experiments illustrating keypoint prediction and correction under occlusion. Each set of three rows corresponds to a single experiment. The first row shows real-time keypoint detection on occluded images using Keypoint R-CNN (blue circles), with missing keypoints corrected using the Unscented Kalman Filter (red circles). Ground truth keypoints obtained under full visibility are overlaid in green for reference. The second row uses inpainted images as input to the Keypoint R-CNN, followed by UKF correction of missing keypoints. The third row shows the corresponding inpainted reconstructions used in the second row. The comparison highlights the effect of inpainting on improving keypoint prediction accuracy and visual stability under occlusion.}
  \label{fig:filter_videos}
\end{figure*}

\subsection{Keypoint Detection Accuracy - Partial Visibility}
We measure keypoint detection accuracy with a margin of error of $10$ pixels with the same sets of videos described in \autoref{ssec:filter_videos}. 

\autoref{tab:kp_det_acc} show the average keypoint detection accuracy over the 5 videos with only inpainted images and then with UKF correction.
\begin{table}[h]
\centering
\caption{Keypoint detection accuracy before and after UKF correction.}
\label{tab:kp_det_acc}
\begin{tabular}{|c|c|c|}
\hline
\textbf{Experiment} & \makecell{\textbf{Only}\\\textbf{Inpainting}\\\textbf{(\%)}} & \makecell{\textbf{After UKF}\\\textbf{Correction}\\\textbf{(\%)}} \\
\hline
exp\_01 & 99.526 & 100.000 \\
\hline
exp\_02 & 99.341 & 99.615 \\
\hline
exp\_03 & 88.767 & 89.425 \\
\hline
exp\_04 & 94.619 & 97.310 \\
\hline
exp\_05 & 77.229 & 83.976 \\
\hline
\end{tabular}
\end{table}

From experiments \texttt{exp\_01} and \texttt{exp\_02}, the difference between raw and corrected accuracy is minimal, indicating that the inpainting model alone performs sufficiently well in moderate occlusion scenarios. However, in more challenging settings such as \texttt{exp\_03} to \texttt{exp\_05}, the UKF significantly improves detection accuracy by providing temporally consistent corrections when raw detections fail or are unreliable. As illustrated in \autoref{fig:filter_videos}, the second group of rows corresponds to such a challenging occlusion scenario, where larger portions of the robot are obscured by visually complex objects. The actual videos are added in Multimedia Extension $1$.

\subsection{Effect of Inpainting: Comparison with Filtering-Only and Occlusion-Trained Detection}
\label{ssec:inpaint_ablation}
 
To examine whether inpainting is necessary to estimate keypoints under occlusion, or whether a simpler strategy is sufficient, we evaluate five detection and estimation pipelines on the same five occluded sequences described in \autoref{ssec:filter_videos}. We score each pipeline against the full-visibility ground truth on the $5$ keypoints used as control features. The five pipelines are:
 
\begin{itemize}
    \item[(i)] the clean-trained Keypoint R-CNN applied directly to the occluded frames,
    \item[(ii)] the same detector followed by UKF correction,
    \item[(iii)] our full pipeline, in which each frame is first reconstructed by the Attention U-Net GAN and then passed to the detector and the UKF,
    \item[(iv)] a Keypoint R-CNN trained directly on artificially occluded images, with no inpainting at runtime, archived on Zenodo \textit{(DOI: 10.5281/zenodo.20978897)}, and
    \item[(v)] the occlusion-trained detector with UKF correction.
\end{itemize}
 
The first three pipelines share a single detector trained only on clean, unoccluded images, so the comparison between (ii) and (iii) isolates the effect of inpainting, since the detector and the filter are identical and the only difference is image reconstruction. The last two pipelines follow the alternative of retraining the detector on occluded data instead of reconstructing the image.
 
\autoref{tab:inpaint_ablation} summarizes the results. The clean detector applied directly to the occluded frames is the most accurate on the keypoints it detects ($1.48$ px), but it recovers only $73.28\%$ of the control keypoints, because an independent per-keypoint detector has nothing to detect in a region that is fully occluded. It therefore cannot provide a complete feature set for control, which motivates the need for a recovery mechanism. Filtering alone restores the detection rate to $100\%$, but the mean error rises to $11.92$ px and the failure rate to $22.48\%$. This is primarily because the UKF can hold a keypoint that remains visible but has no measurement for one that is occluded, so its prediction drifts during longer occlusions. When we reconstruct the frame before detection, the missing measurement is restored, and the mean error reduces from $11.92$ px to $2.82$ px and the failure rate from $22.48\%$ to $5.09\%$, while the detection rate remains at $100\%$.
 
The occlusion-trained detector also reaches a $100\%$ detection rate and performs well ($3.38$ px mean error, $3.02\%$ failure rate), which confirms that training a detector directly on occluded images is a viable alternative. Our inpainting pipeline is more accurate ($2.82$ px against $3.38$ px), although the occlusion-trained detector has a marginally lower failure rate at the $10$ px threshold. The main difference between the two is the detector itself. The occlusion-trained detector must be retrained for the occlusion condition, whereas our pipeline handles occlusion at runtime using the same clean-trained detector that we use elsewhere in this work. As shown in \autoref{ssec:bkgrnd_robustness}, this single clean-trained detector also transfers to a cluttered, previously unseen background without retraining, although that transfer concerns the visible keypoints and not the occlusion handling, which we revisit in \autoref{ssec:occ_clutter}. The inpainting stage additionally produces a reconstructed image, which is useful for inspection and for tasks other than control.
 
\begin{table}[t]
\caption{Keypoint estimation under occlusion, averaged across the five occluded sequences and scored on the $5$ control keypoints. The first three pipelines share one detector trained on clean images, while the last two use a detector retrained on occluded images.}
\label{tab:inpaint_ablation}
\centering
\resizebox{0.5\textwidth}{!}{%
\begin{tabular}{|>{\centering\arraybackslash}p{3.2cm}||>{\centering\arraybackslash}p{1.8cm}|>{\centering\arraybackslash}p{1.8cm}|>{\centering\arraybackslash}p{1.8cm}|>{\centering\arraybackslash}p{1.8cm}|}
\hline
\textbf{Pipeline} & \textbf{Mean Error (px)} & \textbf{RMSE (px)} & \textbf{Failure Rate \%} & \textbf{Detection \%} \\
\hline
\hline
Visible-only                       & $1.48$  & $2.42$  & $26.97$ & $73.28$ \\
\hline
Filtering-only (UKF)               & $11.92$ & $24.50$ & $22.48$ & $100.00$ \\
\hline
\textbf{Inpainting + UKF (ours)}   & $\mathbf{2.82}$ & $\mathbf{5.35}$ & $\mathbf{5.09}$ & $\mathbf{100.00}$ \\
\hline
Occlusion-trained                  & $3.38$  & $5.44$  & $3.02$  & $100.00$ \\
\hline
Occlusion-trained + UKF            & $3.21$  & $5.02$  & $2.74$  & $100.00$ \\
\hline
\end{tabular}
}
\end{table}

\subsection{Runtime and Per-frame Latency}
\label{ssec:latency}
 
To address the per-frame latency of the perception pipeline, we profiled each stage separately on an NVIDIA GeForce RTX 3080 laptop GPU, averaged over $485$ frames pooled across the five occluded sequences of \autoref{ssec:filter_videos}, timing each stage with a synchronized wall-clock timer. \autoref{tab:latency} reports the result. Keypoint detection is the dominant cost at $63.7$ ms per frame, the inpainting reconstruction adds $21.2$ ms, and the outlier gate together with the UKF update account for under $2.2$ ms. With inpainting active, the full pipeline runs at $87.1$ ms per frame, which corresponds to $11.5$ Hz. Under full visibility, where the reconstruction step is skipped, the pipeline runs at $65.9$ ms per frame, or $15.2$ Hz. Both rates remain within the $100$ ms budget of the $10$ Hz visual servoing loop described in \autoref{ssec:ukf}, so the pipeline operates in real time, with the inpainting stage accounting for roughly a quarter of the per-frame cost.
 
\begin{table}[t]
\caption{Per-frame latency of each pipeline stage on an NVIDIA GeForce RTX 3080 laptop GPU, averaged over $485$ frames pooled across the five occluded sequences.}
\label{tab:latency}
\centering
\resizebox{0.5\textwidth}{!}{%
\begin{tabular}{|>{\centering\arraybackslash}p{5.0cm}||>{\centering\arraybackslash}p{2.5cm}|}
\hline
\textbf{Stage} & \textbf{Latency (ms)} \\
\hline
\hline
Inpainting (Attention U-Net)        & $21.2$ \\
\hline
Keypoint detection (Keypoint R-CNN) & $63.7$ \\
\hline
Outlier gate                        & $0.1$  \\
\hline
UKF update                          & $2.0$  \\
\hline
\hline
End-to-end, with inpainting         & $87.1$ \\
\hline
End-to-end, full visibility         & $65.9$ \\
\hline
\end{tabular}
}
\end{table}

\subsection{Robustness to Background Clutter}
\label{ssec:bkgrnd_robustness}
 
All the experiments above were conducted against a uniform black backdrop. To assess whether the trained perception pipeline transfers to a visually different environment, we recorded five additional motion sequences in a cluttered laboratory scene. The scene contains several distractors with a robot-like appearance, including a coat rack holding a dark jacket against a light wall, reflection of a fluorescent tube directly behind the arm and other specular reflections from a glass door, exposed ceiling pipes, and the wooden bench on which the robot is placed has colored tape near the base. None of these conditions are present in the training data.
 
To obtain ground truth in this setting, we placed ArUco markers at the $8$ keypoint locations along the arm and recorded each sequence twice, once with markers and once without, driving the robot through the same start and goal poses so that the frames correspond by index. The marker centers provide the reference keypoint positions, and the detector runs on the marker-free sequence. We evaluate the deployed detector, which is the same model used throughout this work and is trained only on the original black-background data, using three pipelines:
 
\begin{itemize}
    \item raw detection,
    \item detection with the UKF, and
    \item detection with the outlier gate described in \autoref{ssec:ukf} followed by the UKF.
\end{itemize}
 
\begin{figure*}[!t]
    \centering
    \includegraphics[width=\textwidth]{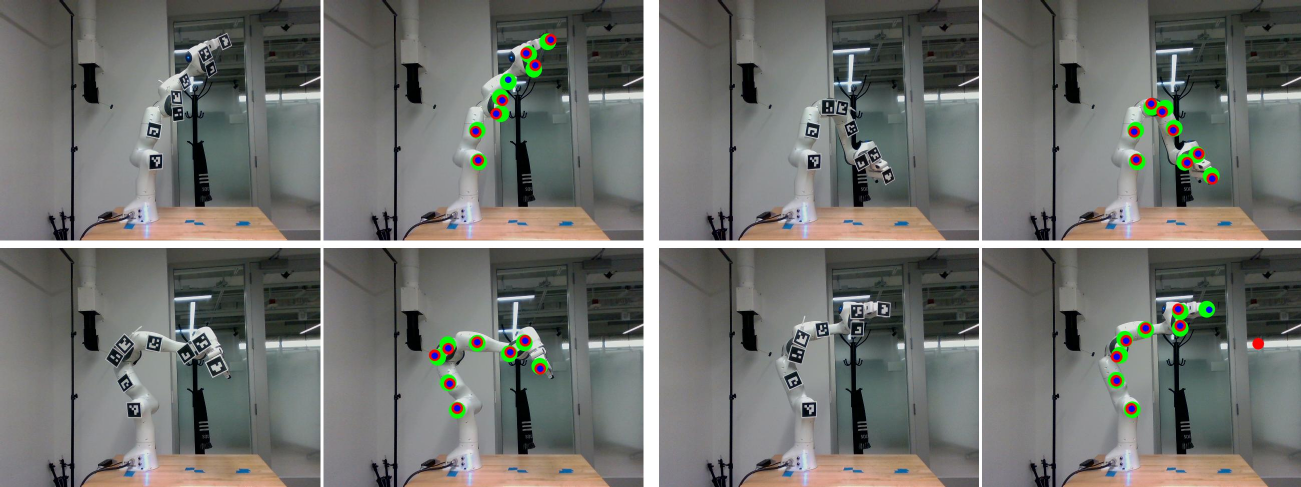}
    \caption{
    Qualitative keypoint detection on the cluttered laboratory background. Four sequences are shown, each as a pair with the ArUco-marker frame on the left providing the ground-truth keypoint positions and the corresponding marker-free frame on the right. On the right frame of each pair, the green circle is the ArUco ground truth, the red circle is the raw Keypoint R-CNN detection, and the blue circle is the final estimate after the outlier gate and the UKF. In the top-left pair the raw detector misses one keypoint, which therefore has no red circle, and the gate together with the UKF recovers it so the blue estimate stays on the keypoint. In the top-right and bottom-left pairs the raw detector localizes every keypoint correctly, so all three circles coincide. In the bottom-right pair a raw red detection lands far from its marker, an outlier caused by the background clutter, and the gate rejects it so the blue estimate stays on the true keypoint. The detector is the deployed model trained only on the black-background data, with no retraining for this scene.
    }
    \label{fig:bkgrnd_qualitative}
\end{figure*}

\autoref{tab:bkgrnd_robustness} summarizes the results across the five sequences ($8$ keypoints, $178$ frames each). The raw detector locates the marked keypoints to within $5.33$ px on average at a $98.04\%$ detection rate, even though it has never seen this background. Adding the gate and the UKF brings the mean error to $5.05$ px, the RMSE from $8.92$ px to $5.46$ px, and the failure rate to $1.93\%$ at a $100\%$ detection rate. The gate accounts for most of this improvement, since it rejects the occasional large detection error caused by the distractors before that error reaches the filter, reducing the RMSE by $39\%$ compared to the raw output. \autoref{fig:bkgrnd_qualitative} illustrates this, with a raw detection landing far from its marker while the gated estimate stays on the keypoint. 
 
We note two points about these results. The ground truth here is the ArUco marker center, which is a different reference from the clean-detector ground truth used under full visibility, so these values are not directly comparable to the controlled-background results and should not be read as a degradation. We also characterize robustness on a single cluttered scene, and a broader study across illumination and background conditions is left for future work. Even with these caveats, a mean error close to $5$ px and a $1.93\%$ failure rate on an out-of-distribution scene, without any retraining, indicate that the clean-trained detector transfers to a visually diverse background. This transfer holds for the keypoints that remain visible. Whether the occlusion handling also transfers to a changed background is a separate question, which we examine in \autoref{ssec:occ_clutter}. See Multimedia Extension $10$ for the demonstration videos.
 
\begin{table}[t]
\caption{Keypoint detection on a cluttered laboratory background with ArUco marker centers as ground truth, averaged across five sequences. The detector is the deployed model, trained only on the black-background data, with no retraining for this scene.}
\label{tab:bkgrnd_robustness}
\centering
\resizebox{0.5\textwidth}{!}{%
\begin{tabular}{|>{\centering\arraybackslash}p{3.4cm}||>{\centering\arraybackslash}p{1.8cm}|>{\centering\arraybackslash}p{1.8cm}|>{\centering\arraybackslash}p{1.8cm}|>{\centering\arraybackslash}p{1.8cm}|}
\hline
\textbf{Pipeline} & \textbf{Mean Error (px)} & \textbf{RMSE (px)} & \textbf{Failure Rate \%} & \textbf{Detection \%} \\
\hline
\hline
Raw detection                          & $5.33$ & $8.92$ & $3.91$ & $98.04$ \\
\hline
Detection + UKF                        & $5.42$ & $7.69$ & $3.00$ & $100.00$ \\
\hline
\textbf{Detection + gate + UKF (ours)} & $\mathbf{5.05}$ & $\mathbf{5.46}$ & $\mathbf{1.93}$ & $\mathbf{100.00}$ \\
\hline
\end{tabular}
}
\end{table}

\subsection{Occlusion in a Changed Background}
\label{ssec:occ_clutter}
 
The two preceding subsections separate two findings. \autoref{ssec:inpaint_ablation} shows that against the controlled backdrop inpainting handles occlusion most accurately, and \autoref{ssec:bkgrnd_robustness} shows that the clean detector transfers to a cluttered background without retraining. The remaining question is whether the occlusion handling itself transfers to that changed background. We recorded five additional sequences that introduce partial occlusion into the same cluttered laboratory scene of \autoref{ssec:bkgrnd_robustness}, scored the five control keypoints against the clean-detector reference following the protocol of \autoref{ssec:inpaint_ablation}, and compared four pipelines: the clean detector with the gate and the UKF without inpainting, our inpainting pipeline, a detector trained on artificially occluded images, and a detector trained on a mixture of occluded and unoccluded images, archived on Zenodo \textit{(DOI: 10.5281/zenodo.20979031)}. \autoref{tab:occ_clutter} reports the result.
 
\begin{figure*}[!t]
    \centering
    \includegraphics[width=\textwidth]{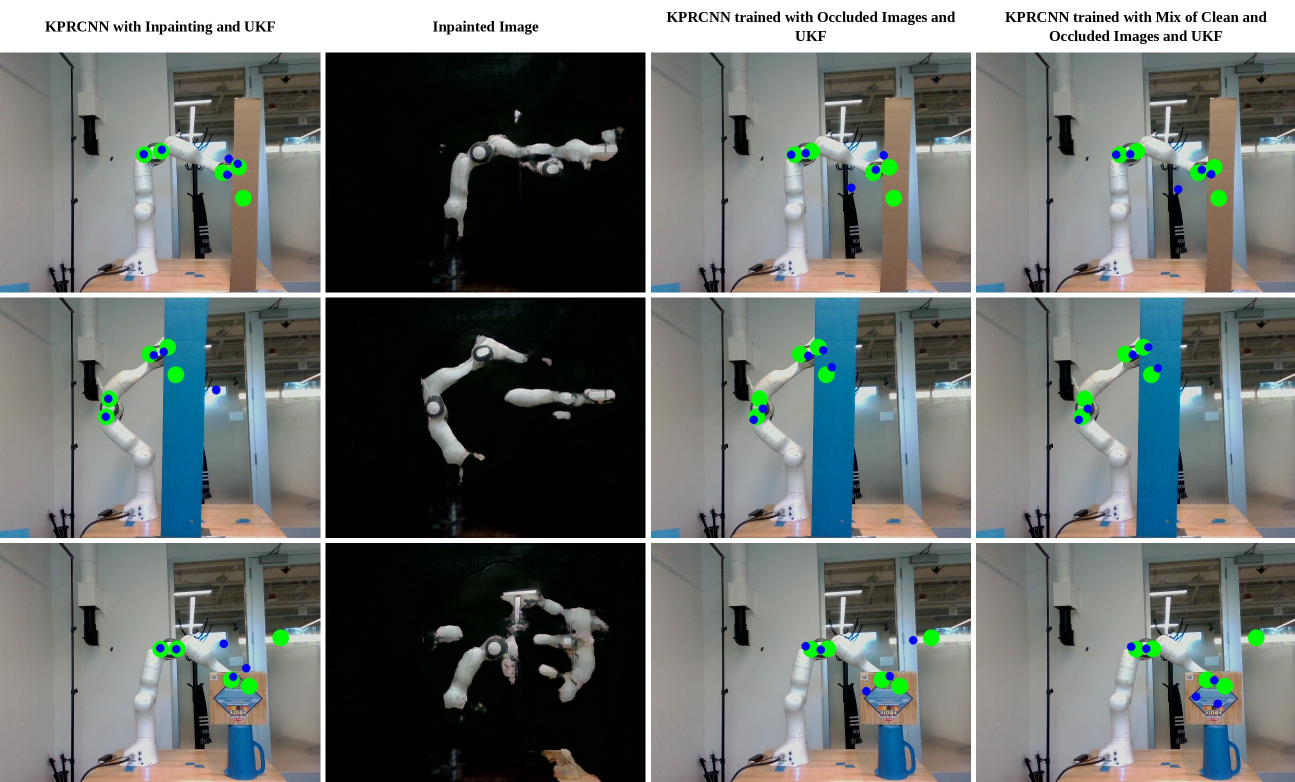}
    \caption{
    Qualitative behavior under combined partial occlusion and background clutter, for three sequences (rows). The columns are, from left to right, the proposed pipeline (clean detector with inpainting, the outlier gate, and the UKF), the inpainted reconstruction that pipeline produces, the occlusion-trained detector, and the mixed-trained detector. In the three overlay columns the green circle is the ground truth, obtained from the clean detector on the unoccluded frame, and the blue circle is the method's prediction. The reconstruction column shows that, against a background the generator never saw, the inpainting stage repaints the robot only inside the suppressed region and does not restore the occluded structure, so the predictions for the occluded keypoints drift from the ground truth for all three methods. The keypoints that remain visible are still localized correctly.
    }
    \label{fig:occ_clutter_qualitative}
\end{figure*}

Inpainting no longer contributes in this regime. Against the controlled backdrop, switching inpainting on reduced the mean error from $11.92$ px to $2.82$ px (\autoref{tab:inpaint_ablation}). The same on-off comparison in the changed background moves the mean error only from $26.55$ px to $26.00$ px, because the generator was trained to reconstruct the robot against the controlled backdrop and carries no prior for the cluttered scene, so it cannot recover a keypoint that is both occluded and set against an unfamiliar background. \autoref{fig:occ_clutter_qualitative} shows this qualitatively, where the inpainted reconstruction restores the robot only inside the suppressed region and the predictions of all three methods drift from the ground truth on the occluded keypoints. The two detectors retrained for occlusion do not transfer either. Their mean error is lower, since they localize the partially visible keypoints more evenly across the body, but they fail on roughly $70\%$ of the keypoints in this regime, against about $31\%$ for the inpainting and no-inpainting pipelines, and neither was trained against clutter. Taken together with \autoref{ssec:bkgrnd_robustness}, these results separate two properties of the pipeline. The clean detector generalizes across backgrounds on the keypoints that remain visible, while occlusion handling, whether by reconstructing the image or by retraining the detector, stays tied to the background it was trained on. Occlusion in a changed background therefore remains an open problem, which we return to in \autoref{sec:conclusion}. See Multimedia Extension $11$ for the experimental videos.
 
\begin{table}[t]
\caption{Keypoint estimation under combined partial occlusion and background clutter, averaged across five sequences and scored on the five control keypoints against the clean-detector reference. The first two pipelines share the clean detector with the outlier gate and the UKF and differ only in whether inpainting is applied. All four are evaluated against a background none of them was trained on.}
\label{tab:occ_clutter}
\centering
\resizebox{0.5\textwidth}{!}{%
\begin{tabular}{|>{\centering\arraybackslash}p{3.4cm}||>{\centering\arraybackslash}p{1.8cm}|>{\centering\arraybackslash}p{1.8cm}|>{\centering\arraybackslash}p{1.8cm}|>{\centering\arraybackslash}p{1.8cm}|}
\hline
\textbf{Pipeline} & \textbf{Mean Error (px)} & \textbf{RMSE (px)} & \textbf{Failure Rate \%} & \textbf{Detection \%} \\
\hline
\hline
Detection + UKF (no inpainting) & $26.55$ & $53.05$ & $30.90$ & $100.00$ \\
\hline
Inpainting + UKF (ours)         & $26.00$ & $49.06$ & $30.83$ & $100.00$ \\
\hline
Occlusion-trained + UKF         & $22.66$ & $32.23$ & $71.46$ & $100.00$ \\
\hline
Mixed-trained + UKF             & $21.09$ & $30.05$ & $70.02$ & $100.00$ \\
\hline
\end{tabular}
}
\end{table}

\subsection{Transient Response - $2$ Active Planar Joints}
\begin{figure}[!t]
          \centering
      \includegraphics[width=0.5\textwidth]{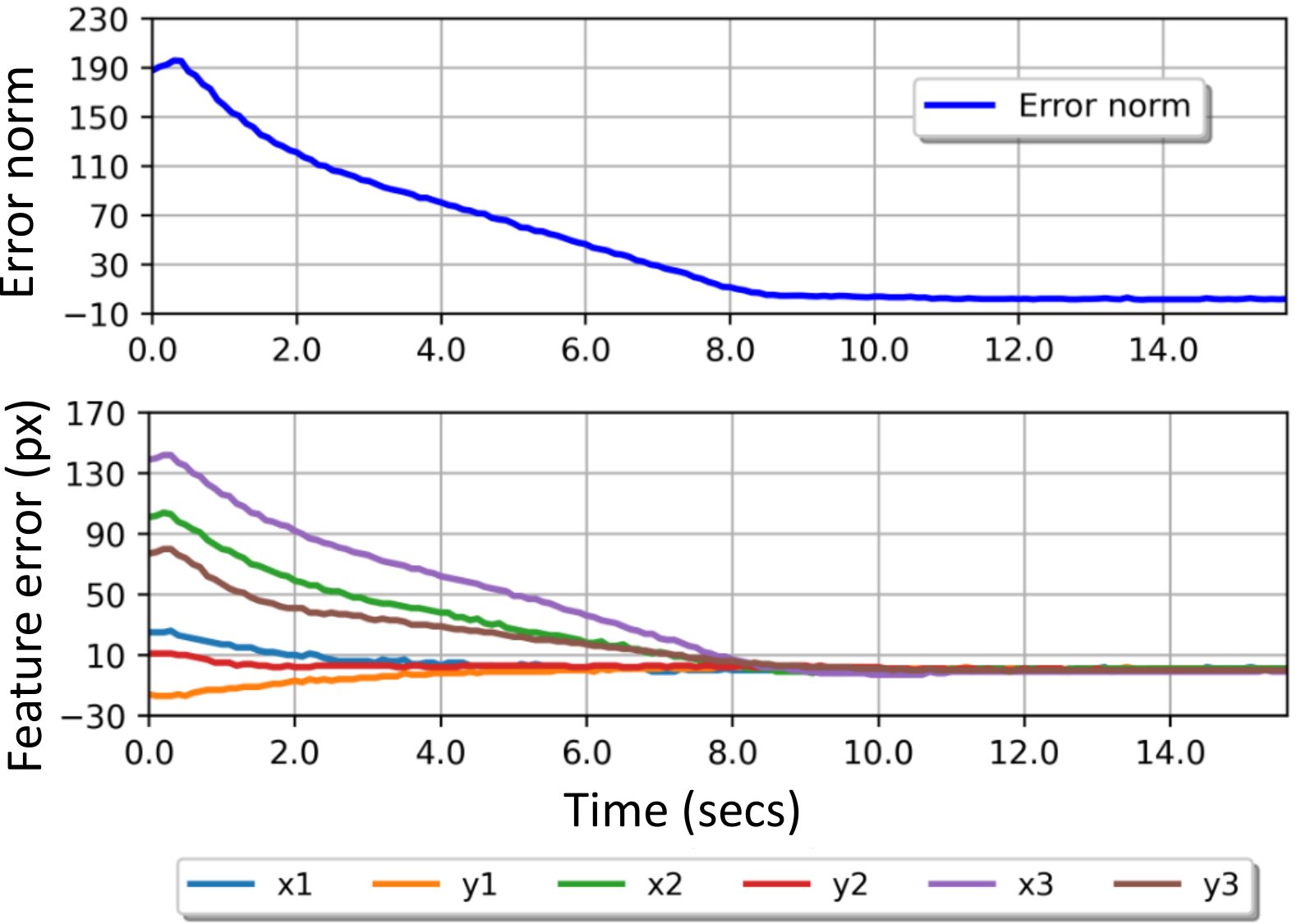}
      \caption{Image feature error norm (top) and individual image feature errors (bottom) for $3$ keypoints. }
      \label{fig:error_plot_03}    
\end{figure}

\begin{figure}[!t]
    \centering
    \vspace{5pt}
    \begin{tabular}{|c|c|c|}
        \hline
         & Prior work & This paper \\
         \hline
         \raisebox{-0.25\height}{\rotatebox[origin=c]{90}{\strut Target 1}} &
            \begin{minipage}[c]{0.20\textwidth}
                \centering
                \includegraphics[width=\linewidth]{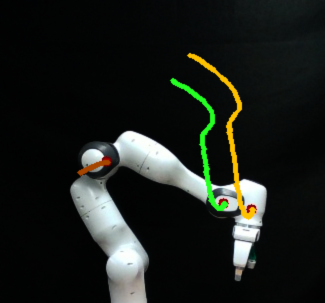}
            \end{minipage} &
            \begin{minipage}[c]{0.20\textwidth}
                \centering
                \includegraphics[width=\linewidth]{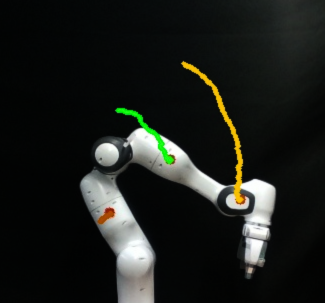}
            \end{minipage} \\
        \hline
        \raisebox{-0.15\height}{\rotatebox[origin=c]{90}{\strut Target 2}} &
            \begin{minipage}[c]{0.20\textwidth}
                \centering
                \includegraphics[width=\linewidth]{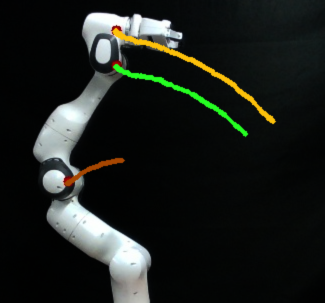}
            \end{minipage} &
            \begin{minipage}[c]{0.20\textwidth}
                \centering
                \includegraphics[width=\linewidth]{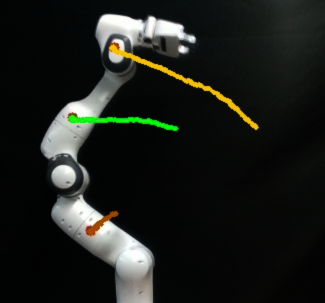}
            \end{minipage} \\ 
        \hline
    \end{tabular}
    \caption{Comparison of trajectory smoothness between the proposed and prior methods. The low-noise profiles indicate that our method achieves robustness comparable to the existing approach.}
    \label{fig:traj_comp_02}
\end{figure}

In order to showcase the utility of our proposed approach, we conducted $20$ vision-based control experiments by utilizing the trained keypoint detection model in real time using $3$ keypoints to control $2$ planar joints, i.e. the second and fourth joints of the Panda Arm we are using. We used exactly the same reference configurations and adaptive visual servoing scheme to our prior work \citep{chatterjee2023keypoints} (which requires camera calibration, robot model, and encoder readings to train the keypoint detection model) so that we can compare the control performance. We tuned the control gains for the system to the lowest rise time and also made sure the overshoot remains within 5\%. We used an Intel RealSense D435i camera. \autoref{performance_data_02} depicts that the visual features generated with our current method are as reliable and efficient as those from the prior keypoint-based method. 
\autoref{fig:error_plot_03} shows the error norm and the individual feature error plot for one of the $20$ control experiments conducted. Please note in \autoref{fig:traj_comp_02}, while the reference configurations are the same between the two experiments, the feature locations are different, since our prior work is trained to detect the joint centers, while the current work detects the locations where the markers were placed in the training phase. Sample experiments of our current method controlling $2$ planar joints can be found in Multimedia Extension $2$.

\begin{table}[!t]
\caption{Performance comparison summary}
\label{performance_data_02}
\centering
\resizebox{0.4\textwidth}{!}{%
\begin{tabular}{
|>{\centering\arraybackslash}m{2cm}||
 >{\centering\arraybackslash}m{2cm}|
 >{\centering\arraybackslash}m{2cm}|
}
\hline
\textbf{Performance metrics} & \textbf{Prior Method} & \textbf{Proposed algorithm}\\
\hline
\hline
\textbf{System rise time (s)} & $6.16 \pm 2.10$ & $5.85 \pm 2.9$ \\
\hline
\textbf{System settling time (s)} & $11.29 \pm 2.42$ & $8.31 \pm 2.9$ \\
\hline
\textbf{End effector rise time (s)} & $5.37 \pm 1.66$ & $5.44 \pm 2.65$  \\
\hline
\textbf{End effector settling time (s)} & $10.05 \pm 3.48$ & $7.72 \pm 3.1$ \\
\hline
{\textbf{Overshoot (\%)}} & $2.9 \pm 2.73$ & $3.43 \pm 3.28$\\
\hline
\end{tabular}
}
\end{table}

\begin{table}[!t]
\caption{Performance results for the control experiments with $5$ keypoints and $3$ joints, with and without occlusion}
\label{performance_data_03}
\centering
\resizebox{0.5\textwidth}{!}{%
\begin{tabular}{
|>{\centering\arraybackslash}m{2cm}||
 >{\centering\arraybackslash}m{2cm}|
 >{\centering\arraybackslash}m{2cm}|
 >{\centering\arraybackslash}m{2cm}|
}
\hline
\textbf{Performance metrics} & \textbf{Baseline keypoint detection (No occlusion)} & \textbf{Small occlusion} & \textbf{Large occlusion} \\
\hline
\hline
\textbf{System rise time (s)} & $10.68 \pm 5.82$ & $13.94 \pm 3.72$ & $14.18 \pm 6.22$ \\
\hline
\textbf{System settling time (s)} & $15.04 \pm 7.19$ & $19.66 \pm 5.85$ & $24.08 \pm 11.49$  \\
\hline
\textbf{End effector rise time (s)} & $8.56 \pm 5.87$ & $9.02 \pm 5.17$ & $12.83 \pm 6.79$ \\
\hline
\textbf{End effector settling time (s)} & $14.02 \pm 6.83$ & $16.66 \pm 8.12$ & $18.98 \pm 7.08$\\
\hline
\textbf{Overshoot (\%)} & $2.33 \pm 1.03$ & $4.11 \pm 2.08$ & $11.37 \pm 9.88$ \\
\hline
\end{tabular}
}
\end{table}

\begin{figure}[!t]
      \centering
      \includegraphics[width=0.5\textwidth]{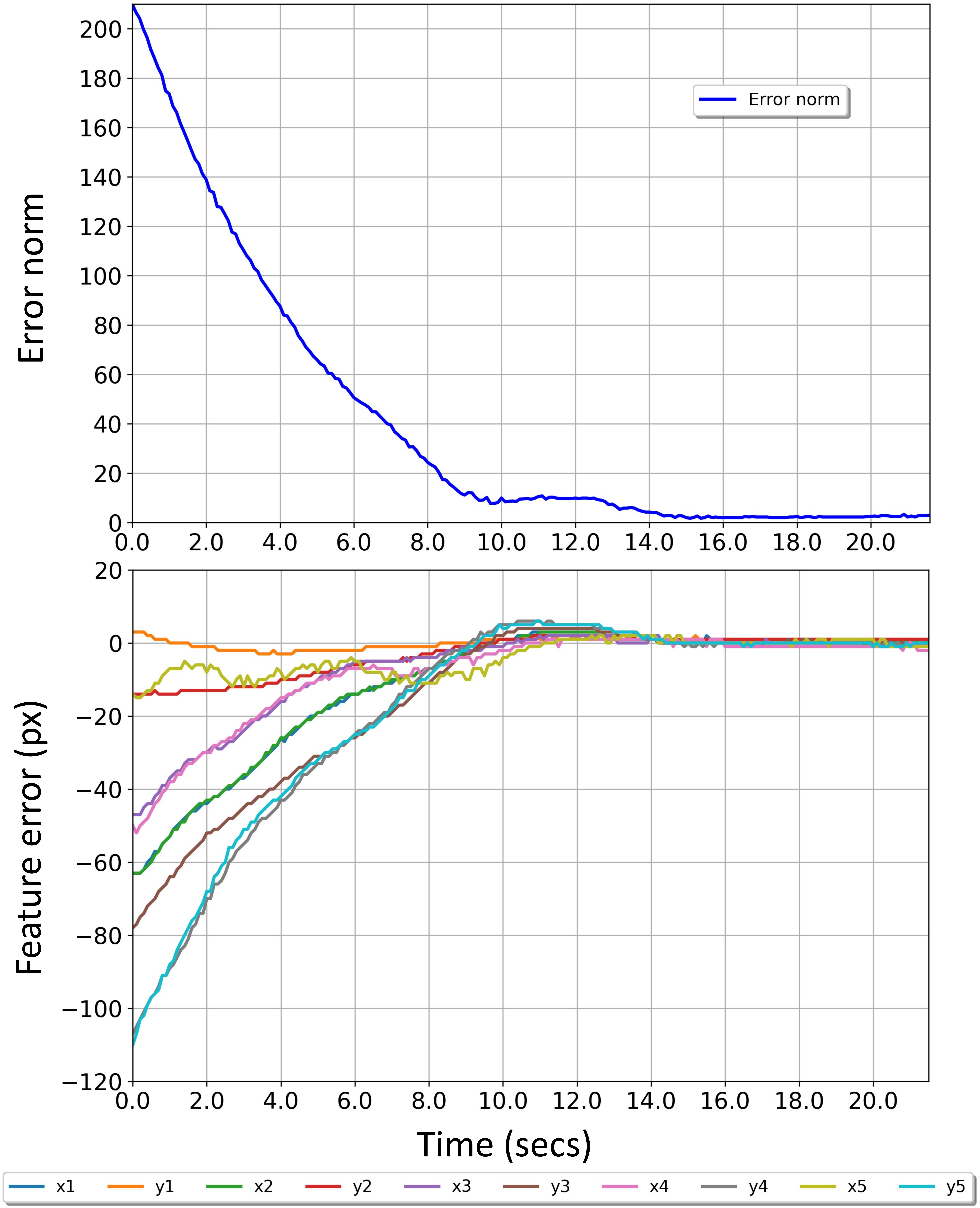}
      \caption{Image feature error norm (top) and individual image feature errors (bottom) for $5$ keypoints without occlusion in the scene. }
      \label{fig:error_plot_05_no_occ}
\end{figure}

\begin{figure}[!t]
      \centering
      \includegraphics[width=0.45\textwidth]{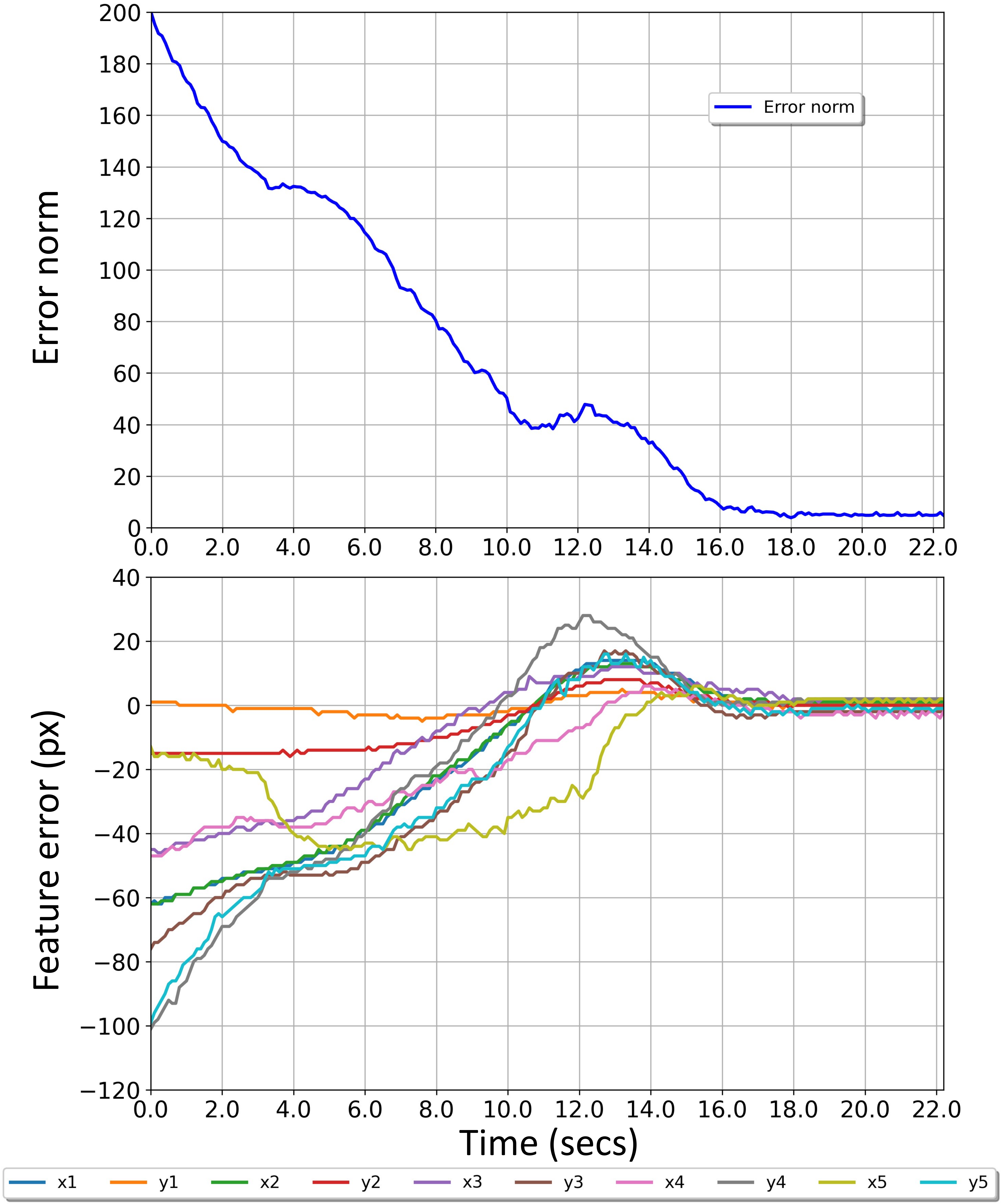}
      \caption{Image feature error norm (top) and individual image feature errors (bottom) for $5$ keypoints with occlusion in the scene. }
      \label{fig:error_plot_05_occ}
\end{figure}

\begin{figure*}[!t]
  \centering

  \includegraphics[width=0.83\textwidth]{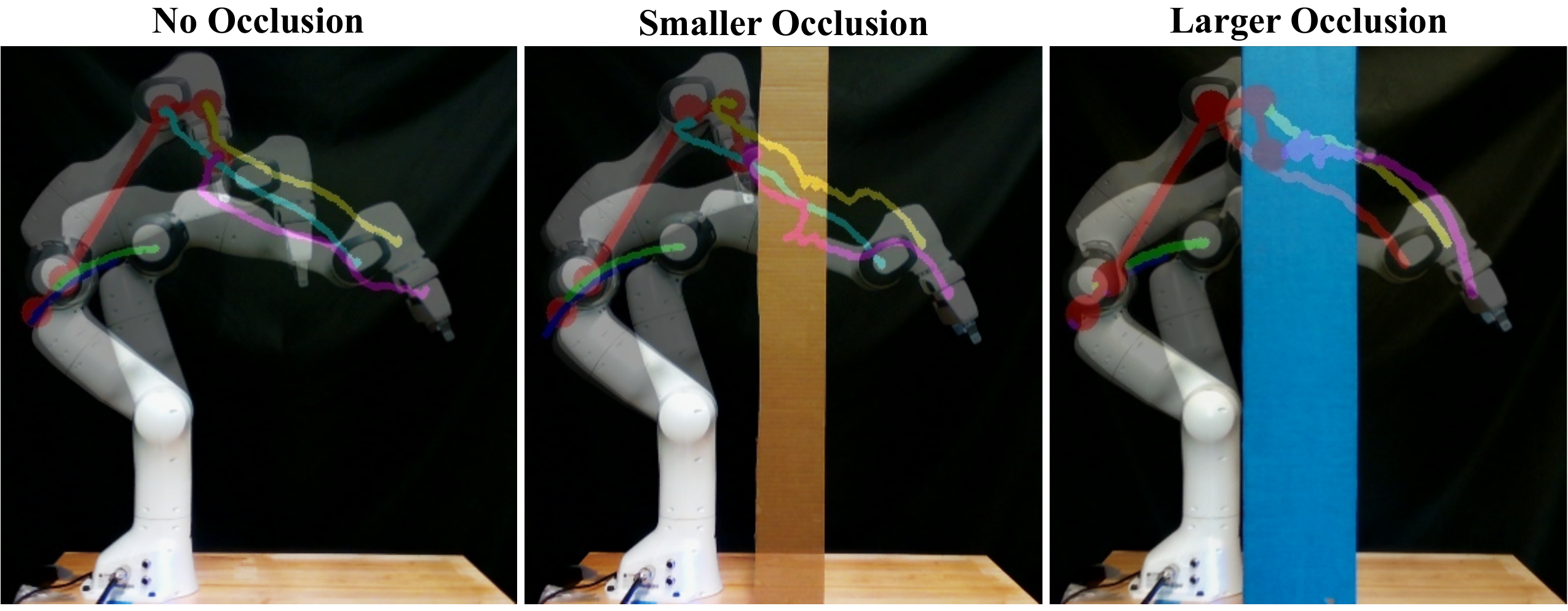}

  \includegraphics[width=0.83\textwidth]{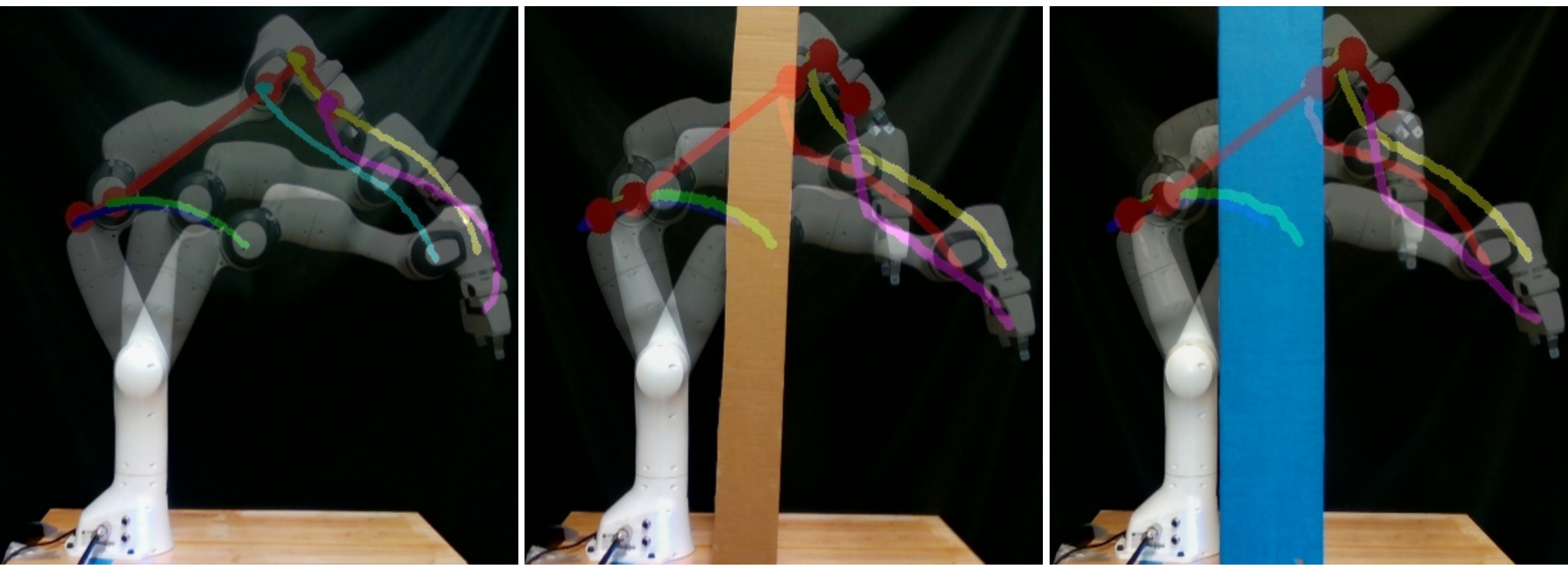}

  \includegraphics[width=0.83\textwidth]{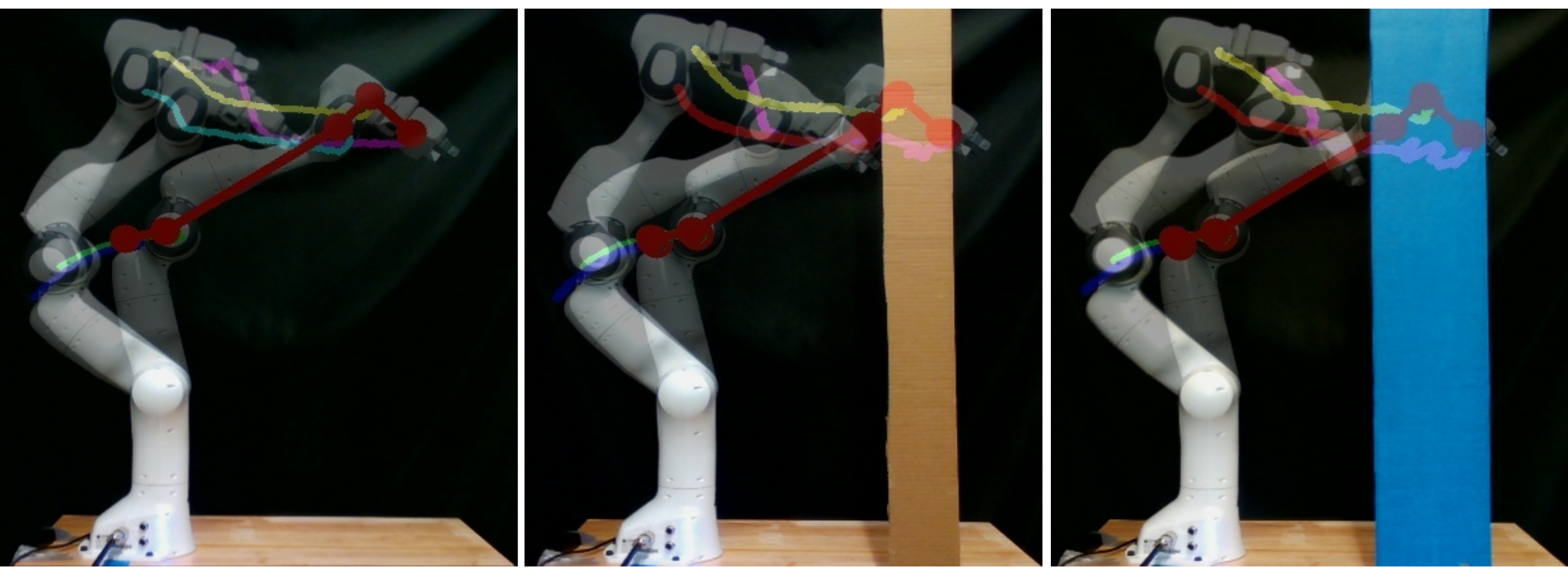}

  \includegraphics[width=0.83\textwidth]{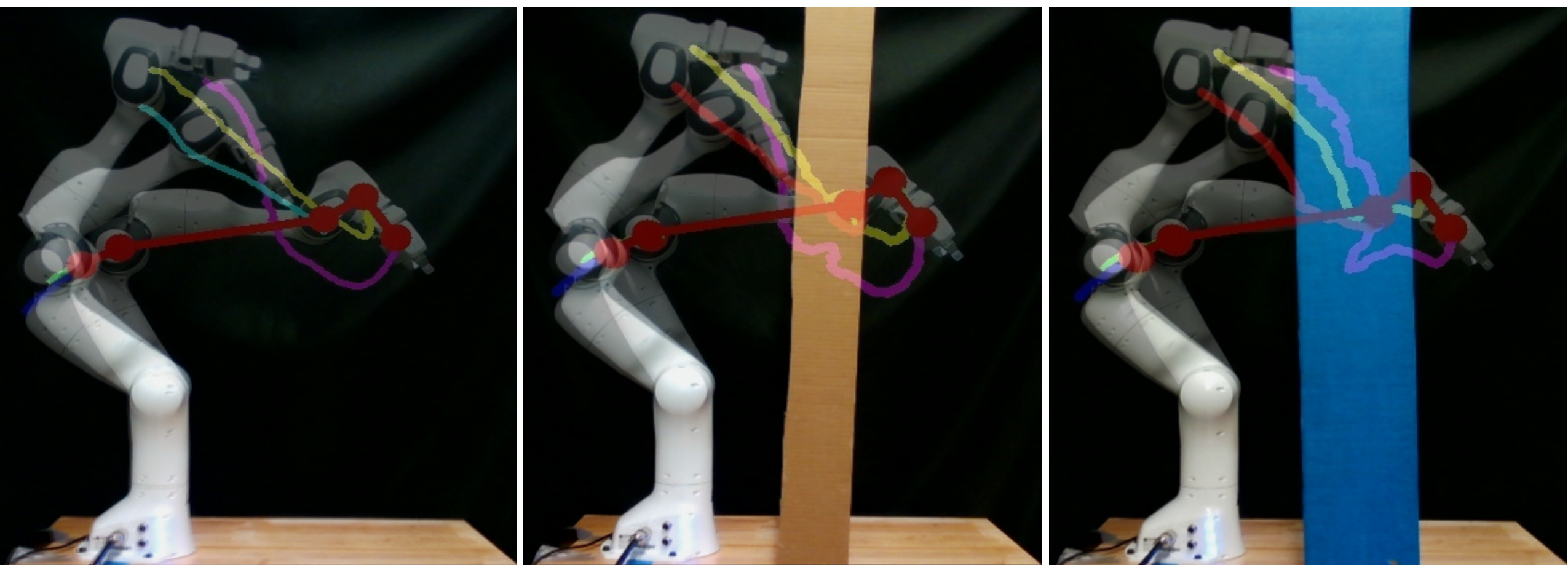}
  
  \vspace{-5pt}

  \caption{Control experiment results comparing baseline keypoint detection with keypoint detection under varying levels of occlusion. Each row corresponds to the same start and goal configuration. The first image in each row shows the baseline without occlusion, the second includes a smaller occlusion, and the third includes a larger occlusion. The robot configuration with red keypoints represents the goal. While the controller successfully converges in all cases, the presence of occlusion leads to visibly noisier trajectories.}
  \label{fig:stacked_control_results}
\end{figure*}

\begin{table}[!t]
\caption{Performance summary for the control experiments in out-of-plane motion}
\label{performance_data_3d}
\begin{center}
\begin{tabular}{|c||c|}
\hline
\textbf{Performance metrics} & \textbf{Proposed algorithm}\\
\hline
\hline
\textbf{System rise time (s)} & $17.69 \pm 10.77$ \\
\hline
\textbf{System settling time (s)} & $34.39 \pm 17.57$ \\
\hline
\textbf{End effector rise time (s)} & $14.19 \pm 8.62$ \\
\hline
\textbf{End effector settling time (s)} & $27.12 \pm 15.6$ \\
\hline
{\textbf{Overshoot (\%)}} & $3.21 \pm 2.32$ \\
\hline
\end{tabular}
\end{center}
\end{table}

\begin{figure*}[!t]
      \centering
      \includegraphics[width=0.75\textwidth]{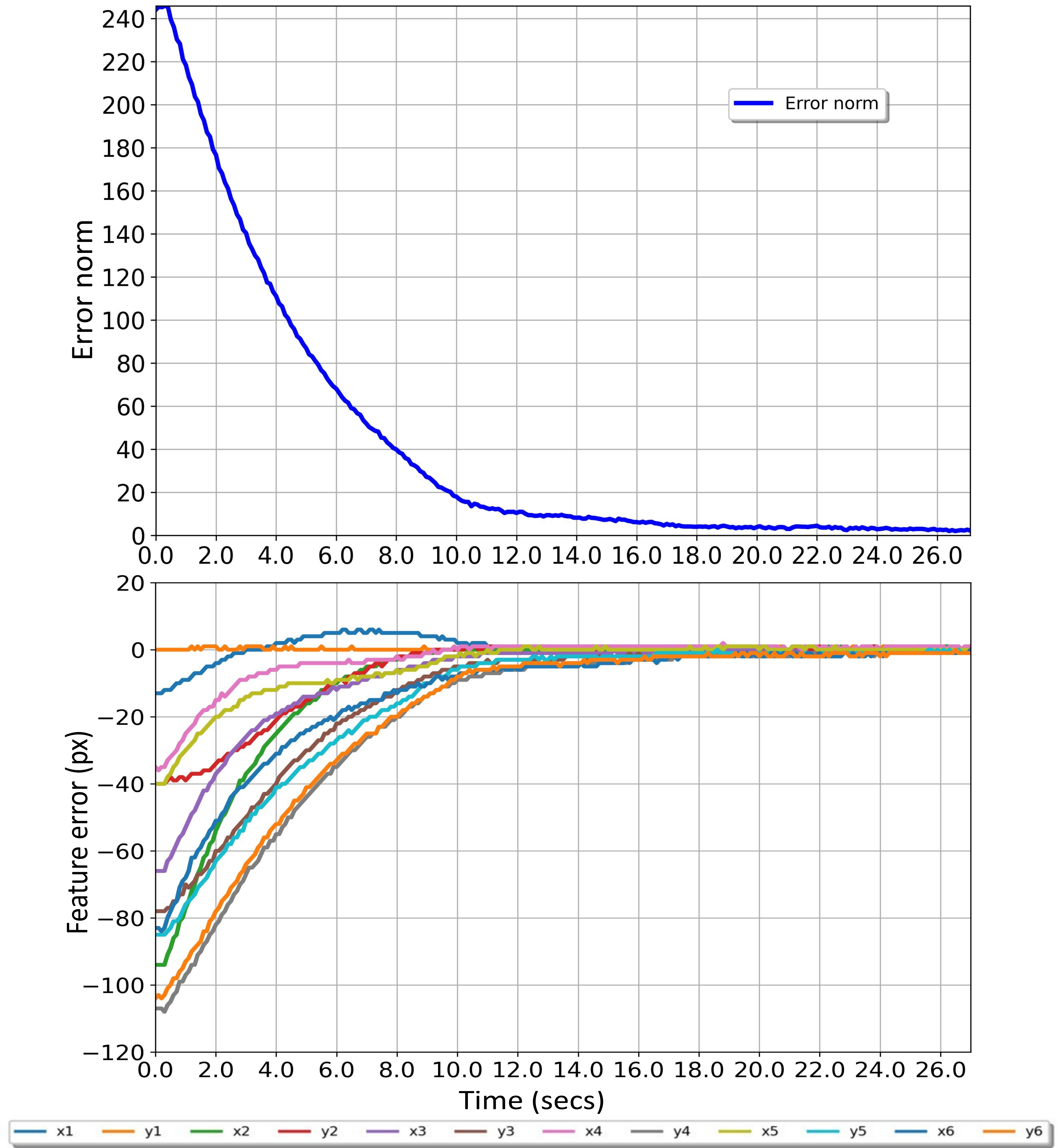}
      \caption{Image feature error norm (top) and individual image feature errors (bottom) for control experiments with 3D motion.}
      \label{fig:error_plot_3d}
\end{figure*}

\begin{figure*}[!t]
  \centering

  \includegraphics[width=0.8\textwidth]{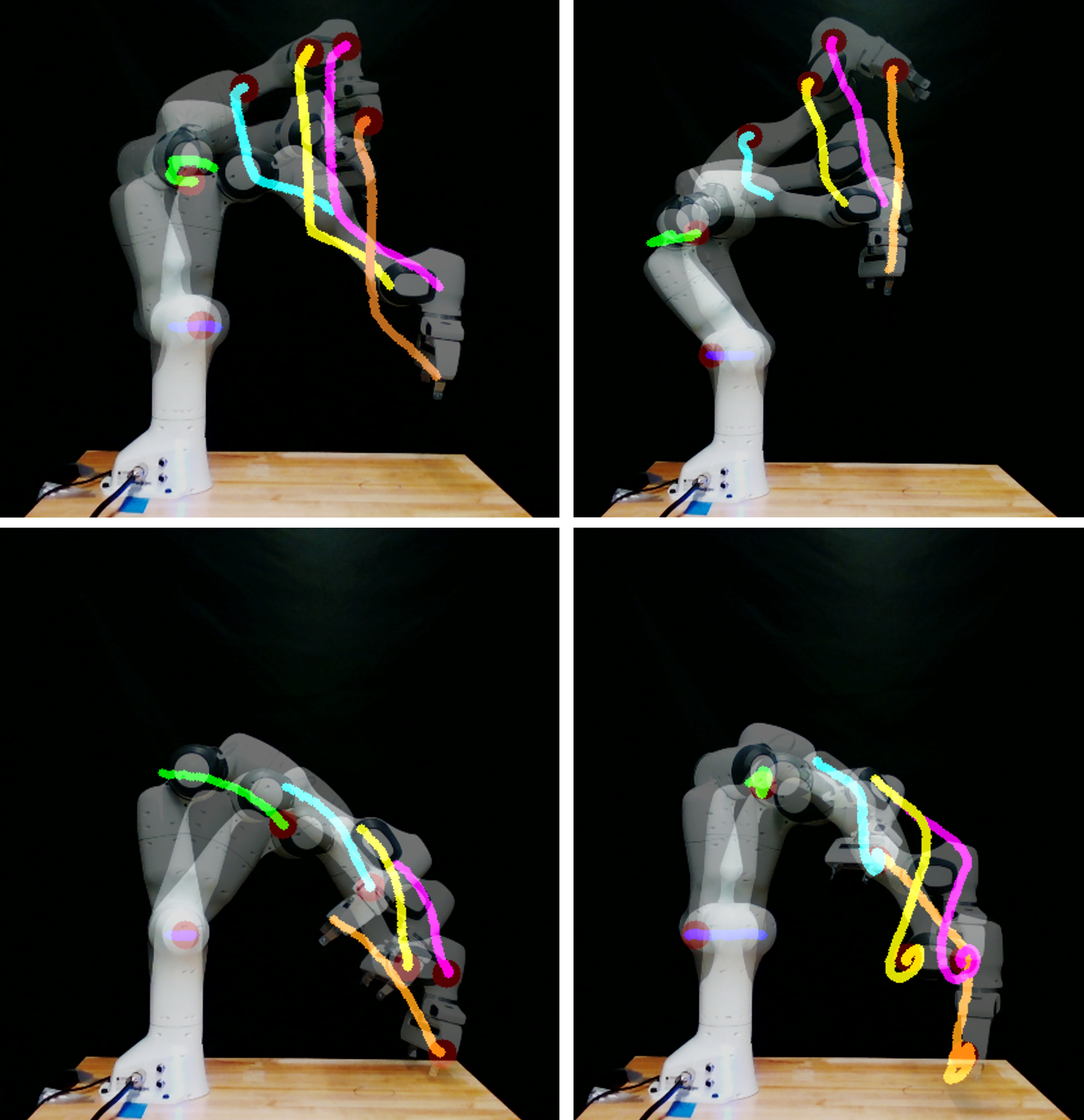}

  \caption{Control experiment results for out-of-plane motions using $6$ keypoints and $4$ joints. The robot configurations with the red keypoints represent the goal. As observed, the controller achieves smooth and noiseless trajectories in all cases.}
  \label{fig:stacked_control_results_3d}
\end{figure*}

\begin{figure*}[!t]
      \centering
      \includegraphics[width=0.75\textwidth]{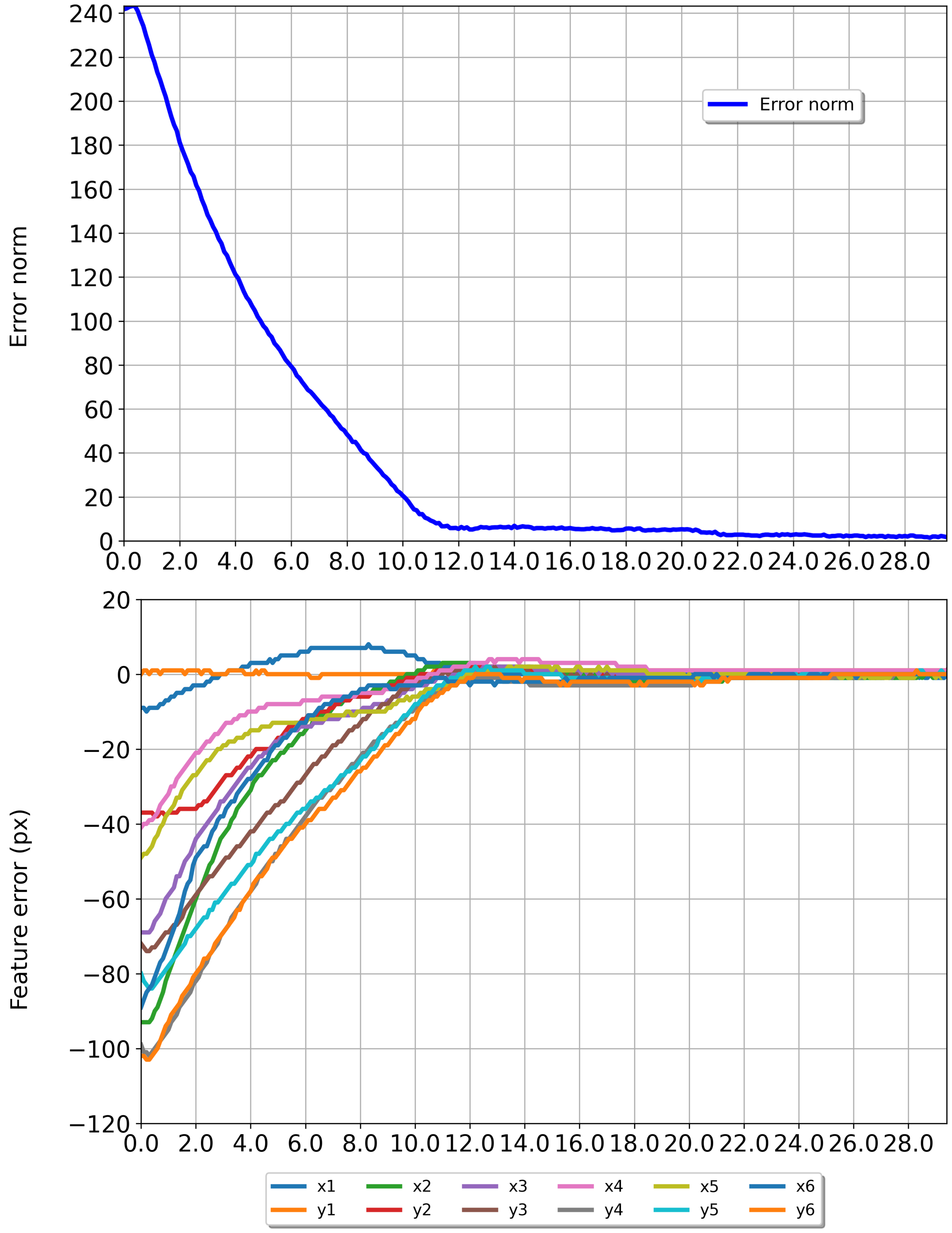}
      \caption{Image feature error norm (top) and individual image feature errors (bottom) for control experiments with 3D motion in clutter.}
      \label{fig:error_plot_clutter}
\end{figure*}

\begin{figure*}[!t]
  \centering
  \includegraphics[width=0.32\textwidth]{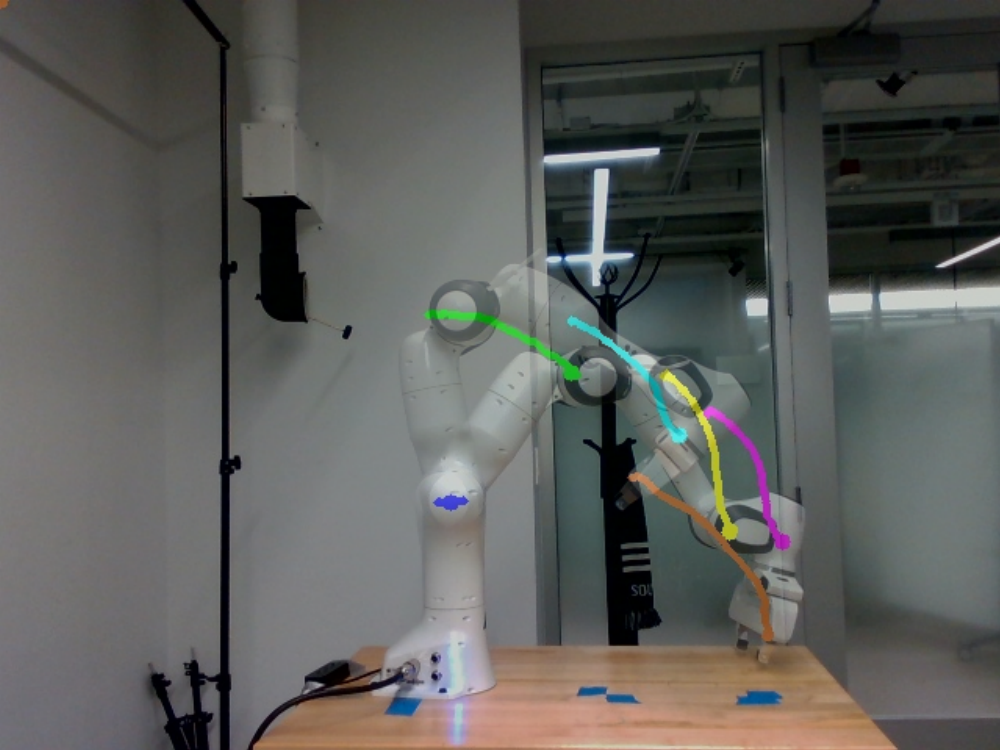}\hfill
  \includegraphics[width=0.32\textwidth]{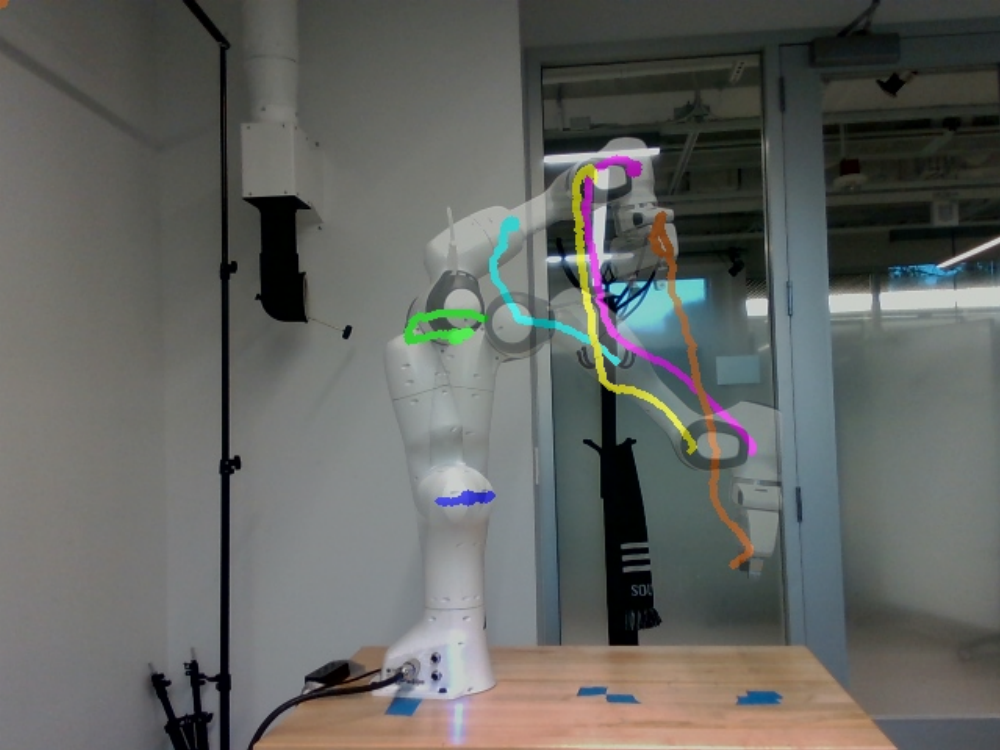}\hfill
  \includegraphics[width=0.32\textwidth]{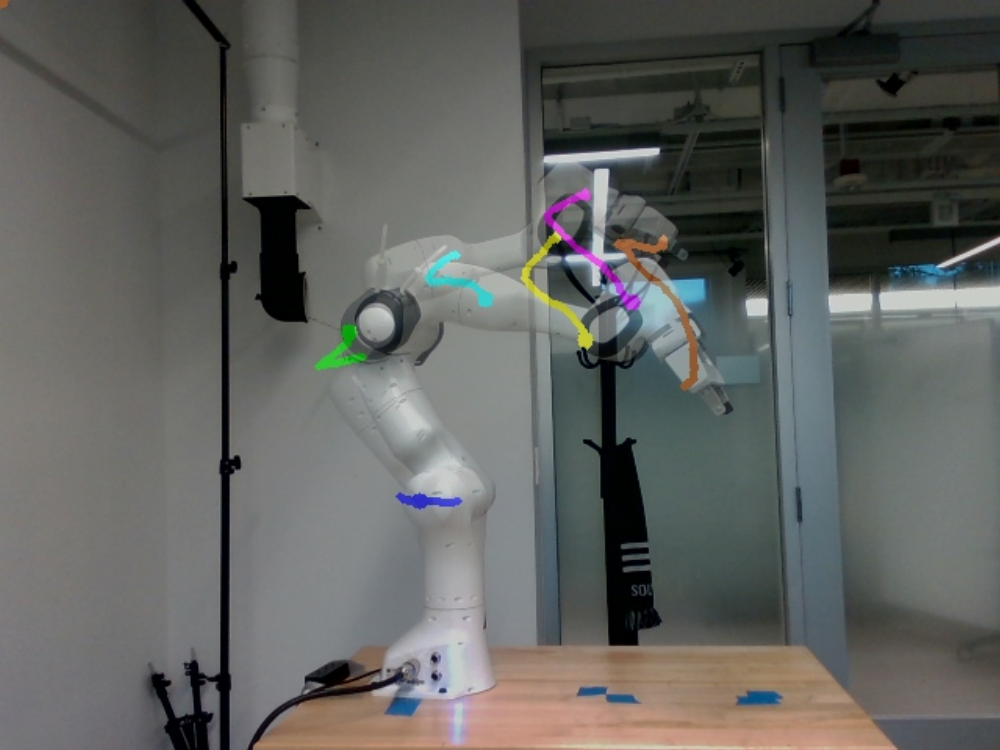}
  \caption{Control experiment results for out-of-plane motions using $6$ keypoints and $4$ joints, with clutter background. In each panel the faded robot denotes the start configuration and the solid robot the goal configuration, and the colored curves trace the image-space trajectory of each keypoint from start to goal. As observed, the controller achieves smooth and noiseless trajectories in all three cases.}
  \label{fig:clutter_control_3d}
\end{figure*}

\subsection{Transient Response - $3$ Active Planar Joints}
To further validate the effectiveness of our proposed pipeline, we conducted $3$ sets of $8$ vision-based control experiments using $5$ visual keypoints to control $3$ planar joints, specifically, the second, fourth, and sixth joints of the Panda arm. Each experiment involved a distinct start–goal configuration pair.

The first set was performed under full visibility and serves as a baseline reference. In the remaining two sets, we introduced visual occlusions to evaluate robustness under partial observability. One set contained large occlusions that covered significant portions of the robot body, while the other involved smaller occlusions allowing us to evaluate the robustness of our system under varying levels of visual obstruction. 

We tuned the control parameters for all three systems for the best performance. The corresponding control experiment videos are provided in Multimedia Extension $3$ (full visibility), Multimedia Extension $4$ (small occlusions), and Multimedia Extension $5$ (large occlusions).

\autoref{performance_data_03} demonstrates that our proposed pipeline maintains comparable rise and settling times under occlusion, though large occlusions result in higher average overshoot with high variance, suggesting occasional but not consistently severe control deviations. 
The table values were computed from the archived control–experiment datasets on Zenodo: baseline (Column~1, \textit{DOI: 10.5281/zenodo.17334885}), small occlusion (Column~2, \textit{DOI: 10.5281/zenodo.17334903}), and large occlusion (Column~3, \textit{DOI: 10.5281/zenodo.17334917}).

\autoref{fig:error_plot_05_no_occ} shows the error norm and the individual feature error plot for one of the $8$ control experiments conducted for scenarios without occlusions, and the \autoref{fig:error_plot_05_occ} shows the same plots for one of the $16$ control experiments conducted for scenarios with occlusion in the visible workspace.

\autoref{fig:stacked_control_results} provides a qualitative comparison of control trajectories under three conditions: baseline keypoint detection (keypoint detection without occlusion), keypoint detection with smaller occlusion, and keypoint detection with larger occlusion in the image. The controller successfully converges in all three cases, though the trajectories become increasingly noisier and less optimal as the level of occlusion increases. See Multimedia Extension $6$ for side-by-side videos of each condition.

We have observed two key outcomes in the control performance experiments. First, the transient response for the $3$-joint configuration, shown in \autoref{performance_data_03} (Column 1), is comparatively slower than that of the $2$-joint configuration in \autoref{performance_data_02} (Column 2). This is primarily because the third joint introduced in the $3$-joint setup has a limited range of motion and contributes minimally to the overall variation in visual features, reducing its influence on control effectiveness.

Second, the average overshoot is noticeably higher in scenarios with larger occlusions, primarily because a greater number of keypoints are obscured, making it more difficult for the system to converge smoothly.

\subsection{Transient Response - 3D Motion}
\vspace{-0.5em}
To evaluate out-of-plane behavior, we conducted $10$ vision-based control experiments using the $6$-keypoints to control $4$ joints - $3$ planar joints (second, fourth, and sixth) together with the first spatial joint of the Franka Emika Panda arm under the same eye-to-hand setup and adaptive visual servoing scheme as the planar studies. Controller gains were tuned with the same procedure as before to balance fast rise time with limited overshoot. As summarized in \autoref{performance_data_3d}, the system exhibits consistent stable convergence despite the added depth variation and viewpoint changes inherent to out-of-plane motion. \autoref{fig:error_plot_3d} shows the error norm and the individual feature error plot for one of the $10$ control experiments conducted. The table values were computed using the control experiments data archived on Zenodo \textit{(DOI: 10.5281/zenodo.17335046)}. \autoref{fig:stacked_control_results_3d} shows that the controller maintains smooth trajectories even under significant depth variation. See Multimedia Extension $7$ for all the 3D control experiment videos.

\subsection{Out-of-Plane Control in a Cluttered Background}
\label{ssec:clutter_control}

The background-robustness study of \autoref{ssec:bkgrnd_robustness} evaluates keypoint detection in the cluttered laboratory scene. We now close the control loop in that same scene to test whether the perception transfer carries through to control. We ran five out-of-plane control experiments in the cluttered background using the same six-keypoint, four-joint setup and adaptive visual servoing scheme as the controlled-background runs of \autoref{performance_data_3d}, each driven to a distinct start and goal configuration. The detector is the deployed model trained only on the black-background data, with no retraining for this scene. \autoref{performance_data_3d_clutter} reports the transient response averaged over the five runs. \autoref{fig:error_plot_clutter} shows the error norm and the individual feature error plot for one of the $5$ control experiments conducted. The table values were computed using the control experiments data archived on Zenodo \textit{(DOI: 10.5281/zenodo.20978321)}. See Multimedia Extension $12$ for all the control experiments.

The controller reaches the goal in every run. \autoref{fig:clutter_control_3d} overlays the start and goal configurations for three representative runs, with the colored curves tracing each keypoint from start to goal, and the trajectories remain smooth even though the detector runs against a background it never saw during training. The mean system settling time is $37.84$ s and the end-effector settling time is $29.80$ s, in the same range as the $34.39$ s and $27.12$ s of the controlled-background case. The start and goal pairs differ between the two sets, so the absolute times are not a controlled comparison, but the transient behavior is of the same order. The mean overshoot is higher at $6.38\%$ against $3.21\%$, consistent with the additional detection noise that the cluttered background introduces. These experiments show that the perception transfer reported in \autoref{ssec:bkgrnd_robustness} extends to closed-loop control, since the system completes the out-of-plane task against a previously unseen background without retraining.

\begin{table}[!t]
\caption{Performance summary for the out-of-plane control experiments conducted in the cluttered laboratory background, averaged over five runs.}
\label{performance_data_3d_clutter}
\begin{center}
\begin{tabular}{|c||c|}
\hline
\textbf{Performance metrics} & \textbf{Proposed algorithm}\\
\hline
\hline
\textbf{System rise time (s)} & $13.94 \pm 8.36$ \\
\hline
\textbf{System settling time (s)} & $37.84 \pm 20.10$ \\
\hline
\textbf{End effector rise time (s)} & $9.94 \pm 4.02$ \\
\hline
\textbf{End effector settling time (s)} & $29.80 \pm 17.13$ \\
\hline
{\textbf{Overshoot (\%)}} & $6.38 \pm 2.87$ \\
\hline
\end{tabular}
\end{center}
\end{table}

\begin{table}[t]
\centering
\caption{Direct evaluation of the online interaction-matrix estimate $\hat{J}$ across
five out-of-plane control runs. $\kappa(\hat{J})$ is the condition number of the estimate.
The direction agreement is the share of steps in which the predicted feature-motion
direction lies within ninety degrees of the observed direction, and the cosine is the
median over the run, both at a ten-step horizon.}
\label{tab:jacobian_eval}
\resizebox{0.5\textwidth}{!}{%
\begin{tabular}{|c||c|c|c|c|}
\hline
\textbf{Run} & \textbf{Steps} & $\boldsymbol{\kappa(\hat{J})}$ \textbf{median} & \textbf{Within} $\boldsymbol{90^\circ}$ & \textbf{Cosine} \\
\hline
\hline
1 & 224 & 10.9 & 91.0\% & 0.88 \\
\hline
2 & 175 & 14.4 & 100.0\% & 0.94 \\
\hline
3 & 293 & 9.2 & 100.0\% & 0.89 \\
\hline
4 & 687 & 11.2 & 99.8\% & 0.74 \\
\hline
5 & 505 & 10.5 & 97.7\% & 0.72 \\
\hline
\hline
\textbf{Mean} & & \textbf{11.2} & \textbf{97.7\%} & \textbf{0.83} \\
\hline
\end{tabular}
}
\end{table}

\begin{table}[!t]
\caption{Summary of repeatability tests - $2$ Joints and $3$ Keypoints }
\label{repeat_data_02}
\begin{center}
\begin{tabular}{|c||c|}
\hline
& {\bf Proposed algorithm}\\
\hline
\hline
\bf{System rise time (s)} & $6.82 \pm 0.11$\\
\hline
\bf{System settling time} (s) & $9.12 \pm 0.23$\\
\hline
\bf{End effector rise time} (s) & $6.82 \pm 0.11$\\
\hline
\bf{End effector settling time} (s) & $9.0 \pm 0.1$\\
\hline
{\textbf{Overshoot (\%)}} & $1.096 \pm 0.27$\\
\hline
% Steady state error (px) & - & -\\
% \hline
\end{tabular}
\end{center}
\end{table}

\begin{table}[!t]
\caption{Summary of repeatability tests - $3$ Joints, $5$ Keypoints, No occlusion}
\label{repeat_data_03}
\begin{center}
\begin{tabular}{|c||c|}
\hline
& {\bf Proposed algorithm}\\
\hline
\hline
\bf{System rise time (s)} & $9.68 \pm 0.35$\\
\hline
\bf{System settling time} (s) & $19.72 \pm 0.19$\\
\hline
\bf{End effector rise time} (s) & $7.42 \pm 0.41$\\
\hline
\bf{End effector settling time} (s) & $19.72 \pm 0.19$\\
\hline
{\textbf{Overshoot (\%)}} & $3.99 \pm 0.68$\\
\hline

\end{tabular}
\end{center}
\end{table}

\begin{table}[!t]
\caption{Summary of repeatability tests - $4$ Joints, $6$ Keypoints, 3D motion}
\label{repeat_data_3d}
\begin{center}
\begin{tabular}{|c||c|}
\hline
& {\bf Proposed algorithm}\\
\hline
\hline
\bf{System rise time (s)} & $31.54 \pm 2.36$\\
\hline
\bf{System settling time} (s) & $40.76 \pm 3.63$\\
\hline
\bf{End effector rise time} (s) & $21.44 \pm 1.68$\\
\hline
\bf{End effector settling time} (s) & $33.66 \pm 3.72$\\
\hline
{\textbf{Overshoot (\%)}} & $1.74 \pm 0.59$\\
\hline

\end{tabular}
\end{center}
\end{table}

\begin{table}[!t]
\caption{Summary of repeatability tests - $3$ Joints, $5$ Keypoints, With occlusion}
\label{repeat_data_03_occ}
\begin{center}
\begin{tabular}{|c||c|}
\hline
& {\bf Proposed algorithm}\\
\hline
\hline
\bf{System rise time (s)} & $13.2 \pm 0.41$\\
\hline
\bf{System settling time} (s) & $19.78 \pm 1.05$\\
\hline
\bf{End effector rise time} (s) & $13.2 \pm 0.41$\\
\hline
\bf{End effector settling time} (s) & $16.10 \pm 0.26$\\
\hline
{\textbf{Overshoot (\%)}} & $5.48 \pm 0.34$\\
\hline

\end{tabular}
\end{center}
\end{table}

\subsection{Accuracy of the Online Interaction-Matrix Estimate}
\label{ssec:jacobian_eval}
 
The controller of \autoref{ssec:adaptive_vs} drives the robot with an online
estimate $\hat{J}$ of the combined interaction matrix, refit at every step by
window-based least squares from the most recent feature and command pairs, with no
robot or camera model. Convergence of the loop on its own does not establish that
this estimate is accurate, so we evaluate $\hat{J}$ directly across five out-of-plane
control runs conducted against the controlled backdrop. The conventional reference for such an evaluation, the analytical composite Jacobian $J = J_{\text{img}} J_{\text{robot}}$, requires the kinematic model
and the camera calibration that our method is designed to avoid, and it does not exist
on the weakly modeled platforms the method targets, such as the origami arm of
\autoref{ssec:origami_extension}. We therefore assess $\hat{J}$ with two measures
computed from quantities the controller already records, the condition number of the
estimate and the agreement between the feature motion it predicts and the feature
motion that is observed.
 
The condition number $\kappa(\hat{J}) = \sigma_{\max}/\sigma_{\min}$ reports whether the
estimate stays full rank and well spread across the feature space over the run. For the
second measure we compare, over a short horizon, the feature displacement the estimate
predicts for the commanded actuator motion, $\hat{J}\,\Delta r$, against the observed
displacement $\Delta s$, and report the cosine of the angle between them. The cosine is
the control-relevant quantity, because an image-based servo loop reduces the feature
error as long as the estimated mapping points the commanded velocity within ninety
degrees of the descent direction, which corresponds to a positive cosine.
 
\begin{figure}[t]
\centering
\includegraphics[width=\linewidth]{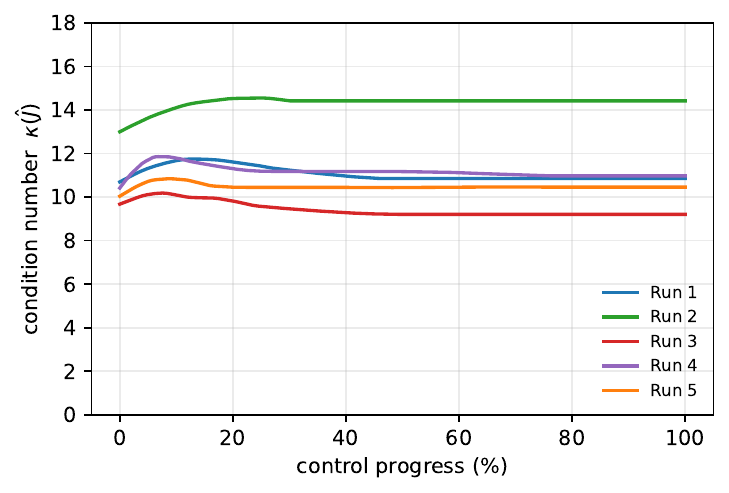}
\caption{Conditioning of the online interaction-matrix estimate over the control
runs. The condition number of $\hat{J}$ stays in a narrow well-conditioned band and
is flat over each run, including at convergence.}
\label{fig:jacobian_conditioning}
\end{figure}

\begin{figure}[t]
\centering
\includegraphics[width=\linewidth]{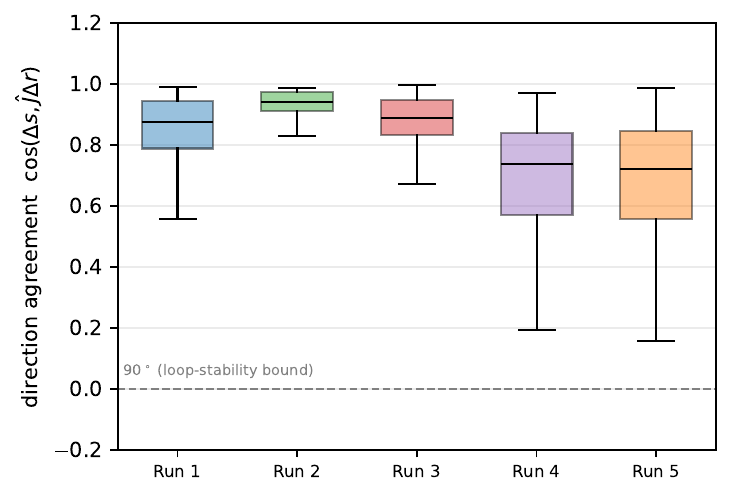}
\caption{Predicted versus observed feature-motion direction across the control runs.
The cosine between the predicted and the observed feature-motion direction stays above
the ninety-degree loop-stability bound (dashed) for nearly all steps, so the estimate
keeps the commanded velocity in a descent direction throughout.}
\label{fig:jacobian_direction}
\end{figure}

The condition number stays between 9 and 15 across all five runs and is flat within each
run, including at convergence, with a mean per-run median of 11.2
(\autoref{fig:jacobian_conditioning}). The estimate does not approach rank deficiency at any
point. The predicted direction of feature motion agrees with the observed direction in
97.7\% of steps on average, ranging from 91\% to 100\% across runs, with a median
direction cosine of 0.83 (\autoref{fig:jacobian_direction}, \autoref{tab:jacobian_eval}). The
estimate is therefore well-conditioned and directionally reliable throughout, which is
the property the servo law depends on and which explains why the loop converges without a
model.
 
The magnitude of the prediction is less consistent than its direction. On the shorter
runs the estimate predicts the size of the feature motion well, while on the longer runs this magnitude prediction degrades as the local linear estimate becomes stale over larger configuration changes. This behavior is expected for an estimate that is re-fit at every step, and it bears directly on the two regimes raised in review. Faster convergence increases the per-step feature motion relative to the detector's pixel-quantization floor and so improves the signal available to the estimator, which means speed favors the estimate rather than stressing it. Continuous trajectory tracking places the opposite demand, a sustained requirement on magnitude accuracy under exactly the conditions where the local estimate goes stale. Our experiments are point-to-point regulation rather than trajectory tracking, and we identify model-free interaction-matrix estimation for continuous trajectory tracking as a direction for future work.
 
\begin{figure*}[!t]
    \centering
    \includegraphics[width=\textwidth, center, trim=0 0 0 0, clip]{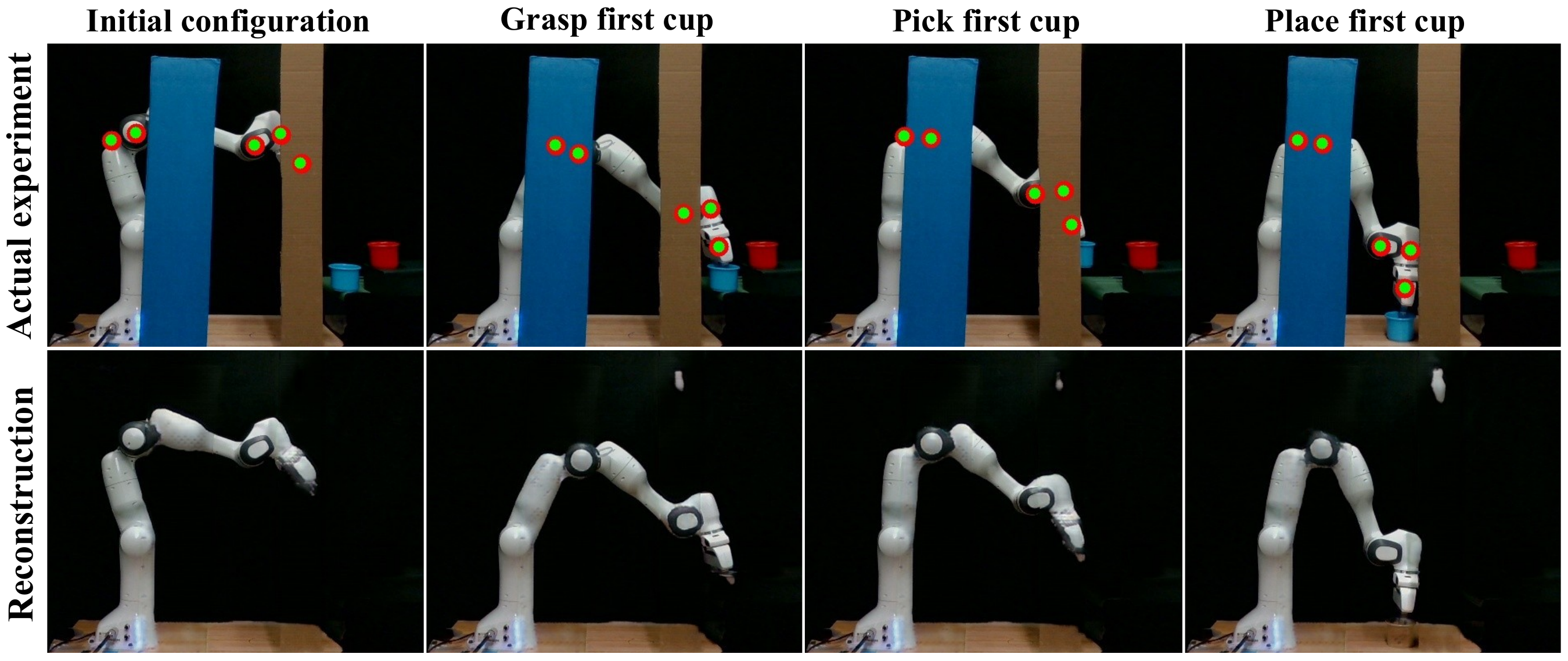}
    \vspace{-5pt}
    \includegraphics[width=\textwidth, center, trim=0 0 0 0, clip]{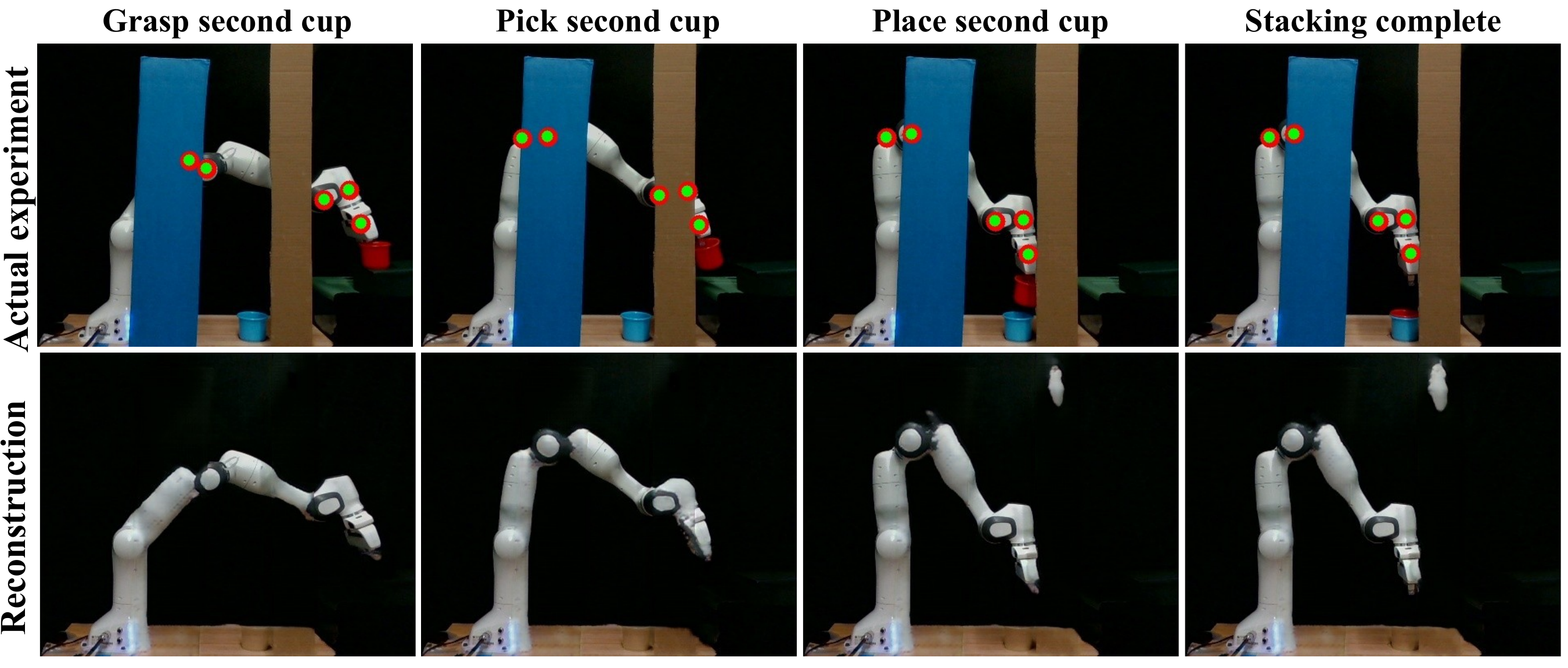}
    \vspace{-5pt}
     \caption{
    A real-time cup stacking experiment using adaptive visual servoing under occlusion. The robotic manipulator stacks cups while tracking body keypoints (red-on-green circles), which are used as visual features for adaptive control. These keypoints are detected from reconstructed images generated via inpainting. The rows labeled “Real-time experiment” show the original occluded images, while the “Reconstructed view” rows display the inpainted images used for keypoint detection. This approach enables accurate and continuous control despite partial visual occlusions during the task.
    }
    \label{fig:cup_stacking}
    \vspace{-0.5em}
\end{figure*}

\subsection{Repeatability:}
To evaluate the repeatability of our proposed method, we conducted control experiments five times using the same initial and target positions for both the $3$-keypoints with $2$-joints and $5$-keypoints with $3$ joints setups under full visibility. We applied the same protocol to the 3D setup, where $6$ keypoints are used as visual features to control four joints (three planar $+$ one spatial). The results are summarized in \autoref{repeat_data_02} and \autoref{repeat_data_03} for the respective planar full visibility cases, and in \autoref{repeat_data_3d} for the 3D full-visibility case. We also conducted the same repeatability test on a randomly selected experiment with occlusion in the workspace, with the results summarized in \autoref{repeat_data_03_occ}. Across all the runs, the rise time, settling time, and overshoot values showed only minor variations, demonstrating consistent performance.

\subsection{Real-World Demonstration - Cup Stacking:}

To demonstrate the effectiveness of our overall pipeline, we present a real-time cup-stacking experiment shown in \autoref{fig:cup_stacking}. The Franka Emika Panda robot successfully completes the task under partial occlusion by relying on detected keypoints along its body as visual features for control. These keypoints are predicted in real time using our proposed pipeline. Even when significant portions of the arm are occluded, the system is able to reconstruct the missing regions, detect keypoints accurately, and maintain continuous control throughout the sequence. The figure displays both the original occluded images and their corresponding inpainted reconstructions, illustrating the robustness of our method in maintaining keypoint detection under uncertain conditions. See Multimedia Extension $8$ for the demonstration video.

\begin{figure*}[!b]
    \centering
    \includegraphics[width=\textwidth, center, trim=0 0 0 0, clip]{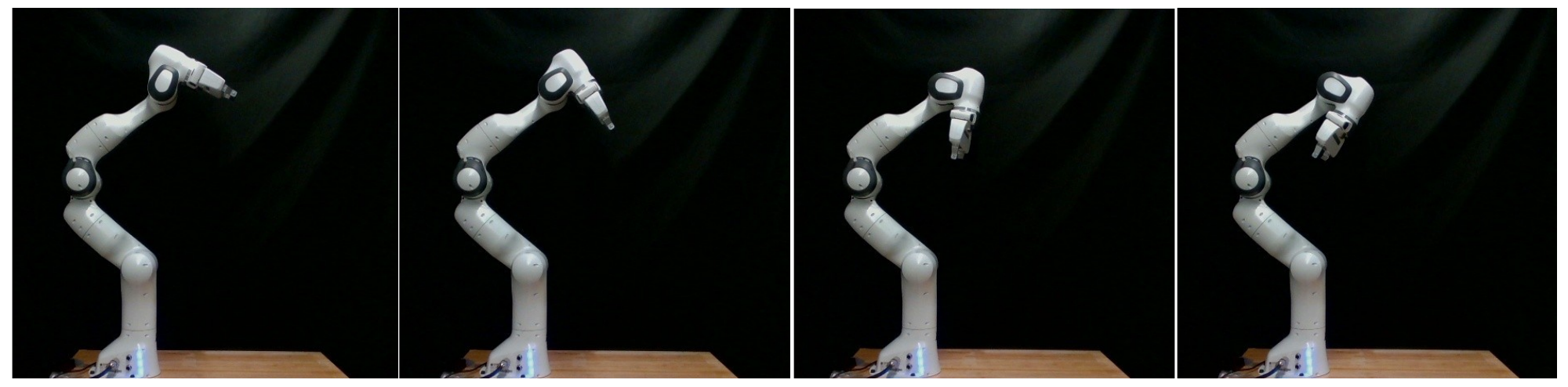}
    \caption{
    Example of pose ambiguity introduced by occlusion. All four images show the same configuration for the first two joints, but the occluded end-effector is reconstructed differently in each case, resulting in multiple plausible but incorrect poses for the last joint.
    }
    \label{pose_input}
\end{figure*}

\section{Failure Use Cases and Observations} \label{sec:failures}
    % Failure Use Cases and Observations

\begin{figure}[!t]
    \centering
    \includegraphics[width=\linewidth, center, trim=0 0 0 0, clip]{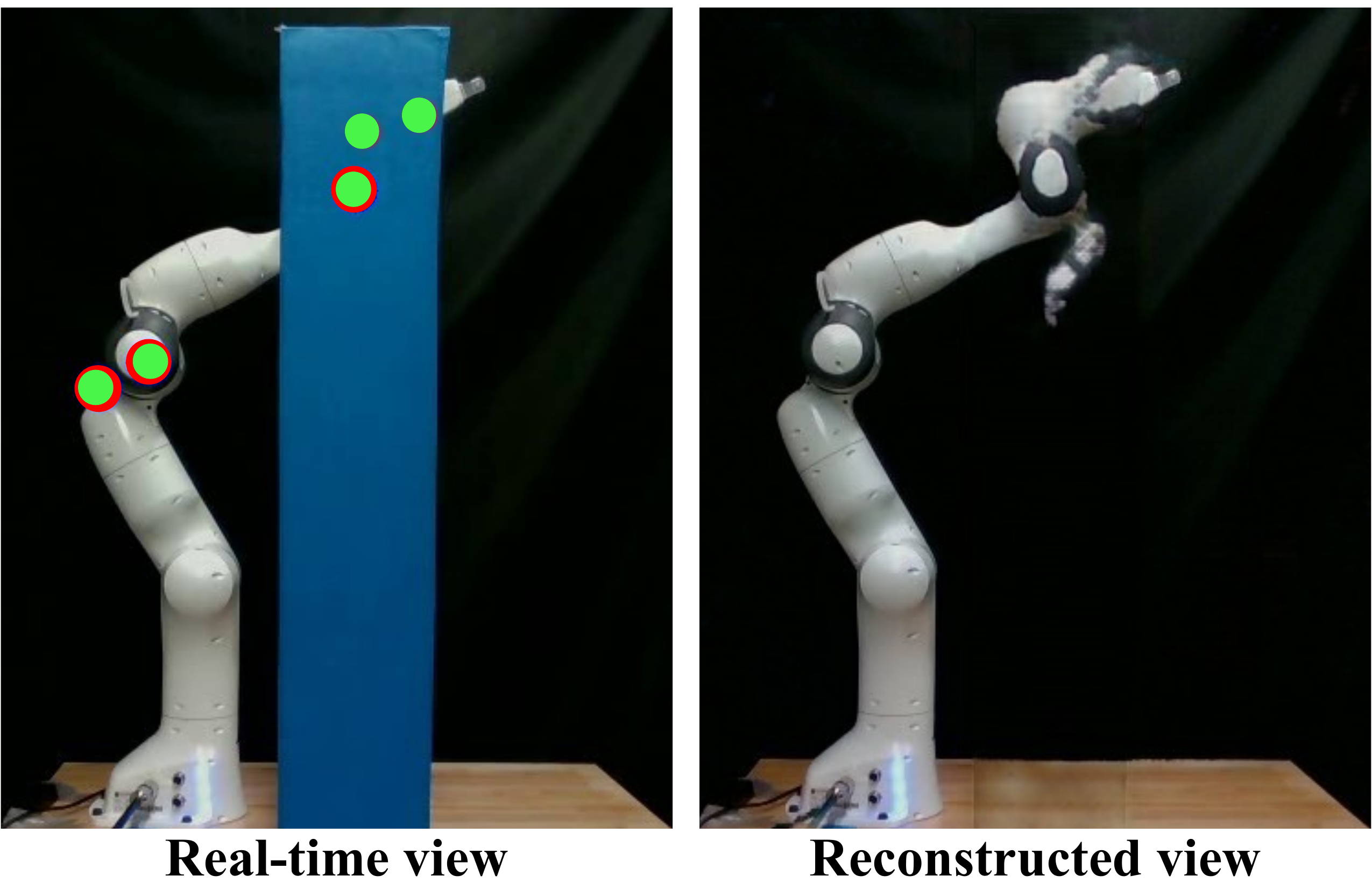}
    \caption{
    The first image contains a large occlusion completely covering the end effector. The inpainting model reconstructs two end effector regions due to pose ambiguity as seen in second image. As a result, the initial keypoint prediction (red circles) is missed. Despite this, the UKF, leveraging distance-based thresholding, is able to generate a reasonably accurate estimate of the keypoint (green circles)
    }
    \label{pose_amb}
\end{figure}

\begin{figure}[!t]
    \centering
    \includegraphics[width=\linewidth, center, trim=0 0 0 0, clip]{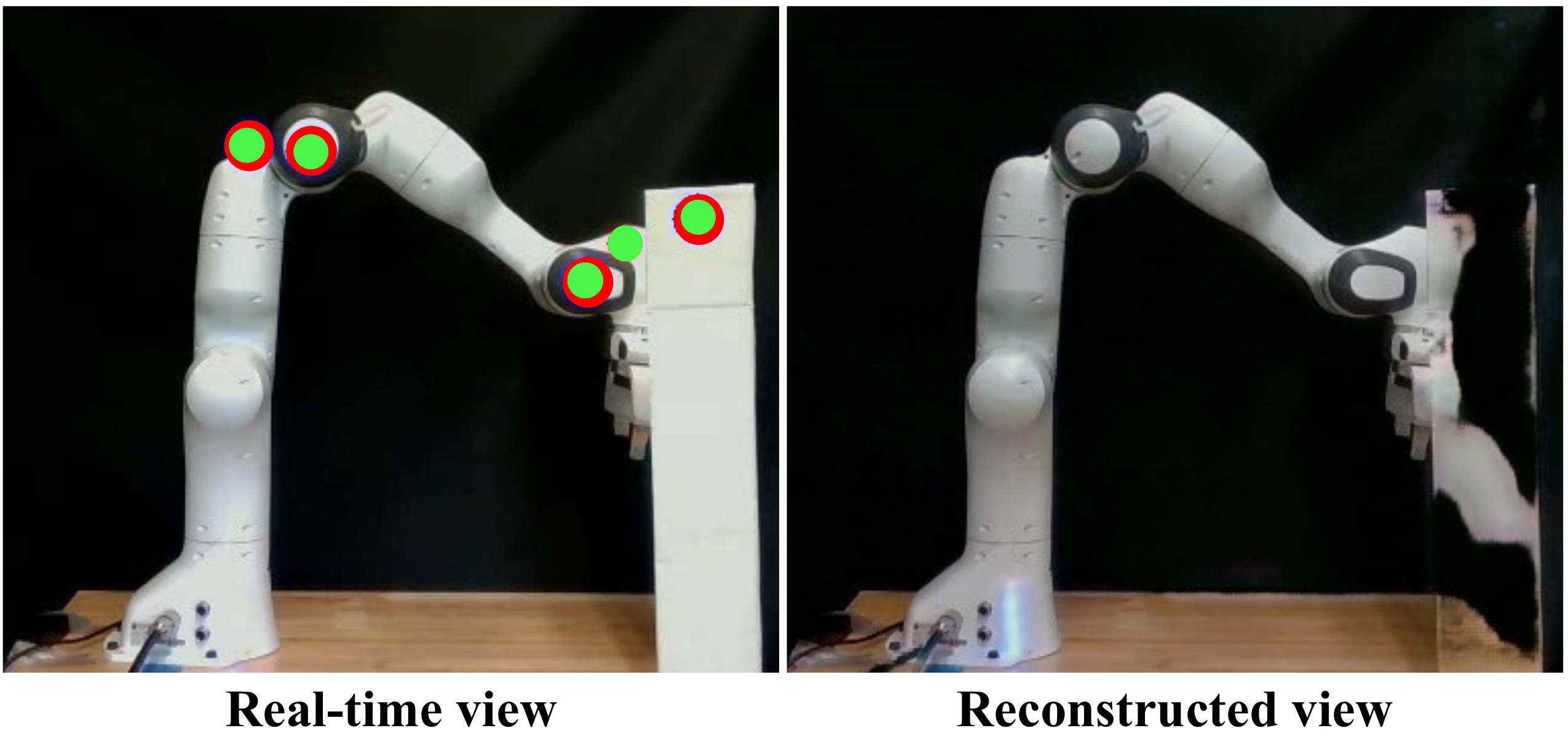}
    \caption{
    The first image contains a bright white occlusion that significantly affects reconstruction quality, as seen in the second image. As a result, the initial keypoint prediction (red circle) is severely mislocalized, falling outside the correction range of the UKF (green circle).
    }
    \label{bright}
\end{figure}

Despite the improved robustness introduced by our inpainting-based pipeline and the integration of a UKF for keypoint correction, we observe three important failure cases that can significantly impact system performance in real-world scenarios. These failure modes are described below. 
\subsection{Pose Ambiguity Under End-Effector Occlusion}
One common failure mode arises when the end-effector of the robot is completely occluded. In such cases, the inpainting model attempts to hallucinate the missing structure based on learned priors. As illustrated in \autoref{pose_input}, these learned priors come from the network’s exposure to thousands of robot images during training, leading it to ``fill in" missing regions, such as an occluded end-effector—based on what appears kinematically plausible. However, without sufficient contextual cues, the model may reconstruct the end-effector in multiple plausible but incorrect poses, introducing \textit{pose ambiguity}. This ambiguity becomes especially problematic in downstream keypoint detection, as different reconstructed poses lead to spatially inconsistent keypoint predictions. Since this error originates from the inpainting stage, where the image itself is incorrectly reconstructed, filtering methods like UKF cannot fully correct it. The UKF is designed to handle temporary occlusions or small localization errors by smoothing keypoint trajectories over time. However, when the structure of the robot is reconstructed incorrectly, such as showing a duplicated (\autoref{pose_amb}) or misplaced end-effector, the keypoint predictions are fundamentally wrong. In such cases, the UKF cannot recover the correct location because the visual input itself is misleading. This ambiguity is only partially addressed using the \hyperlink{ukf_thresholding}{distance-based thresholding} heuristic described in the UKF logic. As illustrated in \autoref{pose_amb}, the reconstruction produces two end effectors, causing the keypoint detection to fail. However, the UKF, aided by distance thresholding, is still able to provide a reasonably localized estimate.

\subsection{Reconstruction Errors from Occlusions Matching Robot Appearance}
A second critical observation involves the presence of objects within the robot's workspace that are both visually similar in color and significantly brighter than the robot body. When such distractor objects lie in the occluded regions or overlap with the robot structure, the inpainting model often incorrectly reconstructs parts of the robot by blending features from the distractor. This leads to structural inaccuracies in the recovered image, such as spurious links or distorted joints. These visual errors propagate to the keypoint detection stage and result in mislocalized or missing keypoints. Unlike mild occlusions, the degradation here is too severe for the UKF to correct, as the initial measurements are already erroneous. Consequently, this failure mode poses a significant challenge to reliable perception in scenes with high-contrast or visually similar clutter. In \autoref{bright}, the occlusion is extremely bright and white, leading to poor image reconstruction. As a result, one of the keypoints is predicted far from its true location, making it too inaccurate for the UKF to correct.

\begin{figure}[!t]
    \centering
    \includegraphics[width=0.5\textwidth]{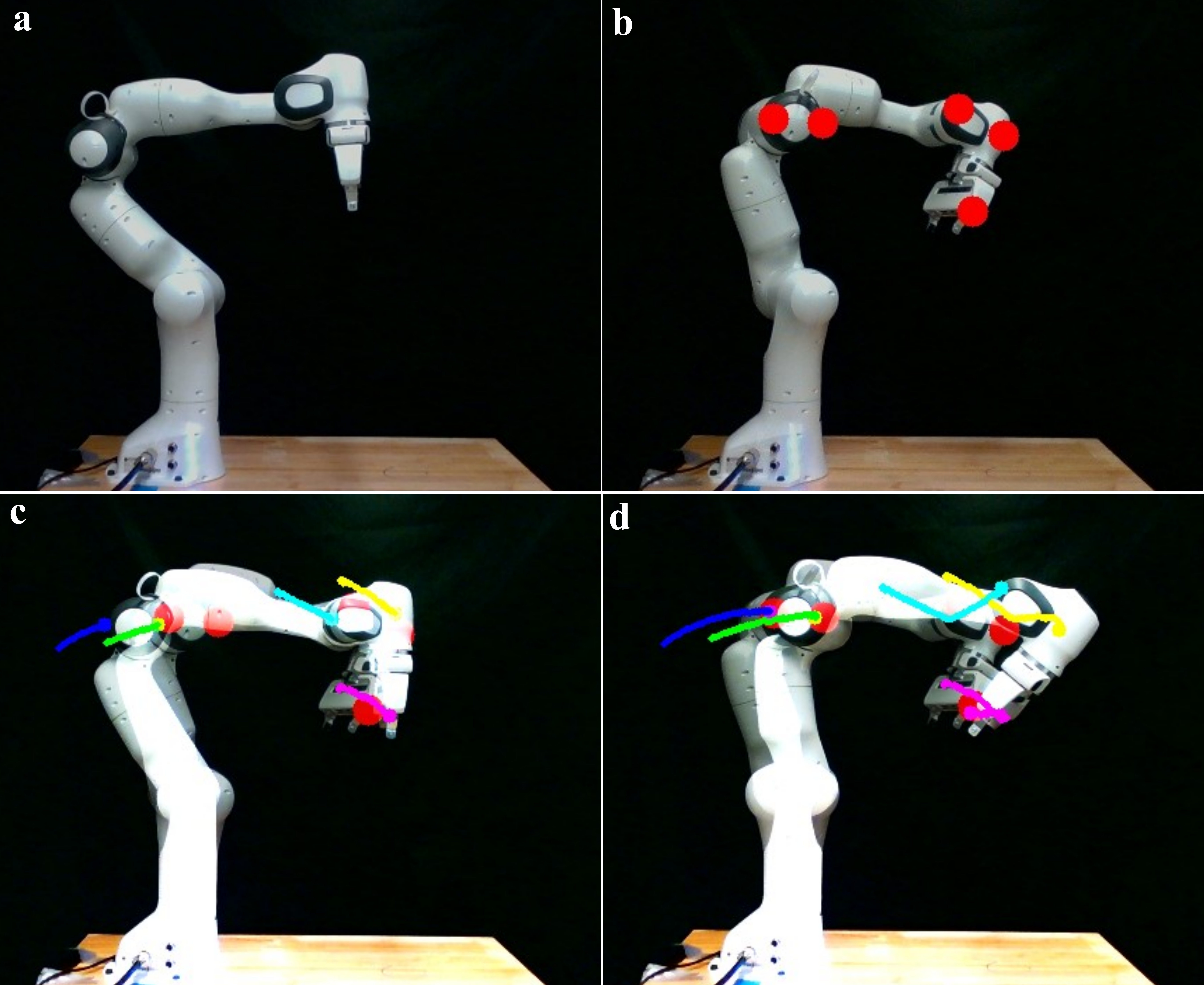}
    \caption{
    (a) Initial image. (b) Goal image with desired keypoints (red). (c–d) Intermediate images during control: although the goal requires motion \emph{away} from the camera, the image-space arrangement of predicted keypoints becomes nearly indistinguishable from the goal, creating a depth-direction ambiguity that initially drives the arm \emph{toward} the camera. 
    }
    \label{fig:pose_amb_3d}
\end{figure}

\subsection{Pose Ambiguity in 3D Motion}
This subsection highlights a representative failure mode of monocular IBVS in 3D—depth-direction (pose) ambiguity, and how it is fixed in this paper.

The robot configurations in \autoref{fig:pose_amb_3d}(a) and (b) are the initial and goal images for an out-of-plane control experiment. \autoref{fig:pose_amb_3d}(c), (d) show intermediate images during the run. Although the goal requires the robot to move \textit{away} from the camera, the controller initially drives it \textit{toward} the camera. The reason is that the predicted image-space keypoints in these intermediate views are nearly indistinguishable from those of the goal, creating a depth-direction ambiguity that misguides the IBVS update, an effect well known in monocular IBVS. See Multimedia Extension $9$ to observe the precise moment this depth-direction failure occurs.

We rectified this by increasing the number and spatial coverage of features: using $8$ keypoints distributed along the arm (\autoref{fig:3d_wkflow}). The richer, more spread-out keypoint set provides stronger and more informative image configuration, disambiguating the depth direction. With this modification, the robot follows the correct motion (away from the camera) and remains stable even under pronounced depth variations.

\subsection{Frequency of the Observed Failure Modes}
We observe these failure modes at a limited rate across our experiments. For the five occluded sequences, the raw detection accuracy after inpainting ranges from $99.53\%$ in the mildest case to $77.23\%$ in the most severe (\autoref{tab:kp_det_acc}). The two lowest values, for \texttt{exp\_03} and \texttt{exp\_05}, correspond to the sequences with the heaviest end-effector occlusion, which is the condition that produces the pose-ambiguity and bright-distractor errors described above. The UKF recovers part of this loss and raises the worst case to $83.98\%$. On the $5$ control keypoints, the full pipeline keeps the failure rate at $5.09\%$ at the $10$ px threshold (\autoref{tab:inpaint_ablation}), so the reconstruction errors that the filter cannot correct affect roughly one keypoint estimate in twenty. We observe the same pattern in the background-clutter experiment, where a single sequence, in which the arm sweeps across the fluorescent tube, contributes a disproportionate share of the raw detection errors, while the gate and the filter absorb them in the final output (\autoref{tab:bkgrnd_robustness}).

\vspace*{-0.7\baselineskip}
\section{Practical Usage: Adapting the Framework to Other Manipulators} \label{sec:prac_app}
    % Practical Usage: Adapting the Framework to Other Manipulators

This procedure can be easily applied to any manipulator (rigid, continuum or soft) with a fixed eye-to-hand camera. Below is a concise summary of steps.

\medskip
\noindent\textbf{Step 1}: Command small velocities to desired actuators computed by \autoref{comp_vel} to traverse the robot in visible workspace while recording RGB frames at a constant rate. We refer to the saved images as \texttt{Image\_set\_01}

\noindent\textbf{Step 2}: Train/fine-tune the modified LaMa inpainting model on \texttt{Image\_set\_01} to learn reconstruction of the robot's body. We refer to the resulting model as \texttt{Model\_01}.

\noindent\textbf{Step 3}: Place ArUco markers along the robot's body. Quantity and placement of the markers depend on the application. Repeat Step 1 to capture a new set of images referred to here as \texttt{Image\_set\_02}. Detect and store marker centers as keypoint labels.

\noindent\textbf{Step 4}: Run \texttt{Model\_01} on \texttt{Image\_set\_02} to remove markers, producing a set of reconstructed images referred to here as \texttt{Image\_set\_03}. Pair each reconstructed image with its keypoint labels to build the keypoint dataset.

\noindent\textbf{Step 5}: Train the modified Keypoint R-CNN on \texttt{Image\_set\_03} and corresponding labels to obtain a model referred to here as \texttt{Model\_02}, which tracks keypoints on robot's body using a single camera in real time.

\noindent\textbf{Step 6}: Utilize the keypoints detected by \texttt{Model\_02} to control the manipulator using an adaptive visual servoing scheme described in \autoref{ssec:adaptive_vs}

\noindent\textbf{Step 7}:
Create \texttt{Image\_set\_04} by artificially occluding parts of the manipulator in \texttt{Image\_set\_01} (occlusion can be of varied shapes/sizes/textures). Train the Attention U-Net–based GAN inpainting model described in \autoref{ssec:gan-desc} to obtain the model we referred to here as \texttt{Model\_03}.

\noindent\textbf{Step 8}: Integrate \texttt{Model\_02}, \texttt{Model\_03}, and a UKF to run concurrently for robust, real-time keypoint detection under uncertainty and partial occlusion.

\subsection{Case Study: Extending the Pipeline to a Soft Origami Arm}
\label{ssec:origami_extension}
 
Our motivation targets platforms where accurate kinematic models and joint sensing are difficult to obtain, such as soft and continuum arms. To show that the perception pipeline carries beyond the rigid Panda arm, we applied it to a two-module and a three-module soft origami arm. We adapt the data collection, the inpainting-based keypoint training, and the temporal tracking stages to this platform and evaluate keypoint detection and tracking on continuous video. We do not close the control loop on the origami arm in this work, so this case study demonstrates transfer of the perception components rather than a control result.
 
Following the data collection of \autoref{ssec:data_collection}, we moved the arm through its visible workspace by applying small velocities to the two camera-facing tendons computed with \autoref{comp_vel}, and recorded images at a fixed rate. We fine-tuned the LaMa inpainting model \citep{Suvorov_2022_WACV} starting from the weights already trained on Panda images, which gives a more structured initialization than random weights for the translucent appearance of the origami arm. Because that translucent body is harder to resolve than the Panda, we captured images at a higher resolution and cropped them to $720 \times 720$ for training. We placed small red square markers along the two-module arm and larger blue markers along the three-module arm, choosing the color and size for visibility against the body and across the larger depth range of the spatial workspace. For each image we detect the marker centers, build a binary mask around each marker, and reconstruct a marker-free image with refinement enabled \citep{kulshreshtha2022feature}. The marker centers in the original images become the keypoint labels.
 
We store the marker centers as keypoint annotations together with the reconstructed images, using the same JSON format and fixed-square bounding boxes as \autoref{ssec:data_gen}. We define five keypoints along the two-module arm for planar motion and seven keypoints along the three-module arm for spatial motion. On these datasets we train the same \textit{keypointrcnn\_resnet50\_fpn} detector used in \autoref{ssec:kp_train}, with the optimizer and hyperparameters selected through Optuna \citep{akiba2019optuna} and the ResNet50 backbone initialized from COCO weights \citep{Lin2014}. The trained three-module origami keypoint detector is archived on Zenodo \textit{(DOI: 10.5281/zenodo.20990811)}.
 
\begin{figure*}[!t]
    \centering
    \includegraphics[width=\textwidth]{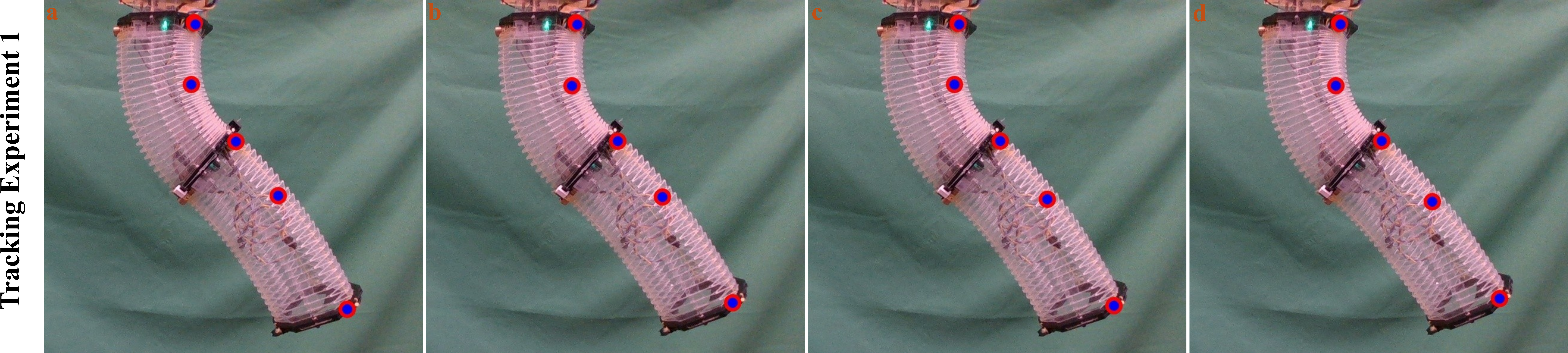}
 
    \includegraphics[width=\textwidth]{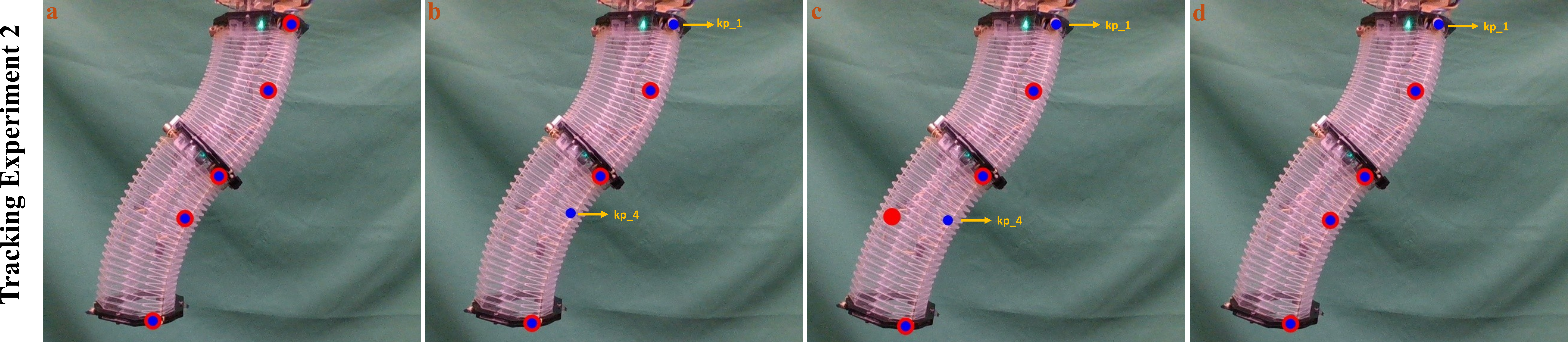}
    \caption{
    Keypoint tracking on the two-module origami arm across two experiments. In the top row, the trained Keypoint R-CNN detects all keypoints along the body from frame (a) to (d), shown as blue circles on the red markers. In the bottom row, frame (a) shows correct detection. In frame (b) the detector misses keypoints kp\_1 and kp\_4, which the temporal tracking module reconstructs. In frames (c) and (d), kp\_1 stays undetected by the network and is held by temporal tracking, while kp\_4 is detected but severely mislocalized in frame (c) and then corrected by the refinement.
    }
    \label{fig:origami_tracking_2d}
\end{figure*}

At runtime we pass each video frame through the trained detector to obtain per-frame keypoints along the arm. Per-frame detection on the compliant arm is less reliable than on the Panda, with keypoints occasionally missed or mislocalized under larger deformations and spatial motion. We therefore add a lightweight image-based temporal refinement that maintains a running estimate of each keypoint. The refinement accepts a new detection when it is present, sufficiently confident, and within a small displacement of the previous estimate, and otherwise propagates the previous estimate forward in time. This step uses no motion model or Kalman filter and relies on the close spacing of consecutive frames and the local smoothness of the keypoint trajectories. Across the sequences we evaluated, the detector and this refinement together produce stable keypoint tracks for both the planar two-module motion and the spatial three-module motion, and the refinement recovers keypoints that the detector intermittently misses. \autoref{fig:origami_workflow} shows the adapted data generation pipeline. \autoref{fig:origami_tracking_2d} shows representative tracking frames for the planar two-module motion, and \autoref{fig:origami_tracking_3d} shows tracking during the spatial three-module motion. See Multimedia Extension $13$ for the demonstration videos of the $3$ modules spatial motion.
 
This case study confirms that the perception components of our framework, namely the data collection, the inpainting-based keypoint training, and the temporal tracking, transfer from a rigid industrial manipulator to a compliant, weakly modeled origami arm with minimal assumptions about its kinematics. Closing the loop with the adaptive visual servoing scheme of \autoref{ssec:adaptive_vs} on the soft arm is the natural next step, which we leave for future work.
 
\begin{figure*}[!t]
    \centering
    \includegraphics[width=\textwidth]{images/origami_wkflow.pdf}
    \caption{
    Extension of the markerless data generation pipeline to the soft origami arm. (a) Sample images of the arm with red markers. (b) Corresponding binary masks of the markers. (c) Reconstructed images with the markers removed by inpainting. (d) Sample annotations of the marker centers and bounding boxes used to build the Keypoint R-CNN training set. (e) and (f) show two configurations with keypoints predicted in real time by the trained Keypoint R-CNN on the origami arm.
    }
    \label{fig:origami_workflow}
\end{figure*}
 
\begin{figure*}[!t]
    \centering
    \includegraphics[width=\textwidth]{images/3D_origami_tracker.pdf}
    \caption{
    Keypoint tracking on the three-module origami arm during spatial motion. Frames (a) to (c) show different arm configurations, where blue circles denote the keypoints predicted in real time by the trained Keypoint R-CNN and the temporal tracking pipeline.
    }
    \label{fig:origami_tracking_3d}
\end{figure*}

\section{Conclusion and Future Work} \label{sec:conclusion}
    % Conclusion and Future Work

In this work, we present a markerless, model-free, full-body vision-based control pipeline that enables robotic manipulators to operate effectively in unstructured and partially observable environments. To build the system we first used ArUco markers on designated locations on the robot's body for identifying keypoints or features, and then artificially removed the markers using a deep learning inpainting framework. By doing so, we completely eliminated the need for camera calibration as well as any prior knowledge of the kinematic model of the robot. We created and open-sourced the data collection pipeline using our proposed method to build large datasets for different robotic manipulators. Using our method, we collected a dataset for the Franka Emika Panda robot and trained a \textit{keypointrcnn\_resnet50\_fpn} model on it. We obtained highly accurate keypoint detection results in real-time, and we were able to control the robot by utilizing the keypoints inferred from this model. To handle occlusions during control, we introduced an attention-based U-Net generator trained using a Wasserstein GAN with gradient penalty, capable of reconstructing the occluded regions of the robot image without using explicit occlusion masks during training or inference, and we verified that this reconstruction step is necessary by comparing it against temporal filtering alone and against a detector retrained on occluded images. We then passed the reconstructed image through a modified \textit{keypointrcnn\_resnet50\_fpn} model to predict keypoints along the robot's body in real-time. To further enhance robustness, we integrated an Unscented Kalman Filter into the pipeline to correct for missing or inaccurately localized keypoints caused by reconstruction noise or pose ambiguity, ensuring smooth and continuous control. These keypoints are then used as natural visual features for an adaptive image-based visual servoing scheme, where the interaction matrix is estimated online without relying on robot or camera models.

Using this framework, we successfully controlled the Panda arm under full visibility in both within-plane (2D) and out-of-plane (3D) motion, and under partial occlusions in 2D. Our results show that keypoints generated with the proposed method are as robust and reliable as those from our prior work, which relied on explicit modeling. We observed minimal performance degradation in the presence of moderate occlusions. Even with successful convergence, larger occlusions led to weaker control performance, mainly showing higher overshoot. We also evaluated the pipeline against a cluttered laboratory background absent from the training data, where the deployed detector localized the keypoints without retraining and the controller completed the out-of-plane task in that scene. We profiled the per-frame cost of the perception pipeline and measured $87.1$ ms with inpainting active and $65.9$ ms under full visibility, both of which stay within the $100$ ms budget of the control loop, so the system operates in real time. We also evaluated the online interaction-matrix estimate directly, since the analytical Jacobian that would normally serve as a reference is unavailable in our setting. The estimate stays well-conditioned across the control runs, with a mean condition number of $11.2$, and the direction of feature motion it predicts agrees with the observed direction in $97.7\%$ of the steps, which is the property the servo law depends on and which explains why the loop converges without a model. Notably, the control loop performs \textit{3D IBVS without depth sensing, camera calibration, or a kinematic robot model}, and uses only monocular image-space keypoints. The robot-specific structure it needs is learned by the detector from images rather than provided as a model. To the best of our knowledge, this is the first demonstration of out-of-plane IBVS on a manipulator under these conditions.

We identified a shortcoming of our approach in occlusion scenarios. When the scene contains a highly reflective or brightly colored object, or an object with a color similar to that of the robot, the inpainting model occasionally reconstructs the robot incorrectly in those regions. This results in invalid keypoints being predicted far from their true locations, which, in severe cases, cannot be reliably corrected even by the Unscented Kalman Filter. Such errors can compromise control performance and highlight the need for improved robustness in ambiguous visual contexts. We also found that occlusion handling does not transfer across backgrounds. The clean-trained detector generalizes to a cluttered scene without retraining on the keypoints that remain visible, but inpainting reconstructs the occluded regions only against the background the generator was trained on, and retraining the detector on occluded images does not carry to a new background either, as shown in \autoref{ssec:occ_clutter}. Occlusion under a changed background therefore remains unsolved. To address both limitations, we aim to shift from reconstructing the robot's image to directly reconstructing the keypoints using spatial context. Specifically, we plan to employ a Siamese architecture \citep{koch2015siamese} using Graph Convolution Network \citep{zhang2019graph} and GraphSage Network \citep{hamilton2017inductive} to model the spatial relationships between keypoints, enabling more accurate prediction of missing or ambiguous keypoints without relying on full image reconstruction. Because reconstruction in keypoint space does not depend on the appearance of the background, we expect it to handle occlusion more consistently across environments. We also identify model-free interaction-matrix estimation for continuous trajectory tracking as a direction for future work, since the magnitude of the estimate's prediction is less consistent than its direction on the longer runs. Beyond the Panda arm, we applied the perception pipeline to a two-module and a three-module soft origami arm, where the data collection, the inpainting-based keypoint training, and the temporal tracking all transferred to a compliant platform without a kinematic model. We will also extend the demonstrated 3D control to operate under occlusion and close the loop on soft and continuum manipulators, building on the perception results we report for the origami arm in \autoref{ssec:origami_extension}.
\section{Multimedia Extensions} \label{sec:multimedia}
    % Multimedia Extensions

\begin{itemize}
    \item \textbf{Multimedia Extension $1$}: Keypoint detection accuracy under partial visibility
    \item \textbf{Multimedia Extension $2$}: Control experiments on the Franka Emika Panda with $3$ keypoints and $2$ planar joints
    \item \textbf{Multimedia Extension $3$}: Control experiments on the Franka Emika Panda with $5$ keypoints and $3$ planar joints under full visibility
    \item \textbf{Multimedia Extension $4$}: Control experiments on the Franka Emika Panda with $5$ keypoints and $3$ planar joints under low-coverage partial occlusion
    \item \textbf{Multimedia Extension $5$}: Control experiments on the Franka Emika Panda with $5$ keypoints and $3$ planar joints under high-coverage partial occlusion
    \item \textbf{Multimedia Extension $6$}: Qualitative comparison of control trajectories and performance under variable visibility
    \item \textbf{Multimedia Extension $7$}: Out-of-plane control experiments on the Franka Emika Panda with $6$ keypoints and $3$ planar and $1$ spatial joints
    \item \textbf{Multimedia Extension $8$}: Real-world demonstration of vision-based control using image keypoints without robot or camera Model under partial occlusion
    \item \textbf{Multimedia Extension $9$}: Failure case in 3D motion - Pose ambiguity
    \item \textbf{Multimedia Extension $10$}: Keypoint detection accuracy in cluttered background
    \item \textbf{Multimedia Extension $11$}: Keypoint detection under occlusion in a cluttered background
    \item \textbf{Multimedia Extension $12$}: Out-of-plane control experiments on the Franka Emika Panda with 6 keypoints and 3 planar and 1 spatial joints with cluttered background
    \item \textbf{Multimedia Extension $13$}: Extension of the keypoint pipeline to a soft origami arm    
\end{itemize}
\section{Data and Code Availability} \label{sec:data_and_code}
    % Data and Code Availability

\textbf{Data:}\\
Research datasets and trained models are deposited on Zenodo. During peer review, these records are given restricted access. A supplementary CSV file is added with the submission where the editors and reviewers have been provided with each record's DOI, title and temporary access link. The datasets and models will be made public at the end of the final review process.
\\
\\
\textbf{Code:} \\
All relevant codes and scripts are available at \url{https://github.com/Jani-C/KPDataGenerator}

% \FloatBarrier
\bibliographystyle{SageH}
\bibliography{references}

\end{document}